\documentclass{src/bmvc2k}

\title{Taming Visually Guided Sound Generation}

\addauthor{Vladimir Iashin}{vladimir.iashin@tuni.fi}{1}
\addauthor{Esa Rahtu}{esa.rahtu@tuni.fi}{1}

\addinstitution{
 Computing Sciences\\
 Tampere University\\
 Tampere, Finland
}

\runninghead{Iashin, Rahtu}{Taming Visually Guided Sound Generation}

\usepackage{tikz}
\newcommand\PlaceText[3]{%
\begin{tikzpicture}[remember picture,overlay]
\node[outer sep=0pt,inner sep=0pt,anchor=south west] 
  at ([xshift=#1,yshift=#2]current page.south west) {#3};
\end{tikzpicture}%
}
\usepackage{media9}
\usepackage{transparent}
\usepackage{fontawesome}
\usepackage{booktabs}
\usepackage{multirow}
\usepackage{colortbl}
\usepackage{pifont}
\definecolor{vggsound2vggsound}{HTML}{99FFFF}
\definecolor{vggsound2vas}{HTML}{CC99FF}
\definecolor{vas2vas}{HTML}{FFCC99}

\def\eg{\emph{e.g}\bmvaOneDot}

\def\etal{\emph{et al}\bmvaOneDot}

\newcommand{\comma}{, \, }
\newcommand{\R}{\mathrm{R}}
\newcommand{\inR}{\in \mathrm{R}}
\DeclareMathOperator{\argmin}{arg\,min}

\begin{document}
\maketitle
\vspace{-2.0em}
\begin{figure*}[ht]
\begin{center}
\includegraphics[width=\linewidth]{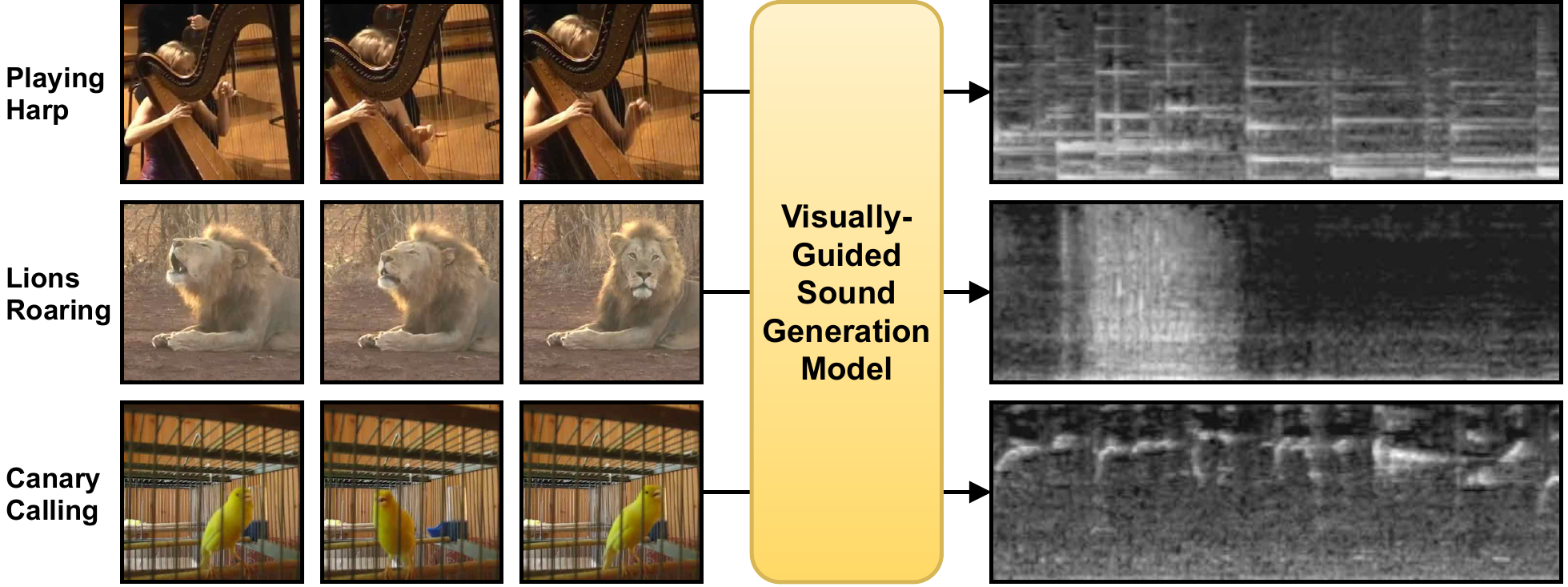}
\PlaceText{105.08mm}{170.61mm}{
\includemedia[ addresource=mp3/harp.mp3, flashvars={ source=mp3/harp.mp3 &autoPlay=false } ]{
\transparent{0.8} \color{green} \tiny \parbox{\dimexpr 0.4\linewidth-2\fboxsep-2\fboxrule\relax}{\faPlayCircleO~\textsf{Click to Play \\ in Adobe Reader \\[2.5em]}}}{APlayer.swf}
}
\PlaceText{105.08mm}{154.11mm}{
\includemedia[ addresource=mp3/lion.mp3, flashvars={ source=mp3/lion.mp3 &autoPlay=false } ]{
\transparent{0.8} \color{green} \tiny \parbox{\dimexpr 0.4\linewidth-2\fboxsep-2\fboxrule\relax}{\faPlayCircleO~\textsf{Click to Play \\ in Adobe Reader \\[2.5em]}}}{APlayer.swf}
}
\PlaceText{105.08mm}{138.11mm}{
\includemedia[ addresource=mp3/canary_calling.mp3, flashvars={ source=mp3/canary_calling.mp3 &autoPlay=false } ]{
\transparent{0.8} \color{green} \tiny \parbox{\dimexpr 0.4\linewidth-2\fboxsep-2\fboxrule\relax}{\faPlayCircleO~\textsf{Click to Play \\ Adobe Reader \\[2.5em]}}}{APlayer.swf}
}
\vspace{-1.5em}
\end{center}
\vspace{-1.5em}
   \caption{\normalsize A single model supports the generation of visually guided, high-fidelity sounds for multiple classes from an open-domain dataset faster than the time it will take to play it.} 
\label{fig:main}
\end{figure*}
\vspace{-1.5em}
\begin{abstract}
\vspace{-1ex}
Recent advances in visually-induced audio generation are based on sampling short, low-fidelity, and one-class sounds.
Moreover, sampling 1 second of audio from the state-of-the-art model takes minutes on a high-end GPU.
In this work, we propose a single model capable of generating visually relevant, high-fidelity sounds prompted with a set of frames from open-domain videos in less time than it takes to play it on a single GPU.

We train a transformer to sample a new spectrogram from the pre-trained spectrogram codebook given the set of video features.
The codebook is obtained using a variant of VQGAN trained to produce a compact sampling space with a novel spectrogram-based perceptual loss.
The generated spectrogram is transformed into a waveform using a window-based GAN that significantly speeds up generation.
Considering the lack of metrics for automatic evaluation of generated spectrograms, we also build a family of metrics called FID and MKL.
These metrics are based on a novel sound classifier, called Melception, and designed to evaluate the fidelity and relevance of open-domain samples.

Both qualitative and quantitative studies are conducted on small- and large-scale datasets to evaluate the fidelity and relevance of generated samples.
We also compare our model to the state-of-the-art and observe a substantial improvement in quality, size, and computation time. 
Code, demo, and samples: \href{https://v-iashin.github.io/SpecVQGAN}{\color{blue} \texttt{\textbf{v-iashin.github.io/SpecVQGAN}}}

\end{abstract}
\vspace{-4ex}
\section{Introduction}
\vspace{-1ex}
\label{sec:intro}

A user-controlled sound generation has many applications for \eg movie and music production. 
Currently, foley designers are required to search through large databases of sound effects to find a suitable sound for a scene.
\pagebreak A less painstaking approach would be to automatically generate a novel and relevant sound, given a few visual cues.
Recent advances in deep learning brought to light many promising models for user-controlled content synthesis.

Previous works have proposed models to controllably generate \eg  images \cite{odena2017conditional,miyato2018cgans,zhang2019self,reed2016generative,zhang2017stackgan,xu2018attngan,ramesh2021zero,patashnik2021styleclip,choi2018stargan,lee2020maskgan,esser2021taming}, videos \cite{chen2019mocycle,mallya2020world,tulyakov2018mocogan,lee2018stochastic,narasimhan2021strumming,tripathy2021facegan,hao2018controllable,tulyakov2018mocogan,wang2018video,chan2019everybody}, and audios \cite{chen2017deep,hao2018cmcgan,tan2020spectrogram,dhariwal2020jukebox,bazin2021spectrogram,greshler2021catch,tomczak2020drum,nistal2020drumgan}, or separate sounds \cite{zhao2018sound,zhao2019sound,gan2020music,zhu2020visually,gan2020foley}.
However, most of the audio works are music-related, and only a few attempts have been made to generate visually guided audio in an open domain setup \cite{zhou2018visual,chen2020generating}.
These methods rely on a one-model-per-class approach, which can be prohibitively expensive to scale to hundreds of classes.

Our goal in this paper is to build a single model that is capable of generating sounds conditioned on visual input from multiple classes with a restricted time budget.
To address this, we propose to learn a prior in a form of the Vector Quantized Variational Autoencoder (VQVAE) codebook \cite{oord2017neural} and operate on spectrograms for efficiency.
To shrink the sampling space more aggressively, we draw on advances in controlled image generation \cite{esser2021taming} relying~on a variant of VQVAE with adversarial loss and introduce a novel spectrogram perceptual loss. 

Such an approach allows us to reliably reconstruct a high-fidelity spectrogram from a smaller representation resolution.
We, thus, can train a transformer on a shorter sequence to sample from the codebook and autoregressively construct a high-fidelity spectrogram while being conditioned on the visual cues. 
Finally, we vocode the spectrogram into a waveform using a variant of MelGAN \cite{kumar2019melgan} suitable for open-domain applications.

Human evaluation of content generation models is an expensive and tedious procedure.
In the image generation field, this problem is bypassed with the automatic evaluation of fidelity using a family of metrics based on an ImageNet-pretrained \cite{deng2009imagenet} Inception model \cite{szegedy2016rethinking} \eg  Inception Score \cite{salimans2016improved}, Fr\'{e}chet- \cite{heusel2017gans} and Kernel Inception Distance \cite{binkowski2018demystifying} (FID \& KID). 
The automatic evaluation of a sound generation model, however, remains an open question. 

FID was adapted to assess fidelity of the generated audio in \cite{kilgour2018fr}. This metric is designed for 
very short sounds (<1 second) and, therefore, has limited applicability for long audio as it may miss long-term cues.
Another challenge in the visually guided sound generation is to reliably estimate the relevance of produced samples.
To mitigate both problems, we propose a family of metrics for fidelity and relevance evaluation based on a novel architecture called Melception, trained as a classifier on VGGSound \cite{chen2020vggsound}, a large-scale open-domain dataset. 

The main contributions of this work are: 
\textbf{(1)} a novel efficient approach for multi-class visually guided sound synthesis that relies on a transformer trained to sample from a codebook-based prior; 
\textbf{(2)} a new perceptual loss for spectrogram synthesis, called LPAPS. The loss relies on a novel general-purpose sound classifier, referred to as VGGish-ish, and helps VQVAE to learn reconstruction of higher-fidelity spectrograms from small-scale representations;
\textbf{(3)}~a~novel set of metrics suitable for automatic evaluation of the fidelity and relevance of spectrogram synthesis, called Melception-based FID and MKL. 
We show the effectiveness of our approach in comparison with prior work and provide an extensive ablation study on small- and large-scale datasets (VAS and VGGSound) for visually guided sound synthesis.

\vspace{-2ex}
\section{Related Work}
\vspace{-1ex}

\paragraph{Codebook-based Content Generation}
The use of condensed prior information in a form of a codebook has been shown to effectively reduce the sampling space of generative algorithms.
The initial idea was proposed in the seminal work \cite{oord2017neural} (VQVAE) and further improved in \cite{razavi2019generating} (VQVAE-2).
Applications of VQVAE for content generation include images \cite{oord2017neural,razavi2019generating}, audio \cite{oord2017neural,dhariwal2020jukebox,zhao2020improved,liu2021conditional}, and videos \cite{rakhimov2020latent,yan2021videogpt}. Recently, it was found to be beneficial to train a transformer to sample from the codebook given a rich condition \eg text \cite{ramesh2021zero,ding2021cogview}, low-resolution image, semantic, edge, and depth-maps \cite{esser2021taming}. 
Our method, in contrast, is conditioned on a sequence of video frames and generates spectrograms.

\vspace{-1em}
\paragraph{Automatic Evaluation of Audio Synthesis}
While still being an open research question, few promising ideas have been proposed for the automatic evaluation of audio synthesis. 
Specifically, Kilgour \etal \cite{kilgour2018fr} adapted FID \cite{heusel2017gans} to evaluate the fidelity of music enhancement algorithms.
Unfortunately, the proposed method operates on 1-second windows and, therefore, does not utilize long-term cues.
A similar approach was shown on a text-to-speech task in \cite{binkowski2019high}. 
Alternatively, a model trained on human judgments has been employed as a perceptual loss during training \cite{manocha2020differentiable}.
However, collecting training material for a large-scale dataset poses significant budget requirements.
In this paper, we propose a set of metrics designed to measure both the fidelity and relevance of prolonged open-domain spectrograms.

\vspace{-1em}
\paragraph{Instrument Music Generation With Visual Cues}
Generating short music audios became a testbed for many cross-modal generation algorithms.
Owens \etal \cite{owens2016visually} pioneered the task by collecting a dataset of short videos containing hitting/scratching drumsticks against objects and used a combination of AlexNet \cite{krizhevsky2012imagenet} and LSTM \cite{hochreiter1997long} as a baseline. 
Chen \etal \cite{chen2017deep} focused on the generation of an image from the audio and vice-versa for single-instrument performance videos from the URMP dataset \cite{li2018creating} using two Generative Adversarial Nets (GAN) \cite{goodfellow2014generative} while Hao~\etal~\cite{hao2018cmcgan} improved the performance of the GAN with cross-modal cycle-consistency \cite{zhou2016learning}.
Furthermore, Tan \etal \cite{tan2020spectrogram} incorporated self-attention \cite{vaswani2017attention} into the GAN architecture and Su~\etal~\cite{NEURIPS2020_227f6afd} proposed to generate a piano sound by vocoding Midi predicted from a video.
Recently, Kurmi \etal \cite{kurmi2021collaborative} brought a generation of short (1s) musical videos into the picture.
These methods, however, focus on short ($\sim$1 second) music videos recorded in a controlled setting while our model operates on open-domain 10-second videos.

\vspace{-1em}
\paragraph{Open-domain Audio Generation Based on Visual Cues}
The generation of audio given a set of open-domain visual cues is a novel and challenging task. 
The first attempt to solve the task was published by Chen \etal \cite{chen2018visually} who proposed to employ a subset of AudioSet \cite{gemmeke2017audio} to train a model to learn a residual to an average spectrogram for a video class. 
However, more relevant and higher-fidelity results were obtained by training a separate model for each video class. 
Namely, Zhou \etal \cite{zhou2018visual} trained a separate SampleRNN \cite{mehri2016samplernn} to generate a waveform for each of the 10 classes in the proposed dataset (VEGAS).
Current state-of-the-art results in the generation of relevant and high-fidelity sounds for a video were shown by Chen \etal \cite{chen2020generating} (RegNet).
They noticed the negative impact of ``unseen'' background sound on training dynamics and introduced a ground-truth-based regularizer and an enhanced version of the VEGAS dataset (VAS). 
While producing the most appealing results, the models are trained for each data class and the sampling speed is slow limiting the applicability of the model. 
In this paper, we propose a model that is capable of generating visually relevant sounds from videos of multiple classes in a time that is less than it takes to play the sound.
\vspace{-1em}


\begin{figure*}[h]
\begin{center}
\vspace{-1ex}
\includegraphics[width=\linewidth]{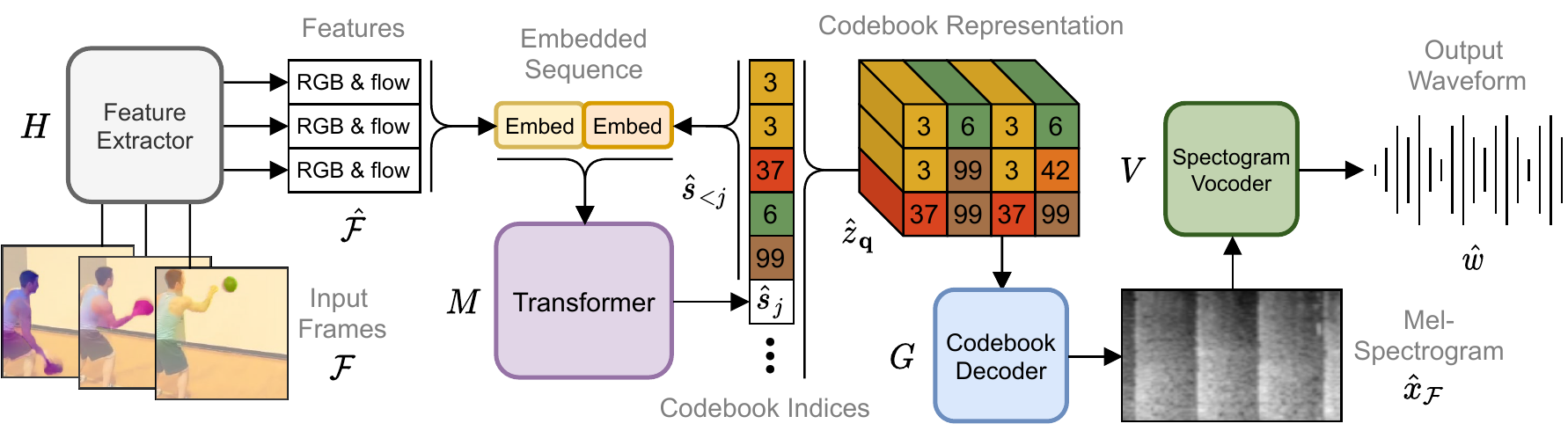}
\end{center}
\vspace{-2em}
   \caption{\normalsize\textbf{Vision-based Conditional Cross-modal Autoregressive Sampler}. A transformer autoregressively samples the next codebook index given a sequence of visual features along with previously generated codebook indices. Once sampling is done, a sequence of generated indices is used to look up a pretrained codebook. Next, a pretrained codebook decoder is used to decode a spectrogram from a codebook representation. Finally, the generated spectrogram is turned into a waveform using a pretrained general-purpose spectrogram vocoder.}
\label{fig:second_stage}
\end{figure*}

\section{Framework}
\vspace{-1ex}
We aim to generate visually relevant and high-fidelity sounds. 
The main challenge is to design a model that handles videos of multiple categories and operates in real-time. 
Thus, we train a transformer to autoregressively compose a concise codebook representation of a spectrogram primed with a small set of frame-wise features obtained from a video (Sec.~\ref{sec:second_stage}).
The representation is then used in the pretrained codebook decoder to produce a spectrogram as outlined in Sec.~\ref{sec:first_stage}.
Finally, a waveform is reconstructed from the spectrogram using a pretrained vocoder as defined in Sec.~\ref{sec:vocoder}.
An overview of the architecture is shown in Fig.~\ref{fig:second_stage}.

\subsection{Perceptually-rich Spectrogram Codebook} \label{sec:first_stage}

The transformer requires the input to be represented as a sequence.
A direct operation on wave samples or raw spectrogram pixels, however, quickly becomes intractable due to the quadratic nature of the dot-product attention. 
Alternatively, one could apply an encoder such as VQVAE \cite{oord2017neural} but the quantized bottleneck representation would be still infeasibly large. 
Our approach draws on VQGAN \cite{esser2021taming}, an efficient autoencoder that allows decoding an image from a smaller-size representation than of VQVAE.
To bridge the gap between image and audio signals, we operate on spectrograms and propose a new perceptual loss (LPAPS).

\paragraph{Spectrogram VQVAE}
Vector-Quantized Variational Autoencoder (VQVAE) \cite{oord2017neural} is trained to approximate an input using a compressed intermediate representation, retrieved from a discrete codebook.
Our adaption of VQVAE, \textit{Spectrogram VQVAE}, inputs a spectrogram $x\inR^{F\times T}$ and outputs a reconstructed version of it $\hat{x}\inR^{F\times T}$.
First, the input $x$ is encoded into a small-scale representation $\hat{z} = E(x) \inR^{F' \times T' \times n_z}$ where $n_z$ is the dimension of the codebook entries and $F'\times T'$ is a reduced frequency and time dimension.
Next, the elements of the encoded representation $\hat{z}$ are mapped onto the closest items in a codebook $\mathcal{Z}=\{z_k\}^K_{k=1} \subset \R^{n_z}$, forming a quantized representation $z_\mathbf{q}\inR^{F'\times T' \times n_z}$:
\begin{align} \label{eq:lookup}
    z_\mathbf{q} = \mathbf{q}(\hat{z}) := \bigg(\underset{z_k\in \mathcal{Z}}{\argmin} ||\hat{z}_{ft} - z_k||  \quad \text{for all $(f\!\comma t)$ in $(F'\times T')$}\bigg).
\end{align}
Since \eqref{eq:lookup} is non-differentiable, we approximate the gradient by a straight-though estimator \cite{bengio2013estimating}.
The reconstructed spectrogram $\hat{x}$ is subsequently decoded from the codebook representation as $\hat{x} = G(z_\mathbf{q}) = G(\mathbf{q}(E(x)))$. 
The full VQVAE objective is defined by
\begin{align}\label{eq:vqvae_loss}
    \mathcal{L}_\text{VQVAE} = \underbrace{\big|\big|x-\hat{x}\big|\big|}_\text{recons loss} + \underbrace{\big|\big|E(x) - \text{sg}[z_\mathbf{q}]\big|\big|^2_2 + \beta \big|\big|\text{sg}[E(x)] - z_\mathbf{q}\big|\big|^2_2}_\text{codebook loss}
\end{align}
where $\text{sg}$ is the stop-gradient operation that acts as an identity during the forward pass but has zero gradient at the backward pass.

The resolution of the intermediate codebook representation ($F'\times T'$) produced by VQVAE remains to be too large for a transformer to operate on. However, more suitable down-sampling rates, \eg $1/16$ of the input size, lead to poor reconstructions as shown in \cite{esser2021taming}.

\begin{figure*}
\begin{center}
\vspace{-1ex}
\includegraphics[width=\linewidth]{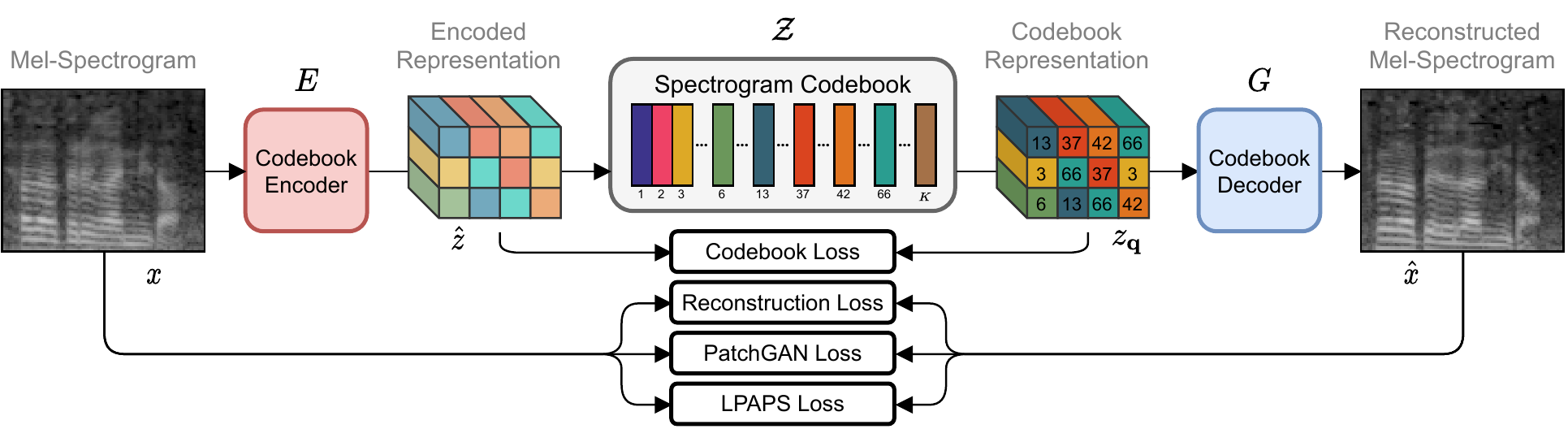}
\end{center}
\vspace{-1.9em}
   \caption{\normalsize\textbf{Training Perceptually-Rich Spectrogram Codebook}. A spectrogram is passed through a 2D codebook encoder that effectively shrinks the spectrogram. Next, each element of a small-scale encoded representation is mapped to its closest neighbor from the codebook. A 2D codebook decoder is then used to reconstruct the input spectrogram. The training of the model is guided by codebook, reconstruction, adversarial, and LPAPS losses.}
\vspace{-2ex}
\label{fig:first_stage}
\end{figure*}

\paragraph{Spectrogram VQGAN and LPAPS}
VQGAN \cite{esser2021taming} is a version of VQVAE, extended with a patch-based adversarial loss \cite{isola2017image} and perceptual loss (LPIPS) \cite{zhang2018unreasonable}, that help to preserve the reconstruction quality when upsampled from a smaller-scale representation.
Since the perceptual loss, used in the original VQGAN, relies on the ImageNet \cite{deng2009imagenet} pretrained VGG-16 \cite{simonyan2014very}, it is unreasonable to expect decent performance on sound spectrograms.
Therefore, we introduce a novel way of guiding spectrogram-based audio synthesis, referred to as Learned Perceptual \textit{Audio} Patch Similarity (LP\textit{A}PS). 

The closest relative of VGG-16 in audio classification is VGGish \cite{hershey2017cnn}, which has the same capacity as VGG-9.
However, we cannot directly build LPAPS on the pretrained VGGish or its architecture, since VGGish digests spectrograms with a rather short time span (<1 second), while our application requires operating on spectrograms spanning up to 10 seconds. 
Moreover, the lack of depth and, therefore, downsampling operations prevents the model from extracting larger-scale features that could be useful in separating real and fake spectrograms.
To address this, we train a variant of the VGG-16 architecture on the VGGSound dataset \cite{chen2020vggsound}.
We refer to the obtained model as VGGish-ish.

Fig.~\ref{fig:first_stage} shows the training procedure for Spectrogram VQGAN with the final loss:
\begin{align}\label{eq:specvqgan_loss}
    \mathcal{L}_\text{SpecVQGAN} = \mathcal{L}_\text{VQVAE} + \underbrace{\log D(x) + \log (1-D(\hat{x}))}_\text{patch-based adversarial loss} + \underbrace{\sum_{s}\frac{1}{F^sT^s}||x^s - \hat{x}^s ||^2_2}_\text{LPAPS loss},
\end{align}
where $D$ is a patch-based discriminator and $x^s\!\comma \hat{x}^s \inR^{F^s \times T^s \times C^s}$ are features from real and fake spectrograms extracted at the $s^\text{th}$ scale of VGGish-ish. 

\subsection{Vision-based Conditional Cross-modal Autoregressive Sampler} \label{sec:second_stage}

The sampler (transformer) is trained to sample a sequence of the codebook indices given a set of visual features.
These should match the indices formed by the codebook encoder for the original audio. 
The conditional prediction of the next token can be formulated as a machine translation task and modeled by the vanilla Encoder-Decoder transformer architecture \cite{vaswani2017attention}.
Alternatively, the problem can be defined in terms of language modeling, that is often approached with a Decoder-only transformer such as GPT \cite{radford2018improving}.
In this paper, we employ a variant of GPT-2 \cite{radford2019language} inspired by its success in autoregressive image synthesis \cite{chen2020generative,esser2021taming}.

As outlined in Fig.~\ref{fig:second_stage}, the sampling starts with the extraction of a sequence of features $\hat{\mathcal{F}} = \{\hat{f}_i\}_{i=1}^N \subset \R^{D_r + D_o}$ formed from a stack of RGB and optical flow frames $\mathcal{F}=\{f_i^r\!\comma f_i^o\}_{i=1}^N$.
The sequence of features $\hat{\mathcal{F}}$ is obtained by applying a frame-wise feature extractor $H$ that consists of two pretrained models (for RGB and flow modalities) such that $\hat{\mathcal{F}} = H(\mathcal{F})$.
Given a sequence of previously generated codebook indices $\hat{s}_{<j}=(\hat{s}_1\!\comma \hat{s}_2\!\comma \dots\comma \hat{s}_{j-1})$ along with the features $\hat{\mathcal{F}}$, an autoregressive step for the transformer $M$ is defined by 
\begin{align}\label{eq:autoregressive-step}
    p\big(s_j|\hat{s}_{<j},\hat{\mathcal{F}}\big) = M\big([\hat{\mathcal{F}}:\hat{s}_{<j}]\big),
\end{align}
where $[:]$ is a stacking operation and $ p\big(s_j|\hat{s}_{<j},\hat{\mathcal{F}}\big) \in [0, 1]^{n_z}$ is a probability distribution over all codebook indices. 
The next codebook index $\hat{s}_j$ is sampled from the multinomial distribution with weights provided by $p$.
The sampling is initialized at $j=1$ and primed only with the input features $\hat{\mathcal{F}}$.
Once $j=F'\cdot T'$, the sampling stops. 
The sequence of predicted codebook indices $\hat{\mathcal{S}} = \{\hat{s}_j\}_{j=1}^{F'\cdot T'}$ is used to lookup the codebook $\mathcal{Z}$ so that, after unflattening, the codebook representation $\hat{z}_\mathbf{q}\inR^{F' \times T' \times n_z}$ is formed. 
The transformer is trained with a typical cross-entropy loss, comparing the predicted codebook indices to those obtained from the ground truth spectrogram.
Finally, given the codebook representation $\hat{z}_\mathbf{q}$, we decode a spectrogram $\hat{x}_\mathcal{F}$ using the decoder $G$ pretrained during the codebook training stage (Sec.~\ref{sec:first_stage}).

We note the importance of unflattening the sequence into a 2D form in a column-major way, precisely as shown in the middle part of Fig.~\ref{fig:second_stage}, opposed to the row-major approach used for image synthesis \cite{chen2020generative,esser2021taming}. 
Employing the row-major unflatteting during training restricts model applications as it would correspond to reconstructing the lower frequencies given the higher ones.
Specifically, we found that a model trained this way produces poor samples when prompted with a few seconds of real audio.

\vspace{-1ex}
\subsection{Spectrogram Vocoder} \label{sec:vocoder}
During the final stage, a waveform $\hat{w}$ is reconstructed from the decoded spectrogram using the pretrained vocoder $V$.
Natural candidates for such vocoding are the Griffin-Lim algorithm \cite{griffin1984signal} and WaveNet (used in prior work \cite{chen2020generating}).
The Griffin-Lim procedure is fast, easy to implement, and it handles the diversity of an open-domain dataset.
However, it produces low-fidelity results when operating on mel-spectrograms. 
In contrast, WaveNet provides high-quality results but remains to be relatively slow on test-time (25 mins per 10-sec sample on a \textbf{G}PU).
For these reasons, we employ MelGAN \cite{kumar2019melgan} that is a non-autoregressive approach to reconstruct a waveform and, therefore, takes only 2 secs per sample on a \textbf{C}PU, while still achieving decent quality. 
Since MelGAN is originally trained for speech or music data, the pretrained models cannot be used in our open-domain scenario.
Therefore, we train a MelGAN on the open-domain dataset (VGGSound).

\vspace{-1ex}
\subsection{Automatic Quality Assessment of Spectrogram-based Synthesis}\label{sec:automatic_eval}

\paragraph{Fidelity}
Our goal is to automatically evaluate both the fidelity and relevance of the generated samples.
In the image generation domain, ImageNet pretrained InceptionV3 \cite{szegedy2016rethinking} is often used to form an opinion on the fidelity of the generated samples. 
Specifically, Inception Score \cite{salimans2016improved} hypotheses low entropy in conditional label distribution and high entropy on a marginal probability distribution for high-fidelity and diverse samples.
More consistent evaluation results were achieved by computing Fr\'{e}chet Distance between the distributions of pre-classification layer's features of InceptionV3 between fake and real samples (FID) \cite{heusel2017gans}.
Considering the domain gap between spectrograms and RGB images, we adapt the Inception architecture for a spectrogram input size and train the model on the VGGSound~dataset.

\begin{table}
\begin{minipage}[c]{0.495\textwidth}
\centering 
\small
\vspace{-1.2ex}
\setlength\tabcolsep{0.42em}
\begin{tabular}{ll rrr}
\toprule
Trained on & Evaluated on & FID↓ & $\overline{\text{MKL}}$↓ \\
\midrule
\cellcolor{vggsound2vggsound!50} VGGSound & \cellcolor{vggsound2vggsound!50} VGGSound & 1.0 & 0.8 \\
\cellcolor{vggsound2vas!50} VGGSound &\cellcolor{vggsound2vas!50} VAS & 3.2 & 0.7 \\
\cellcolor{vas2vas!50} VAS & \cellcolor{vas2vas!50} VAS & 6.0 & 1.0 \\
\bottomrule
\end{tabular} 

\end{minipage}\hfill
\begin{minipage}[c]{0.24\textwidth}
\includegraphics[width=\linewidth]{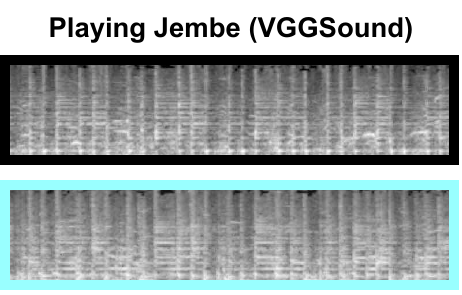}
\end{minipage}\hfill
\begin{minipage}[c]{0.24\textwidth}
\includegraphics[width=\linewidth]{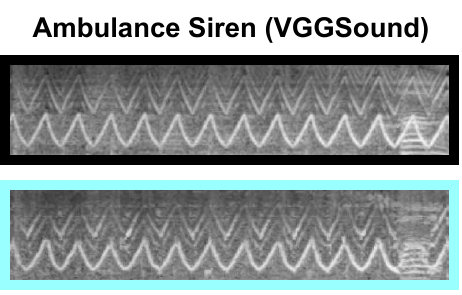}
\end{minipage}
\begin{minipage}[c]{0.24\textwidth}
\includegraphics[width=\linewidth]{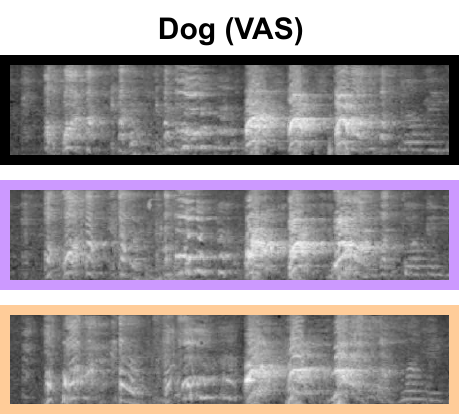}
\end{minipage}\hfill
\begin{minipage}[c]{0.24\textwidth}
\includegraphics[width=\linewidth]{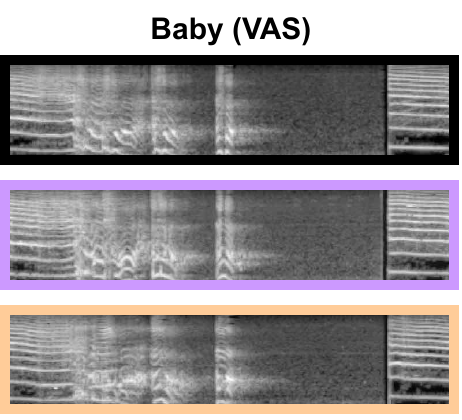}
\end{minipage}\hfill
\begin{minipage}[c]{0.24\textwidth}
\includegraphics[width=\linewidth]{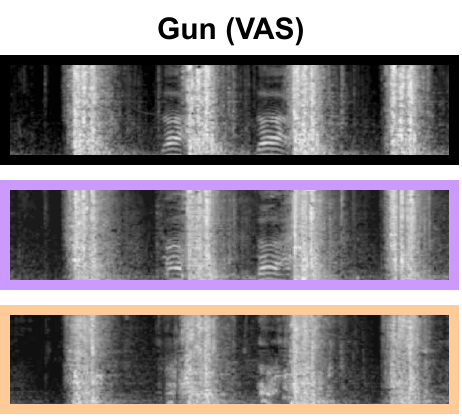}
\end{minipage}\hfill
\begin{minipage}[c]{0.24\textwidth}
\includegraphics[width=\linewidth]{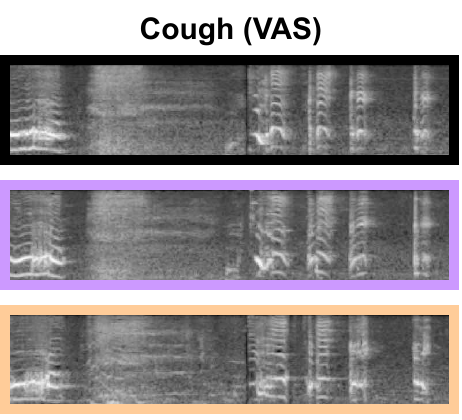}
\end{minipage}
\vspace{0.5ex}
\caption{\normalsize\textbf{Spectrogram VQGAN shows strong reconstruction ability on hold-out sets of VGGSound and VAS}. Metrics are Melception-based FID and mean MKL. On the top-right: ground truth reconstruction results for two classes are shown for a model trained on VGGSound. The bottom triplets show a comparison of VGGSound-trained and VAS-trained models on four classes from VAS. Adobe Reader can be used to listen for reconstructions.}\label{tab:reconstruction}
\vspace{-5ex}
\PlaceText{88.58mm}{211.11mm}{
\includemedia[ addresource=mp3/djembe_gt.mp3, flashvars={ source=mp3/djembe_gt.mp3 &autoPlay=false } ]{
\transparent{0.8} \color{green} \tiny \parbox{\dimexpr 0.25\linewidth-2\fboxsep-2\fboxrule\relax}{\faPlayCircleO~\textsf{Play Me\\[1.2em]}}}{APlayer.swf}
}
\PlaceText{88.58mm}{202.61mm}{
\includemedia[ addresource=mp3/djembe_vggsound.mp3, flashvars={ source=mp3/djembe_vggsound.mp3 &autoPlay=false } ]{
\transparent{0.8} \color{green} \tiny \parbox{\dimexpr 0.25\linewidth-2\fboxsep-2\fboxrule\relax}{\faPlayCircleO~\textsf{Play Me\\[1.2em]}}}{APlayer.swf}
}
\PlaceText{121.08mm}{211.11mm}{
\includemedia[ addresource=mp3/siren_gt.mp3, flashvars={ source=mp3/siren_gt.mp3 &autoPlay=false } ]{
\transparent{0.8} \color{green} \tiny \parbox{\dimexpr 0.25\linewidth-2\fboxsep-2\fboxrule\relax}{\faPlayCircleO~\textsf{Play Me\\[1.2em]}}}{APlayer.swf}
}
\PlaceText{121.08mm}{202.61mm}{
\includemedia[ addresource=mp3/siren_vggsound.mp3, flashvars={ source=mp3/siren_vggsound.mp3 &autoPlay=false } ]{
\transparent{0.8} \color{green} \tiny \parbox{\dimexpr 0.25\linewidth-2\fboxsep-2\fboxrule\relax}{\faPlayCircleO~\textsf{Play Me\\[1.2em]}}}{APlayer.swf}
}
\PlaceText{23.08mm}{191.11mm}{
\includemedia[ addresource=mp3/dog_gt.mp3, flashvars={ source=mp3/dog_gt.mp3 &autoPlay=false } ]{
\transparent{0.8} \color{green} \tiny \parbox{\dimexpr 0.25\linewidth-2\fboxsep-2\fboxrule\relax}{\faPlayCircleO~\textsf{Play Me\\[1.2em]}}}{APlayer.swf}
}
\PlaceText{23.08mm}{182.61mm}{
\includemedia[ addresource=mp3/dog_vggsound.mp3, flashvars={ source=mp3/dog_vggsound.mp3 &autoPlay=false } ]{
\transparent{0.8} \color{green} \tiny \parbox{\dimexpr 0.25\linewidth-2\fboxsep-2\fboxrule\relax}{\faPlayCircleO~\textsf{Play Me\\[1.2em]}}}{APlayer.swf}
}
\PlaceText{23.08mm}{174.11mm}{
\includemedia[ addresource=mp3/dog_vas.mp3, flashvars={ source=mp3/dog_vas.mp3 &autoPlay=false } ]{
\transparent{0.8} \color{green} \tiny \parbox{\dimexpr 0.25\linewidth-2\fboxsep-2\fboxrule\relax}{\faPlayCircleO~\textsf{Play Me\\[1.2em]}}}{APlayer.swf}
}
\PlaceText{55.08mm}{191.11mm}{
\includemedia[ addresource=mp3/baby_gt.mp3, flashvars={ source=mp3/baby_gt.mp3 &autoPlay=false } ]{
\transparent{0.8} \color{green} \tiny \parbox{\dimexpr 0.25\linewidth-2\fboxsep-2\fboxrule\relax}{\faPlayCircleO~\textsf{Play Me\\[1.2em]}}}{APlayer.swf}
}
\PlaceText{55.08mm}{182.61mm}{
\includemedia[ addresource=mp3/baby_vggsound.mp3, flashvars={ source=mp3/baby_vggsound.mp3 &autoPlay=false } ]{
\transparent{0.8} \color{green} \tiny \parbox{\dimexpr 0.25\linewidth-2\fboxsep-2\fboxrule\relax}{\faPlayCircleO~\textsf{Play Me\\[1.2em]}}}{APlayer.swf}
}
\PlaceText{55.08mm}{174.11mm}{
\includemedia[ addresource=mp3/baby_vas.mp3, flashvars={ source=mp3/baby_vas.mp3 &autoPlay=false } ]{
\transparent{0.8} \color{green} \tiny \parbox{\dimexpr 0.25\linewidth-2\fboxsep-2\fboxrule\relax}{\faPlayCircleO~\textsf{Play Me\\[1.2em]}}}{APlayer.swf}
}
\PlaceText{87.08mm}{191.11mm}{
\includemedia[ addresource=mp3/gun_gt.mp3, flashvars={ source=mp3/gun_gt.mp3 &autoPlay=false } ]{
\transparent{0.8} \color{green} \tiny \parbox{\dimexpr 0.25\linewidth-2\fboxsep-2\fboxrule\relax}{\faPlayCircleO~\textsf{Play Me\\[1.2em]}}}{APlayer.swf}
}
\PlaceText{87.08mm}{182.61mm}{
\includemedia[ addresource=mp3/gun_vggsound.mp3, flashvars={ source=mp3/gun_vggsound.mp3 &autoPlay=false } ]{
\transparent{0.8} \color{green} \tiny \parbox{\dimexpr 0.25\linewidth-2\fboxsep-2\fboxrule\relax}{\faPlayCircleO~\textsf{Play Me\\[1.2em]}}}{APlayer.swf}
}
\PlaceText{87.08mm}{174.11mm}{
\includemedia[ addresource=mp3/gun_vas.mp3, flashvars={ source=mp3/gun_vas.mp3 &autoPlay=false } ]{
\transparent{0.8} \color{green} \tiny \parbox{\dimexpr 0.25\linewidth-2\fboxsep-2\fboxrule\relax}{\faPlayCircleO~\textsf{Play Me\\[1.2em]}}}{APlayer.swf}
}
\PlaceText{119.58mm}{191.11mm}{
\includemedia[ addresource=mp3/cough_gt.mp3, flashvars={ source=mp3/cough_gt.mp3 &autoPlay=false } ]{
\transparent{0.8} \color{green} \tiny \parbox{\dimexpr 0.25\linewidth-2\fboxsep-2\fboxrule\relax}{\faPlayCircleO~\textsf{Play Me\\[1.2em]}}}{APlayer.swf}
}
\PlaceText{119.58mm}{182.61mm}{
\includemedia[ addresource=mp3/cough_vggsound.mp3, flashvars={ source=mp3/cough_vggsound.mp3 &autoPlay=false } ]{
\transparent{0.8} \color{green} \tiny \parbox{\dimexpr 0.25\linewidth-2\fboxsep-2\fboxrule\relax}{\faPlayCircleO~\textsf{Play Me\\[1.2em]}}}{APlayer.swf}
}
\PlaceText{119.58mm}{174.11mm}{
\includemedia[ addresource=mp3/cough_vas.mp3, flashvars={ source=mp3/cough_vas.mp3 &autoPlay=false } ]{
\transparent{0.8} \color{green} \tiny \parbox{\dimexpr 0.25\linewidth-2\fboxsep-2\fboxrule\relax}{\faPlayCircleO~\textsf{Play Me\\[1.2em]}}}{APlayer.swf}
}
\end{table}

\paragraph{Relevance}
Since Inception Score and FID metrics rely on dataset-level distributions, they are not suitable to assess the conditional content synthesis.
To this end, we propose a metric, called MKL, that individually compares the distances between output distributions of fake and real audio associated with a condition (\eg frames from a video).
As the distance measure, we rely on KL-divergence and use the Melception classifier to build the distributions.


\vspace{-3ex}
\section{Experiments}

\vspace{-1ex}
\paragraph{VGGSound and VAS Datasets}
VAS dataset \cite{chen2020generating} consists of 12.5k $\sim$6.73-second clips for 8 classes: \textit{Dog}, \textit{Fireworks}, \textit{Drum}, \textit{Baby}, \textit{Gun}, \textit{Sneeze}, \textit{Cough}, \textit{Hammer}.
We follow the same train-test splitting procedure as \cite{chen2020generating} for a fair comparison.
VGGSound dataset \cite{chen2020vggsound} consists of $\sim$200k+ 10-second clips from YouTube spanning 309 classes with audio-visual correspondence.
The classes can be grouped as \textit{people}, \textit{sports}, \textit{nature}, \textit{home}, \textit{tools}, \textit{vehicles}, \textit{music}, etc.
VGGSound is substantially larger, but less curated than VAS due to the automatic collecting procedure.
We managed to download $\sim$190k clips from the dataset as some of the videos were removed from YouTube.
Our split is similar to the original with the exception that the train part is further split into train and validation.
The validation split is formed to match the same number of \textit{videos} per class as in the test set.
As a result, we have 156.5k \textit{clips} in the train, 19.1k in the validation, and 14.5k in the test sets.
This splitting strategy is used across all training procedures including Melception, MelGAN, and VGGish-ish.
To the best of our knowledge, we are the first to use the VGGSound dataset for sound synthesis. 

\vspace{-2ex}
\paragraph{Metrics}
The proposed model is evaluated in quantitative and qualitative studies.
In quantitative evaluation, we rely on Melception-based metrics, namely MKL (averaged across the dataset) and FID for relevance and fidelity evaluation (as defined in Sec.~\ref{sec:automatic_eval}).

\vspace{-2ex}
\paragraph{Details}
We extract log mel-spectrograms of size $80\times 848$ and 212 visual features with dimension $D_r=D_o=1024$ from $\sim$9.8-second videos before training.
The codebook encoder and decoder are generic 2D Conv stacks with two extra attention layers before $\hat{z}$ and after $z_\mathbf{q}$.
The downsampling and upsampling operations are parametrized.
The variant of GPT-2 has 24 layers.
Visual features and codebook indices are embedded to match the transformer dimension (1024).
Training requires at least one 12GB GPU. 
See more in the supplementary.

\begin{figure}[h!]
    \centering
    \vspace{-0,5ex}
    \includegraphics[width=\linewidth]{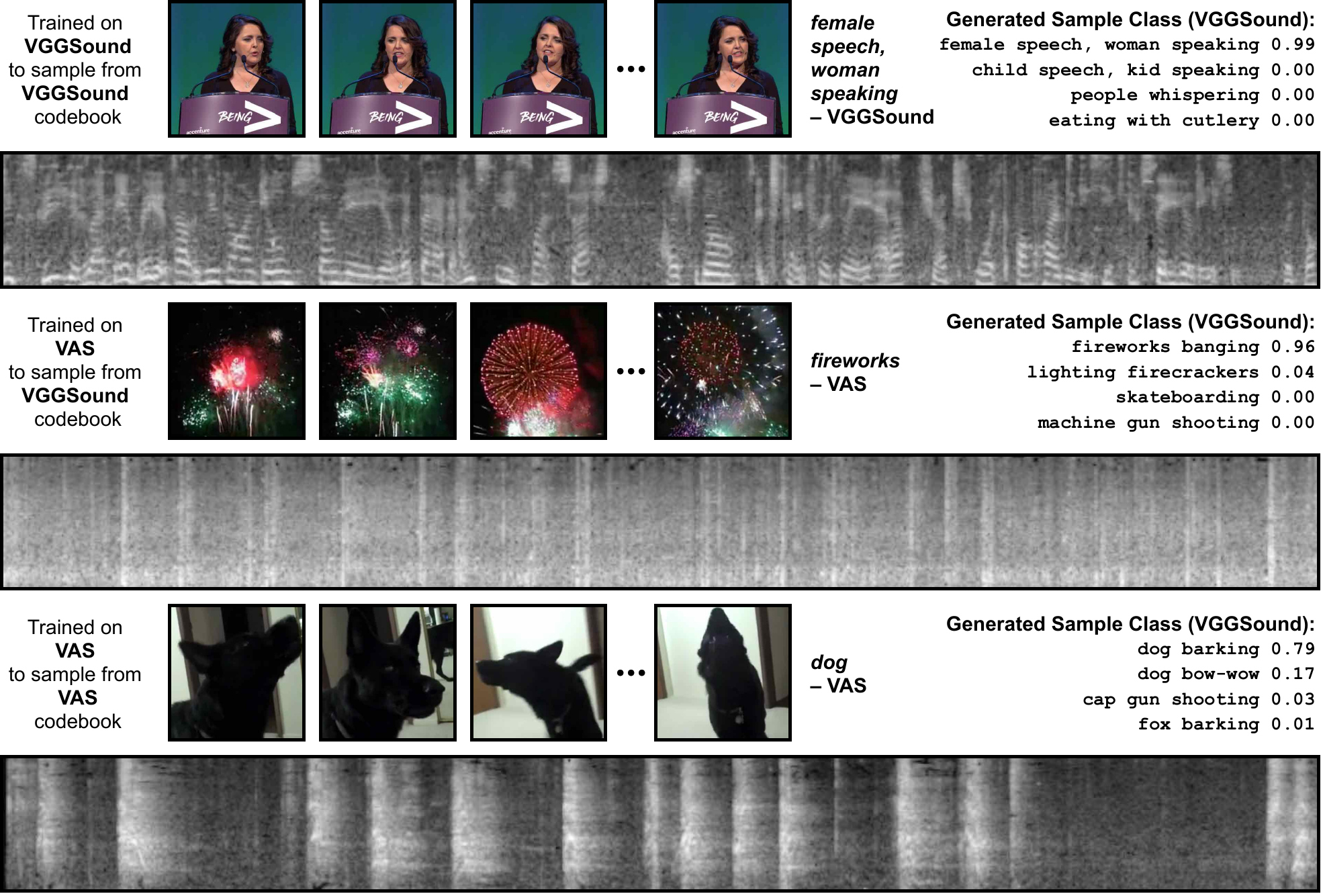}
    \vspace{-4.5ex}
    \caption{\normalsize\textbf{Samples produced by conditional cross-modal sampler are relevant and have high fidelity.} The top row shows results of a model trained on VGGSound to sample from a VGGSound codebook (``from VGGSound for VGGSound''), the middle is ``from VGGSound for VAS'', the bottom is: ``from VAS to VAS''. An ``opinion'' of Melception is on the right.}
    \label{fig:sampling_main}
\PlaceText{21.08mm}{195.11mm}{
\includemedia[ addresource=mp3/female.mp3, flashvars={ source=mp3/female.mp3 &autoPlay=false } ]{
\transparent{0.8} \color{green} \tiny \parbox{\dimexpr 0.98\linewidth-2\fboxsep-2\fboxrule\relax}{\faPlayCircleO~\textsf{Click to Play \\ in Adobe Reader \\[2.8em]}}}{APlayer.swf}
}
\PlaceText{21.08mm}{165.61mm}{
\includemedia[ addresource=mp3/fireworks.mp3, flashvars={ source=mp3/fireworks.mp3 &autoPlay=false } ]{
\transparent{0.8} \color{green} \tiny \parbox{\dimexpr 0.98\linewidth-2\fboxsep-2\fboxrule\relax}{\faPlayCircleO~\textsf{Click to Play \\ in Adobe Reader \\[2.8em]}}}{APlayer.swf}
}
\PlaceText{21.08mm}{136.11mm}{
\includemedia[ addresource=mp3/dog.mp3, flashvars={ source=mp3/dog.mp3 &autoPlay=false } ]{
\transparent{0.8} \color{green} \tiny \parbox{\dimexpr 0.98\linewidth-2\fboxsep-2\fboxrule\relax}{\faPlayCircleO~\textsf{Click to Play \\ in Adobe Reader \\[2.8em]}}}{APlayer.swf}
}
\vspace{-4ex}
\end{figure}

\begin{table}
\begin{minipage}[c]{0.57\textwidth}
    \begin{minipage}[c]{0.46\textwidth}
        \vspace{-2ex}
        \footnotesize
        \setlength\tabcolsep{0.20em}
        \begin{tabular}{cc r r}
        \toprule
        GAN       &    LPAPS  & FID↓  & $\overline{\text{MKL}}$↓ \\
        \midrule
                  &           & 130.4 & 9.6 \\
        \ding{52} &           & 1.4 & 1.1 \\
        \ding{52} & \ding{52} & 1.0 & 0.8 \\
        \bottomrule
        \end{tabular}  
        \vspace{3ex}
    \end{minipage}\hfill
    \begin{minipage}[c]{0.52\textwidth}
        \vspace{-5ex}
        \caption{Adversarial and perceptual losses improve reconstruction results on the VGGSound test set.}\label{tab:ablation_rec}
    \end{minipage}
    \footnotesize
    \setlength\tabcolsep{0.25em}
    \begin{tabular}{rr |rr|rr|rr|r}
\toprule
\multicolumn{2}{r|}{Condition} & FID↓ & $\overline{\text{MKL}}$↓ & FID↓ & $\overline{\text{MKL}}$↓ & FID↓ & $\overline{\text{MKL}}$↓ & \faClockO↓\\
\midrule
& No Feats & 13.5 & 9.7 & 33.7 & 9.6 & 28.7 & 9.2 & 7.7 \\
\midrule
\parbox[t]{2mm}{\multirow{3}{*}{\rotatebox[origin=c]{90}{\textit{ResNet}}}} & 1 Feat\hspace{0.8ex} &  11.5 & 7.3 & 26.5 & 6.7 & 25.1 & 6.3 & 7.7 \\
& 5 Feats   &  11.3 & 7.0 & 22.3 & 6.5 & 20.9 & 6.1 & 7.9 \\
& 212 Feats &  10.5 & 6.9 & 20.8 & 6.2 & 22.6 & 5.8 & 11.8 \\
\midrule
\parbox[t]{2mm}{\multirow{3}{*}{\rotatebox[origin=c]{90}{\scriptsize\textit{Inception}}}} & 1 Feat\hspace{0.8ex} &  8.6 & 7.7 & 38.6 & 7.3 & 25.1 & 6.6 & 7.7 \\
& 5 Feats   &  9.4 & 7.0 & 29.1 & 6.9 & 24.8 & 6.2 & 7.9 \\
& 212 Feats &  9.6 & 6.8 & 20.5 & 6.0 & 25.4 & 5.9 & 11.8 \\
\midrule
\multicolumn{2}{r|}{Codebook} & \multicolumn{2}{c|}{VGGSound} & \multicolumn{2}{c|}{VGGSound} & \multicolumn{2}{c|}{VAS} & \\
\multicolumn{2}{r|}{Sampling for} & \multicolumn{2}{c|}{VGGSound} & \multicolumn{2}{c|}{VAS}      & \multicolumn{2}{c|}{VAS} & \\
\multicolumn{2}{r|}{Setting} & \multicolumn{2}{c|}{\textbf{(a)}} & \multicolumn{2}{c|}{\textbf{(b)}}      & \multicolumn{2}{c|}{\textbf{(c)}} & \\
\bottomrule
\end{tabular}
    \caption{\normalsize\textbf{The number of features is an important factor for relevance and sampling speed on both datasets}. Fidelity and relevance are measured by FID and mean MKL, speed is in seconds to generate a $\sim$10-second audio sample.}\label{tab:main_fid_rel_time}
\end{minipage}\hfill
\begin{minipage}[c]{0.41\textwidth}
    \vspace{-1ex}
\includegraphics[width=\linewidth]{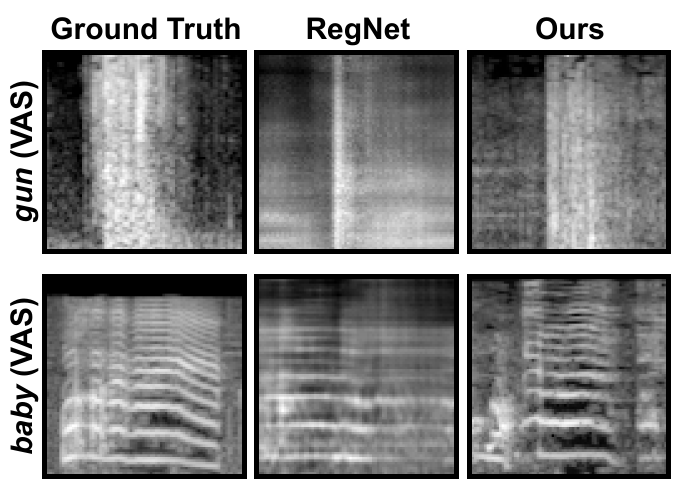}
\vspace{-4ex}
\footnotesize
\setlength\tabcolsep{0.25em}
\begin{tabular}{l r rrr r}
\toprule
& Params & FID↓ & $\overline{\text{MKL}}$↓ &  \faClockO↓ \\
\midrule
Ours \textbf{(b)} & $379\text{M}$ & 20.5 & 6.0 & 12 \\
Ours \textbf{(c)} & $377\text{M}$ & 25.4 & 5.9 & 12 \\
\midrule
RegNet \cite{chen2020generating} & $8 \times 105$M & 78.8 & 5.7 & 1500 \\
Ours \textbf{(b)} + cls & $379\text{M}$ & 20.2 & 5.7 & 12 \\
Ours \textbf{(c)} + cls & $377\text{M}$ & 24.9 & 5.5 & 12 \\
\bottomrule
\end{tabular} 
\caption{\normalsize\textbf{Compared to state-of-the-art, our model generates higher fidelity samples faster and with similar relevance w/ and w/o 
providing the class label}. RegNet size is multiplied by the num. of classes in VAS.}\label{tab:regnet}
\PlaceText{102.08mm}{205.11mm}{
\includemedia[ addresource=mp3/gt_gun.mp3, flashvars={ source=mp3/gt_gun.mp3 &autoPlay=false } ]{
\transparent{0.8} \color{green} \tiny \parbox{\dimexpr 0.3\linewidth-2\fboxsep-2\fboxrule\relax}{\faPlayCircleO~\textsf{Play Me \\[5em]}}}{APlayer.swf}
}
\PlaceText{102.08mm}{187.61mm}{
\includemedia[ addresource=mp3/gt_baby.mp3, flashvars={ source=mp3/gt_baby.mp3 &autoPlay=false } ]{
\transparent{0.8} \color{green} \tiny \parbox{\dimexpr 0.3\linewidth-2\fboxsep-2\fboxrule\relax}{\faPlayCircleO~\textsf{Play Me \\[5em]}}}{APlayer.swf}
}
\PlaceText{118.58mm}{205.11mm}{
\includemedia[ addresource=mp3/regnet_gun.mp3, flashvars={ source=mp3/regnet_gun.mp3 &autoPlay=false } ]{
\transparent{0.8} \color{green} \tiny \parbox{\dimexpr 0.3\linewidth-2\fboxsep-2\fboxrule\relax}{\faPlayCircleO~\textsf{Play Me \\[5em]}}}{APlayer.swf}
}
\PlaceText{118.58mm}{187.61mm}{
\includemedia[ addresource=mp3/regnet_baby.mp3, flashvars={ source=mp3/regnet_baby.mp3 &autoPlay=false } ]{
\transparent{0.8} \color{green} \tiny \parbox{\dimexpr 0.3\linewidth-2\fboxsep-2\fboxrule\relax}{\faPlayCircleO~\textsf{Play Me \\[5em]}}}{APlayer.swf}
}
\PlaceText{135.58mm}{205.11mm}{
\includemedia[ addresource=mp3/ours_gun.mp3, flashvars={ source=mp3/ours_gun.mp3 &autoPlay=false } ]{
\transparent{0.8} \color{green} \tiny \parbox{\dimexpr 0.3\linewidth-2\fboxsep-2\fboxrule\relax}{\faPlayCircleO~\textsf{Play Me \\[5em]}}}{APlayer.swf}
}
\PlaceText{135.58mm}{187.61mm}{
\includemedia[ addresource=mp3/ours_baby.mp3, flashvars={ source=mp3/ours_baby.mp3 &autoPlay=false } ]{
\transparent{0.8} \color{green} \tiny \parbox{\dimexpr 0.3\linewidth-2\fboxsep-2\fboxrule\relax}{\faPlayCircleO~\textsf{Play Me \\[5em]}}}{APlayer.swf}
}

\end{minipage}
\vspace{-5ex}
\end{table}

\subsection{Results} 
\vspace{-1ex}

\paragraph{Reconstruction with Spectrogram VQGAN}\label{sec:results_rec}
When compared to ground truth spectrograms, the reconstructions are expected to have high fidelity (low FID) and to be relevant (low mean MKL).
Tab.~\ref{tab:reconstruction} contains quantitative and qualitative results produced by our Spectrogram VQGAN (Sec.~\ref{sec:first_stage}).
The results imply high fidelity and relevance on a variety of classes from both VGGSound (test) and VAS (validation) datasets.
Notably, the performance of the VGGSound-pretrained codebook is better than of the VAS-pretrained codebook even when applied on the VAS validation set due to larger and more diverse data seen during training.
The implementation details and more examples are provided in the Supplementary.
Moreover, in Tab.~\ref{tab:ablation_rec} we show the results of the ablation study on the impact of losses on reconstruction quality.
In particular, the absence of the adversarial loss results in significant blurriness (which agrees with the findings in \cite{esser2021taming}) in reconstructed spectrograms and expected substantial downgrade in metrics.

\vspace{-3ex}
\paragraph{Visually-Guided Sound Generation}\label{sec:results_sampling}
We benchmark the visually-guided sound generation qualitatively and quantitatively using three different settings: 
\textbf{a)} trained the transformer on \textit{VGGSound} to sample from the \textit{VGGSound} codebook, 
\textbf{b)} trained on \textit{VAS} with the \textit{VGGSound} codebook, and
\textbf{c)} trained on \textit{VAS} with the \textit{VAS} codebook.
Fig.~\ref{fig:sampling_main} shows a few examples obtained with different settings along with the ``opinion'' of the Melception classifier on 
the generated sample label and in Tab.~\ref{tab:main_fid_rel_time}, we compare a different number of priming features 
including sampling without a condition (\textit{No Feats}), which can be seen as the upper-bound on the relevance metric (mean MKL).
The qualitative results are provided for two sets of ImageNet-pretrained features: BN-Inception (RGB + flow) and ResNet-50 (RGB).

We observe that:
1) In general, the more features from a corresponding video are used, the better the result in terms of relevance.
However, there is a trade-off imposed by the sampling speed which decreases with the size of the conditioning.
2) A large gap (log-scale) in mean MKL between visual and ``empty'' conditioning suggests the importance of visual conditioning in producing relevant samples.
3) When the sampler and codebook are trained on the same dataset---settings (a) and (c)---the fidelity remains on a similar level if visual conditioning is used. This suggests that it is easier for the model to learn ``features-codebook'' (visual-audio) correspondence even from just a few features. However, if trained on different datasets (b), the sampler benefits from more visual information.
4) Both BN-Inception and ResNet-50 features achieve comparable performance, with BN-Inception being slightly better on VGGSound and with
longer conditioning in each setting.
Notably, the ResNet-50 features are RGB-only which significantly eases practical applications.
We attribute the small difference between the RGB+flow features and RGB-only features to the fact that ResNet-50 is a stronger architecture
than BN-Inception on the ImageNet benchmark \cite{bianco2018benchmark}.
See the technical details, more examples, ablations, and human studies in Supplementary~Material.

\vspace{-3ex}
\paragraph{Comparison with the state-of-the-art}\label{sec:results-regnet}
In Tab.~\ref{tab:regnet}, we compare our model to RegNet \cite{chen2020generating}, which is currently the strongest baseline in generating relevant sounds for a visual sequence.
For a fair comparison, we employ the same data preprocessing for audio and visual features as in RegNet \cite{chen2020generating}. 
We use the settings (b) \& (c) (see Tab.~\ref{tab:main_fid_rel_time}) with 212 features in the condition, which is similar to the RegNet input.
Since RegNet limits the sampling space explicitly by training a separate model for each class, it is difficult to fairly compare relevance with our model that is trained on all classes.
To mitigate this to some extent, we include a class label into the transformer conditioning sequence allowing the model to learn to separate parameter subspaces for all 8 classes.
The results suggest that our model produces higher quality spectrograms than RegNet, which is also supported by the lower FID scores.
Moreover, RegNet uses two times more parameters. 
See more examples in the Supplementary Material.

\subsection{Qualitative Analysis of the Model Properties}\label{sec:qualitative-analysis}
\vspace{-1ex}
We conduct a human study by single-handedly inspecting over 2k samples for test-set videos of the VGGSound dataset. 
Despite the biasedness of the study, we believe that the results are worth reporting.
The samples are drawn for a random class and using the model trained on the VGGSound dataset with the VGGSound codebook (the setting (a), \textit{5 Feats}, see Sec.~\ref{sec:results_sampling}).  We divide our observations into three parts: \textbf{general properties of the model}, \textbf{problems with data preprocessing}, and \textbf{dataset-related issues} (see supplementary).
\vspace{-4ex}
\paragraph{General Properties of the Model}
The proposed model supports multiple classes and, especially with some patience budget, generates relevant audio for the majority of classes in the VGGSound.
The mistakes are not rare, but they are often associated with a poor audio-visual correspondence in the video or because the model 
generates a sound of another musical instrument instead of the specific one (\eg, \textit{violin} instead of \textit{cello} -- both are string
instruments).
However, the generation of a sample that belongs to a completely different class group is a rare event, 
\eg, for a bird singing video the model will not generate an audio appropriate for indoor sports activities.
We also observed, for classes such as \textit{zebra braying}, \textit{cat purring}, \textit{pig oinking}, \textit{bee, wasp, etc. buzzing}, \textit{cattle mooing}, \textit{alarm clock ringing}, the model struggles to produce a relevant sample possibly due to the unobservable source of the signal (\eg, the flies are flying around the camera pointed to a tree and the flies are never captured but heard).

The model may confuse visually similar sounds, \eg, \textit{people whistling}, \textit{singing}, \textit {talking}, 
\textit{whispering}, \textit{burping}, etc.
Also, if a video shows a close-up of hands, \eg, \textit{machine sewing}, the model may generate a sound of \textit{keyboard typing} or \textit{computer mouse clicking}.
We also found that an ASMR setup (Autonomous Sensory Meridian Response) enforces the model to produce clean sounds similar to ASMR 
but often of a different class. 
The model struggles to differentiate different types of birds (\eg, \textit{swallow} \textit{chickadee},  \textit{pheasant}, etc) 
or hitting instruments (\eg, \textit{bongo}, \textit{timbales}, \textit{timpani}, \textit{steelpan}, etc), yet it tends to produce the sounds 
of a similar class from, \eg, another bird or instrument. 
These properties are expected from a model trained on a relatively noisy dataset with a vague separation between classes.
\vspace{-4ex}
\paragraph{Data Preprocessing Issues}
After transformation into the mel-scale spectrogram, the audio signal loses the phase and a range of essential frequencies
to differentiate sounds from some classes. 
For instance, by transforming the waveform into mel-scale spectrogram and back, we observed that the sound of 
\textit{cat caterwauling} became indiscernible from \textit{person sobbing}, \textit{crying}, or \textit{dog howling} classes. 
Although the speech segments are recognizable, the words are indecipherable.
To this end, the model can be trained directly on top of the STFT spectrograms at the cost of efficiency during sampling, however.
\vspace{-4ex}
\section{Conclusion}
\vspace{-2ex}
We introduced a new efficient approach for multi-class visually-guided sound generation, which operates on spectrograms and relies on a prior in a form of a codebook representation. 
To train the prior, we proposed a new perceptual loss (LPAPS) which is based on a general-purpose classifier (VGGish-ish).
This loss allows the model to learn to reconstruct higher-fidelity spectrograms from a small-scale representation.
In addition, a novel automatic evaluation procedure is outlined to estimate both fidelity and relevance of generated spectrograms with a new family of metrics based on the Melception classifier.
Our experiments on small- and large-scale datasets show the power and efficiency of our model in both quantitative and qualitative studies compared to the state-of-the-art.

\noindent{\small\textbf{Acknowledgments}~\,Funding for this research was provided by the
Academy of Finland projects 327910 \& 324346. We also acknowledge CSC -- IT Center for Science, Finland, for computational~resources.}


\pagebreak

\section{Supplementary Material}

This section is organized as follows. 
In Section~\ref{sec:implementation-details}, we provide comprehensive guidance on training the model including VGGish-ish, 
Melception, and MelGAN.
In Section~\ref{sec:additional-results}, we report additional results on the experiments mentioned earlier.
A qualitative analysis of the model properties is provided in Section~\ref{sec:qualitative-analysis}.
Besides, we include a package with generated audios in \texttt{.mp3} format used in the figures with this supplementary.

\vspace{-1.5ex}
\subsection{Implementation Details}\label{sec:implementation-details}
\vspace{-0.5ex}
In our experiments we select hyper-parameters on validation sets of datasets, the test set of VGGSound was used to calculate the final results.
Considering the broad space for hyper-parameter and architectural elements search we initialize our search from a set of ``reasonable image-synthesis defaults'' which happen to work well for most of the cases.
Despite relying on a relatively expensive hardware setup for our experimentation (4 $\times$ 40GB Nvidia A100 GPUs), it is possible to train all models on one 12GB Nvidia 2080Ti GPU with smaller batch size.
We used the PyTorch deep learning library \cite{NEURIPS2019_9015} in development.
The code and the pre-trained models will be publicly released.

\vspace{-1.5ex}
\subsubsection{Feature Extraction Pipeline}\label{sec:feature-extraction}
\vspace{-0.5ex}
Before training the models, we first pre-extract audio and visual features from videos.
For both VGGSound and VAS datasets, we follow the same feature extraction pipeline as the baseline work (RegNet) \cite{chen2020generating} as it is shown to produce good results, and for a fair comparison with the baseline.
Specifically, the original videos are first resampled to 21.5 fps and trimmed to be at most 10 seconds long, while the shorter clips are repeated until they span 10 seconds.
Then, for every 10-second clip, we extract audio and visual features as follows.

\vspace{-1em}
\paragraph{Audio Features}
The original audio from a video clip is resampled at 22050 Hz and passed through the short-term Fourier transform (STFT) with 1024 bins and 256 hop length. Next, we transform the absolute values of the spectrogram into mel-scale using 80 mel bands in the range from 125 to 7600 Hz and apply logarithm which results in a log-mel-spectrogram ($x$) of size $(F\times T) = (80 \times 860)$.
For our models, we further center crop the spectrogram ($860 \rightarrow 848$) to be able to downscale it by 2 more times without having a non-zero reminder which is useful for Spectrogram VQGAN.
The same audio features are used across the training of all models including Melception, VGGish-ish, and MelGAN (no center cropping).

\vspace{-1em}
\paragraph{Visual Features}
For \textit{BN-Inception} features, we start by extracting a stack of optical flow frames ($\mathcal{F}$) from a set of RGB frames similar to Temporal Segment Networks (TSN) \cite{TSN2016ECCV}.
From a 10 second clip, 215 RGB and flow frames are extracted.
The feature extractor ($H$) relies on ImageNet \cite{krizhevsky2012imagenet} pre-trained weights.
The optical flow uses the same weights except for the input convolutional layer that originally had 3 channels (for RGB).
The weights of this kernel are averaged across the channel dimension and replicated to have 2 channels (for $x$ and $y$ flow maps).
This procedure is inspired by \cite{wang2015towards}.
For each pair of RGB and optical flow frames, $H$ extracts a pair of features of size $D_r=D_o=1024$.
For our models, we also do center cropping to have 212 features to temporally match the cropped audio features i.e. 
4 audio features for every pair of visual features. 
Similarly, \textit{ResNet-50} feature extractor is pre-trained on ImageNet and outputs 215 RGB features, cropped to 212 features. 
In contrast, the features have $D_r=2048$ dimension which matches the dimension of concatenated flow and RGB features from BN-Inception.
Since extraction of optical flow frames is not required, these features are easier to use in practice.
If not specified, BN-Inception features are used.

\begin{figure*}
\begin{center}
\includegraphics[width=\linewidth]{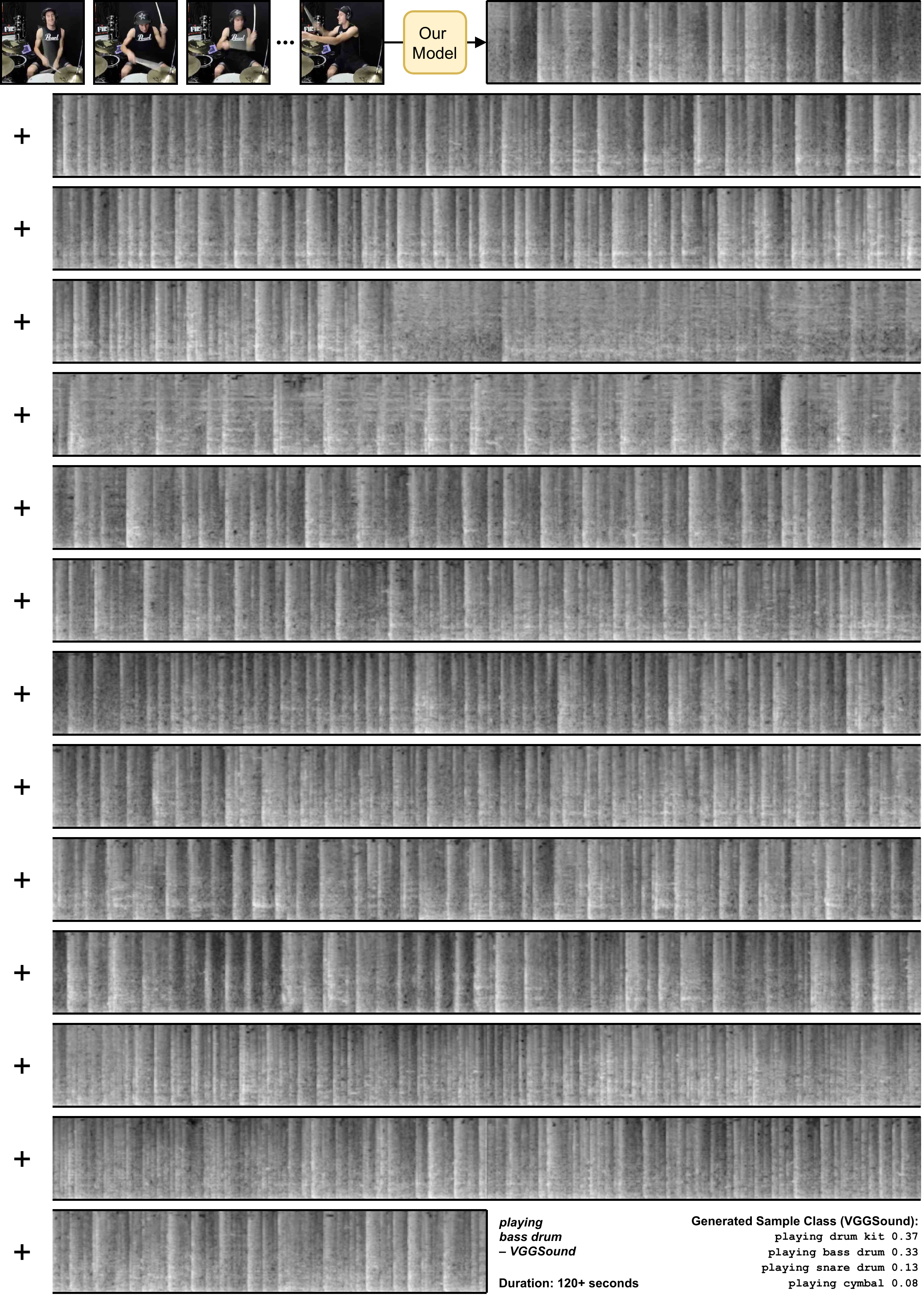}
\PlaceText{88.08mm}{209.11mm}{
\includemedia[ addresource=mp3/drum_solo_audio.mp3, flashvars={ source=mp3/drum_solo_audio.mp3 &autoPlay=false } ]{
\transparent{0.8} \color{green} \tiny \parbox{\dimexpr0.4\linewidth-2\fboxsep-2\fboxrule\relax}{\faPlayCircleO~\textsf{Click to Play \\ in Adobe Reader \\[2.8em]}}}{APlayer.swf}
}
\end{center}
\vspace{-2ex}
\caption{\normalsize ``The great drum solo''. The sample is generated by the model trained on $\sim$10-second spectrograms given 5 RGB and optical flow video frames. Generation time is shorter than it will take to play the sample.}
\label{fig:drum-solo}
\end{figure*}

\begin{table}
\begin{minipage}[c]{0.47\textwidth}
    \centering 
    \small
    \setlength\tabcolsep{0.42em}
    \begin{tabular}{lrrrr}
\toprule
Model & Top-1 & Top-5 & mAP & mAUC \\
\midrule
VGGish-ish & 34.70 & 63.71 & 36.63 & 95.70 \\
Melception & 44.49 & 73.79 & 47.58 & 96.66 \\
\bottomrule
\end{tabular}
\end{minipage}\hfill
\begin{minipage}[c]{0.51\textwidth}
    \centering 
    \small
    \caption{\normalsize Performance of the VGGish-ish and Melception classifiers used for the perceptual loss, LPAPS (see Sec.~\ref{sec:first_stage}), and automatic evaluation (see Sec.~\ref{sec:automatic_eval}). Metrics are Top-1,5 accuracy, mean average precision, and area under curve across all classes. The performance is reported on the test set of VGGSound.}\label{tab:vggishish_melception}
\end{minipage}
\vspace{-3ex}
\end{table}

\subsubsection{Perceptually-Rich Spectrogram Codebook Settings}
The Codebook Encoder ($E$) is a generic 2D-Conv stack with skip-connections \cite{he2016deep} and GroupNorm \cite{wu2018group}, we also additionally add layers of self-attention at the lowest scale which operates on a flattened representation as in \cite{esser2021taming}.
The Codebook Decoder ($G$) has a symmetric architecture to $E$ with an exception for the upsampling layer which doubles the spatial resolution before the conv kernel with the nearest neighbor interpolation. 

To train an efficient spectrogram codebook (defined in Sec.~\ref{sec:first_stage}) for VGGSound and VAS we relied on a dimension of the codebook $n_z=256$.
Thus, for a given mel-spectrogram of size $F\times T = 80 \times 848$ the codebook representation is $ F'\times T' \times n_z = 5\times 53 \times 256$.
The number of codes in VAS and VGGSound is different, though. For VGGSound, we used $|\mathcal{Z}| = 1024$. Since VAS is a magnitude smaller dataset than VGGSound, we went with a smaller number of codes $|\mathcal{Z}| = 128$ otherwise with, for example, $|\mathcal{Z}| = 1024$ the VQVAE exhibited catastrophic index collapse.
Both models are around 75M parameters in size.
Since audio is essentially a 1D signal, we tried a 1D variant of the codebook representation but it did not yield promising reconstruction results.

To stabilize the training procedure, we zero out the adversarial part of the loss in Eq.~\ref{eq:specvqgan_loss} for the first 30k training steps for VGGSound and 2k iterations for VAS, after that the loss is multiplied with a coefficient of 0.8. 
We keep the $\beta$ coefficient to be 0.25 (Eq.~\ref{eq:vqvae_loss}).
The learning rate is fixed and determined as a product of a base learning rate, a number of GPUs, and a batch size. The base learning rate for VGGSound is $4.5\cdot10^{-6}$ and $10^{-6}$ for VAS.
We train the codebook with batches of 8 mel-spectrograms. On VGGSound dataset the training takes about 72 hours and 300k steps (40 epochs) while on VAS it takes 36 hours and 95k steps (260 epochs) with Adam optimizer \cite{kingma2014adam} (\textit{betas} are $(0.5, 0.9)$). 
Despite these estimates on the training time, we note that the model can be stopped much earlier and still produce good reconstruction results.
When making a decision when to stop training, we mostly rely on manual visual inspection of reconstructed samples in the validation set.
We highlight that the VGGSound model did not saturate and kept improving the reconstruction ability.

\subsubsection{Training a model for LPAPS (VGGish-\textit{ish})}\label{sec:vggish-ish}
In order to train a model to produce an efficient small-scale representation of spectrograms, in this paper, we introduce a new perceptual loss LPAPS (as defined in Eq.~\eqref{eq:specvqgan_loss}) which relies on a pretrained classifier model for extraction of perceptually-rich features.
To this end, we proposed a variant of VGGish \cite{hershey2017cnn} that we refer to as VGGish-ish.
We train VGGish-ish on the VGGSound dataset with batch size of 32 mel-spectrograms using Adam optimizer with \textit{betas} $=(0.9, 0.999)$, learning rate $=3\cdot 10^{-4}$ with weight decay $=10^{-3}$. 
The training stops if for 5 consecutive epochs validation loss (cross-entropy) does not improve which, in our case, happened after 4 epochs (2 hours).
The model has 137.6M parameters and was trained on one 12GB Nvidia 2080Ti GPU. 
For performance reference, see Table~\ref{tab:vggishish_melception}.

\subsubsection{Settings for the Vision-based Conditional Cross-modal Autoregressive Sampler}

The autoregressive sampler is a 24-layer 16-head transformer (GPT-2 \cite{radford2019language}) similar to \cite{esser2021taming} used for conditional image synthesis.
Along with the codebook index embedding layer, we transform the visual features into the transformer's hidden dimension space (1024) using one dense layer.
The output of the transformer is passed through a $K$-way softmax classifier which forms a distribution over the next codebook index.
The $K$ is 1024 for VGGSound and 128 for VAS codebooks, corresponding to the number of codes in the codebook. 
The transformer has approximately 310M parameters.

The base learning rate when training the transformer is $5\cdot10^{-6}$ for VGGSound and $10^{-6}$ for VAS datasets. 
We use the batch size of 64 samples. On the VGGSound dataset, the training of the transformer with the longest visual condition takes about 4 hours and 18k steps (6 epochs) while on the VAS dataset it takes 1 hour and 3k steps (16 epochs) with AdamW optimizer \cite{loshchilov2017decoupled}, \textit{betas} are $(0.9, 0.95)$. The model is trained until the loss on the validation set has not improved to 2 consecutive epochs.

When implementing sampling without a condition (the \textit{No Feats} settings in Tab.~\ref{tab:main_fid_rel_time}), we considered the coordinate map as used in VQGAN \cite{esser2021taming} and one randomly distributed vector (uniformly) of the same dimension as visual features.
The implementation of the condition as the coordinate map has a form of integers mimicking codebook tokens increasing monotonically from 0 to $K-1$ along the time dimension of a spectrogram ($F' \times T'$).
This temporal monotonicity gives the transformer an additional ``sense'' of time similar to the positional encoding.
However, we found that using the single random vector yield the same results as the coordinate map while being faster to train and sample due to the shorter attention span because of the reduced conditioning size ($F'\cdot T'$ vs $1$).

\subsubsection{Training a Vocoder (MelGAN)}
We rely on the official implementation of the MelGAN \cite{kumar2019melgan}.
During training, the model inputs a random sequence of 8192 audio samples (@22050 Hz). 
Operation on such short windows ($\sim$1/3 second) and the fully convolutional architecture allows reconstruction of spectrograms of arbitrary temporal dimension on test-time.  
We use the same spectrogram extraction procedure as described in Section~\ref{sec:feature-extraction} except for the center cropping.
The vocoder is trained for 3M iterations with a batch size of 16 mel-spectrograms on one 12GB Nvidia 2080Ti for approximately 14 days (1800 epochs) on the VGGSound dataset.
However, the training did not saturate and kept yielding better results on the test set. 

\subsubsection{Training a Model for Fidelity and Relevance Evaluation (Melception)}
Melception mostly resembles the InceptionV3 architecture \cite{szegedy2016rethinking} except for the input convolutional layer (has 1 channel instead of 3) and absence of max pooling operations to preserve spatial resolution at higher layers.
The model has 25.2M parameters.
The training procedure is similar to VGGish-ish (Sec.~\ref{sec:vggish-ish}). 
However, we use a batch size of 48, no weight decay, and train it on one 40 GB Nvidia A100 for 14 hours, early stopped after 14 epochs.
For performance reference, see Table~\ref{tab:vggishish_melception}.

\subsubsection{Evaluation of Generated Samples}
To evaluate the fidelity and relevance of the samples produced by the model, we generate samples for each video in the hold-out set.
Since mean KL-divergence (MKL) is calculated on each sample individually (see Section~\ref{sec:automatic_eval}), we compare each sample with the corresponding ground truth and average the estimates.
For robustness of the estimates, we generate 10 samples per test video for the VGGSound dataset and 100 samples per validation video for the VAS dataset.

The evaluation, however, requires a significant resource budget since it is necessary to generate 140k samples per VGGSound experiment and 80k for VAS.
For instance, evaluation of the model under setting (a) with 212 visual features in condition (see Table~\ref{tab:main_fid_rel_time}) takes 18 hours on a node with $4\times40$GB Nvidia A100.
We, therefore, parallelize the sampling into several nodes (up to 6).

\begin{figure*}
\begin{center}
\includegraphics[width=\linewidth]{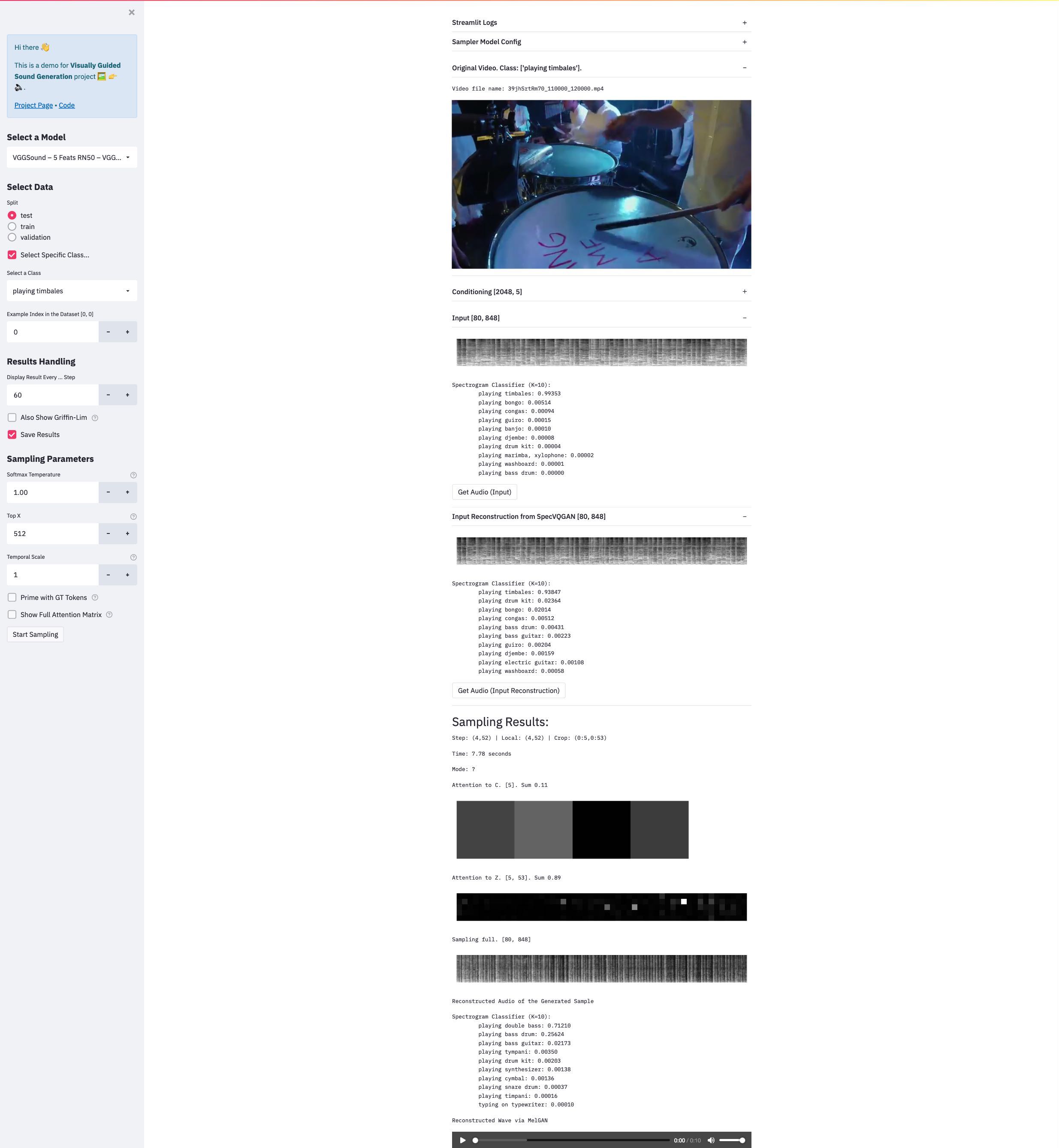}
\end{center}
   \caption{\normalsize The sampling interface that is used to control sampling, and inspect the results of a model. A user can select a model, class, video from a dataset, inspect the reconstruction results, attention activations, and generate a novel audio sample. Best viewed if zoomed-in. The sampling interface is available as a part of our publicly released codebase.}
\label{fig:sampling-interface}
\end{figure*}

\paragraph{Sampling Interface}
Drawing on the sampling interface for conditional image synthesis implemented in \cite{esser2021taming}, we employ \texttt{Streamlit}, an open-source app framework, to build a tool that helps to control the sampling results and qualitatively benchmark the model.
The screenshot of the interface is shown in Figure~\ref{fig:sampling-interface}.

\subsection{Additional Results}\label{sec:additional-results}
\begin{figure*}
\begin{center}
\includegraphics[width=\linewidth]{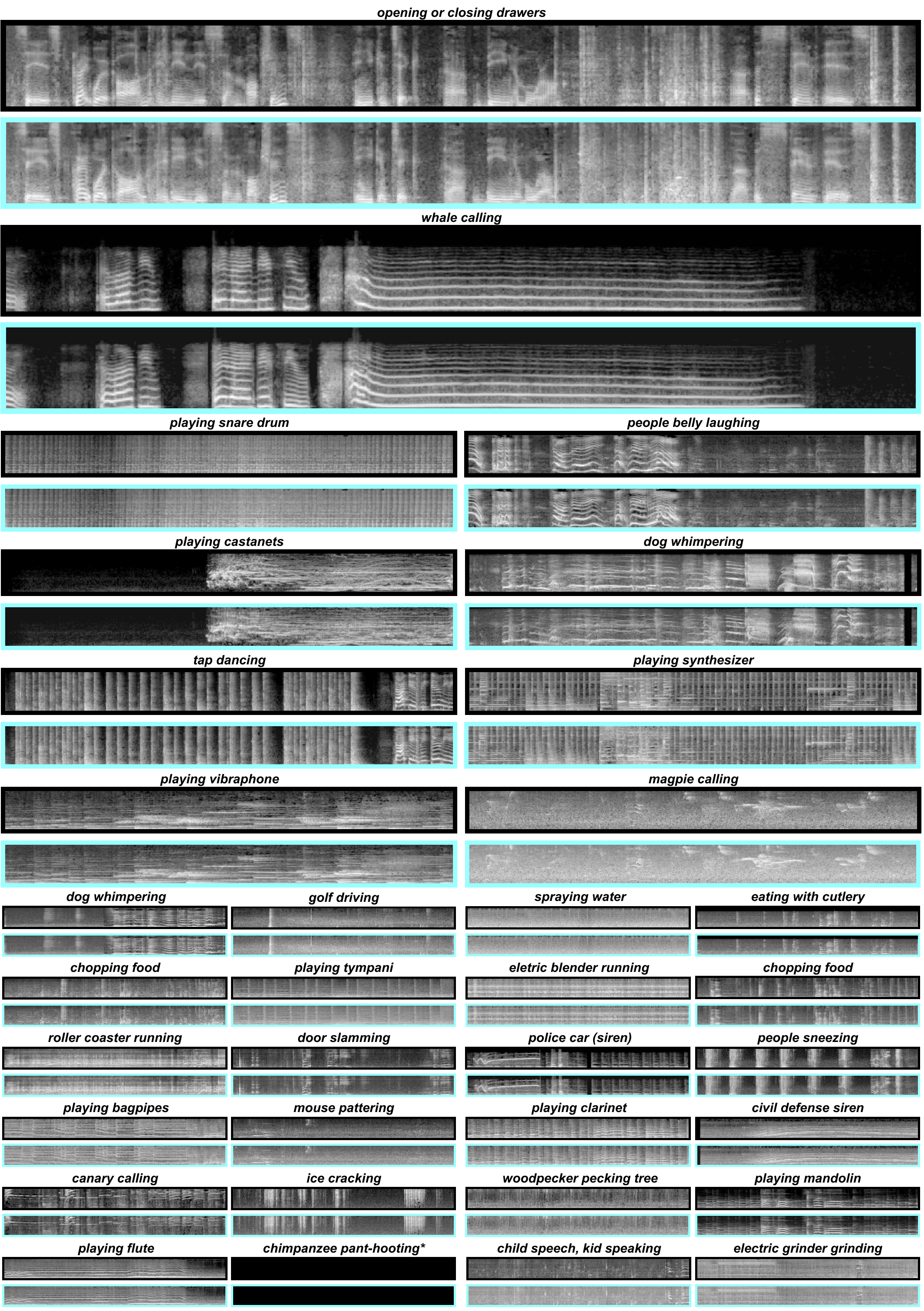}
\end{center}
\vspace{-2ex}
   \caption{\normalsize Random selection of VGGSound test examples ({\color{black} \textbf{top}}) and a Spectrogram VQGAN reconstruction for each ({\color{vggsound2vggsound} \textbf{bottom}}). The model is trained on the VGGSound dataset and shows reconstructions on the test subset of VGGSound. The colors match those in Table~\ref{tab:reconstruction} in Section~\ref{sec:results_rec}. (*)~---~silent audio. The audio samples are provided with this supplementary.}
\label{fig:more-rec-results-vggsound}
\end{figure*}

\begin{figure*}
\begin{center}
\includegraphics[width=\linewidth]{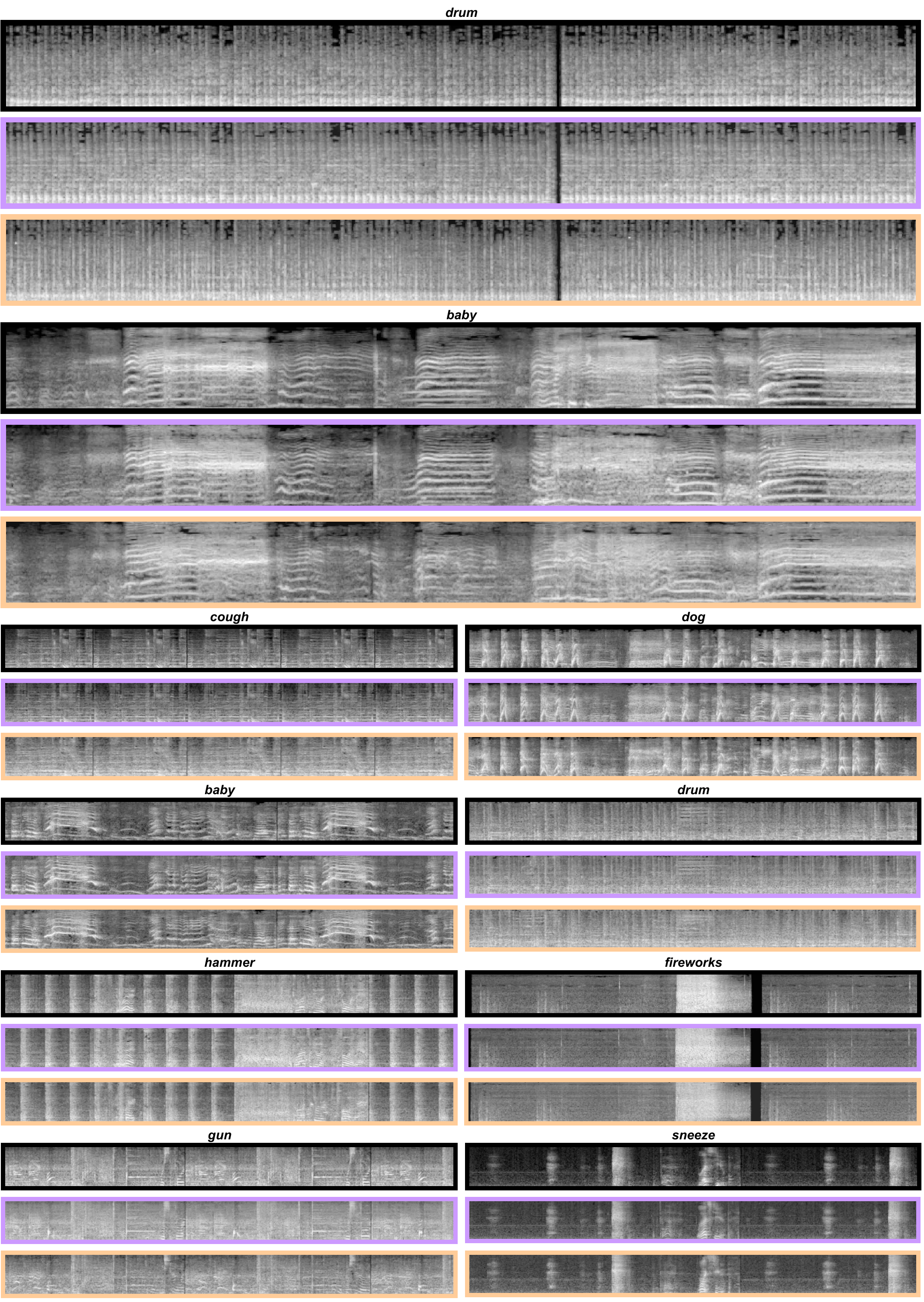}
\end{center}
\vspace{-2ex}
   \caption{\normalsize Random selection of VAS validation examples ({\color{black} \textbf{top}}) as well as reconstruction using Spectrogram VQGAN trained on  VGGSound ({\color{vggsound2vas} \textbf{middle}}) and VAS ({\color{vas2vas} \textbf{bottom}}). The colors match those in Table~\ref{tab:reconstruction} in Section~\ref{sec:results_rec}.  The audio samples are provided with this supplementary.}
\label{fig:more-rec-results-vas}
\end{figure*}

\paragraph{More Spectrogram VQGAN Reconstruction Results}
In Figures~\ref{fig:more-rec-results-vggsound} and \ref{fig:more-rec-results-vas} we show more reconstruction results of the Spectrogram VQGAN (defined in Section~\ref{sec:first_stage}) for randomly drawn samples. 
The results imply strong reconstruction ability of the model on both VAS and VGGSound datasets including the situation when a model is pretrained on one dataset (VGGSound) and applied without finetuning on another (VAS).
See more discussion in the main text, Section~\ref{sec:results_sampling}.

\begin{figure*}
\begin{center}
\includegraphics[width=\linewidth]{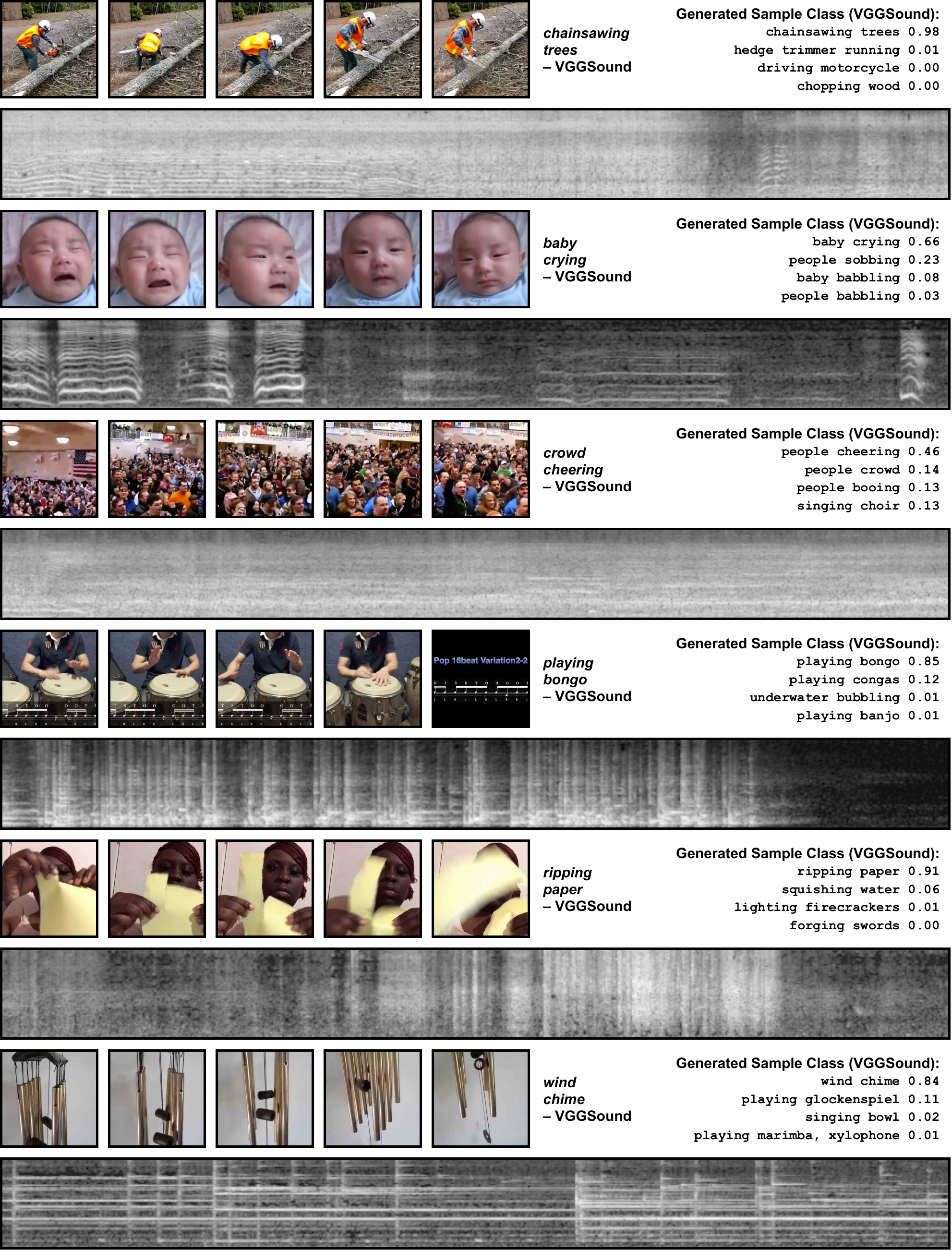}
\end{center}
\PlaceText{20.08mm}{189.11mm}{
\includemedia[ addresource=mp3/chainsawing_trees_audio.mp3, flashvars={ source=mp3/chainsawing_trees_audio.mp3 &autoPlay=false } ]{
\transparent{0.8} \color{green} \tiny \parbox{\dimexpr\linewidth-2\fboxsep-2\fboxrule\relax}{\faPlayCircleO~\textsf{Click to Play \\ in Adobe Reader \\[2.8em]}}}{APlayer.swf}
}
\PlaceText{20.08mm}{160.61mm}{
\includemedia[ addresource=mp3/baby_crying_audio.mp3, flashvars={ source=mp3/baby_crying_audio.mp3 &autoPlay=false } ]{
\transparent{0.8} \color{green} \tiny \parbox{\dimexpr\linewidth-2\fboxsep-2\fboxrule\relax}{\faPlayCircleO~\textsf{Click to Play \\ in Adobe Reader \\[2.8em]}}}{APlayer.swf}
}
\PlaceText{20.08mm}{132.11mm}{
\includemedia[ addresource=mp3/people_cheering_audio.mp3, flashvars={ source=mp3/people_cheering_audio.mp3 &autoPlay=false } ]{
\transparent{0.8} \color{green} \tiny \parbox{\dimexpr\linewidth-2\fboxsep-2\fboxrule\relax}{\faPlayCircleO~\textsf{Click to Play \\ in Adobe Reader \\[2.8em]}}}{APlayer.swf}
}
\PlaceText{20.08mm}{103.61mm}{
\includemedia[ addresource=mp3/playing_bongo_audio.mp3, flashvars={ source=mp3/playing_bongo_audio.mp3 &autoPlay=false } ]{
\transparent{0.8} \color{green} \tiny \parbox{\dimexpr\linewidth-2\fboxsep-2\fboxrule\relax}{\faPlayCircleO~\textsf{Click to Play \\ in Adobe Reader \\[2.8em]}}}{APlayer.swf}
}
\PlaceText{20.08mm}{75.11mm}{
\includemedia[ addresource=mp3/ripping_paper_audio.mp3, flashvars={ source=mp3/ripping_paper_audio.mp3 &autoPlay=false } ]{
\transparent{0.8} \color{green} \tiny \parbox{\dimexpr\linewidth-2\fboxsep-2\fboxrule\relax}{\faPlayCircleO~\textsf{Click to Play \\ in Adobe Reader \\[2.8em]}}}{APlayer.swf}
}
\PlaceText{20.08mm}{47.11mm}{
\includemedia[ addresource=mp3/wind_chime_audio.mp3, flashvars={ source=mp3/wind_chime_audio.mp3 &autoPlay=false } ]{
\transparent{0.8} \color{green} \tiny \parbox{\dimexpr\linewidth-2\fboxsep-2\fboxrule\relax}{\faPlayCircleO~\textsf{Click to Play \\ in Adobe Reader \\[2.8em]}}}{APlayer.swf}
}
\vspace{-2ex}
   \caption{\normalsize Samples produced by the conditional cross-modal sampler trained on the VGGSound dataset with the VGGSound codebook (using the setting (a) with 5 input features -- see Section~\ref{sec:results_sampling}). Also, see Figure~\ref{fig:main} on the title page and \& Figure~\ref{fig:sampling_main} in Section~\ref{sec:results_sampling}.}
\label{fig:more-sampling-results-vggsound}
\end{figure*}

\begin{figure*}
\begin{center}
\vspace{-1.2ex}
\includegraphics[width=\linewidth]{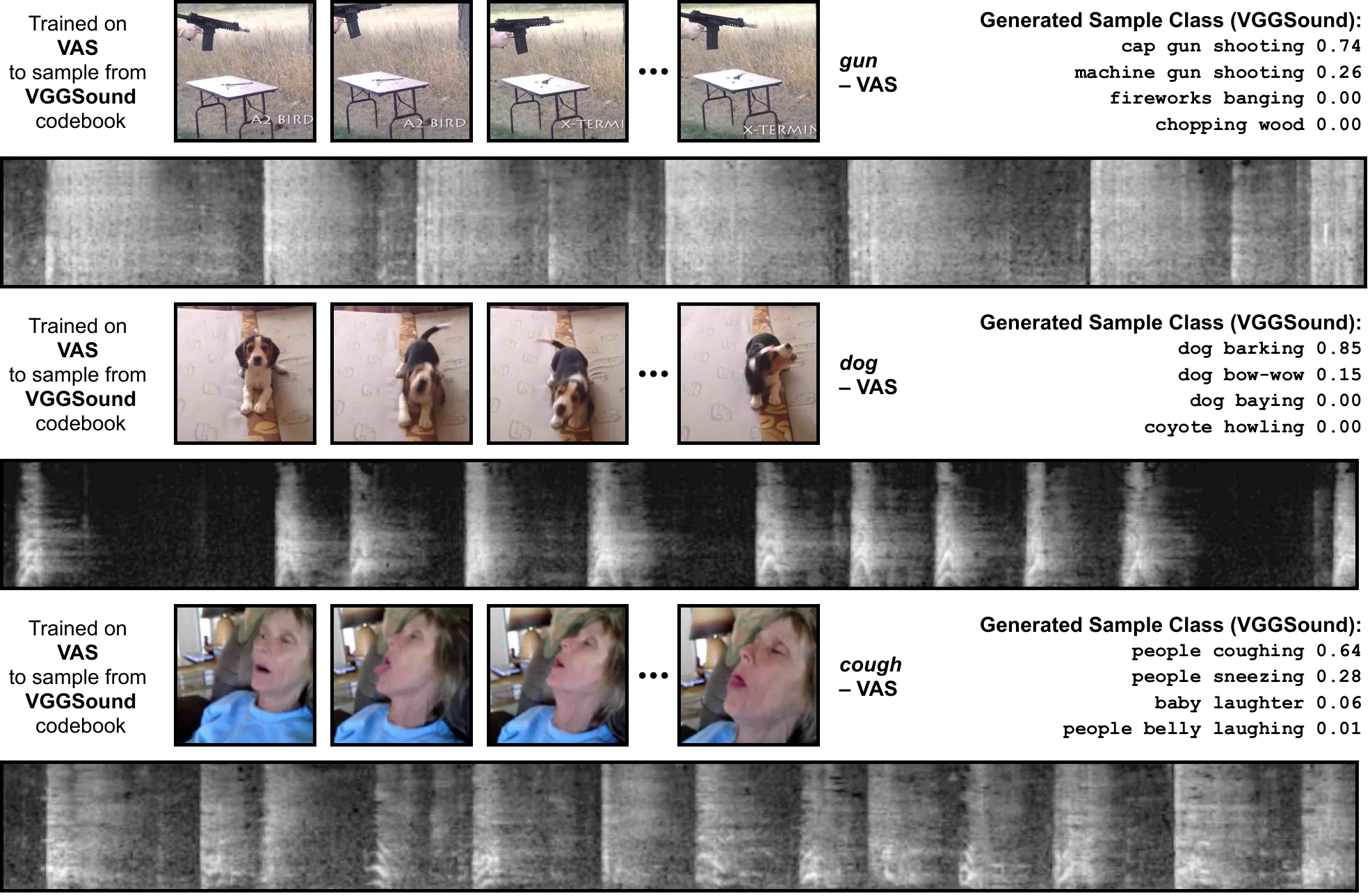}
\end{center}
\PlaceText{23.08mm}{195.11mm}{
\includemedia[ addresource=mp3/gun_audio.mp3, flashvars={ source=mp3/gun_audio.mp3 &autoPlay=false } ]{
\transparent{0.8} \color{green} \tiny \parbox{\dimexpr\linewidth-2\fboxsep-2\fboxrule\relax}{\faPlayCircleO~\textsf{Click to Play \\ in Adobe Reader \\[2.8em]}}}{APlayer.swf}
}
\PlaceText{23.08mm}{166.61mm}{
\includemedia[ addresource=mp3/dog_audio.mp3, flashvars={ source=mp3/dog_audio.mp3 &autoPlay=false } ]{
\transparent{0.8} \color{green} \tiny \parbox{\dimexpr\linewidth-2\fboxsep-2\fboxrule\relax}{\faPlayCircleO~\textsf{Click to Play \\ in Adobe Reader \\[2.8em]}}}{APlayer.swf}
}
\PlaceText{23.08mm}{138.11mm}{
\includemedia[ addresource=mp3/cough_audio.mp3, flashvars={ source=mp3/cough_audio.mp3 &autoPlay=false } ]{
\transparent{0.8} \color{green} \tiny \parbox{\dimexpr\linewidth-2\fboxsep-2\fboxrule\relax}{\faPlayCircleO~\textsf{Click to Play \\ in Adobe Reader \\[2.8em]}}}{APlayer.swf}
}
\vspace{-7ex}
   \caption{\normalsize Samples produced by conditional cross-modal sampler trained on the VAS dataset with the VGGSound codebook (setting (b) with 212 input features -- see Sec.~\ref{sec:results_sampling} and Fig.~\ref{fig:sampling_main}. The number of training videos for \textit{gun}, \textit{dog}, and \textit{cough}: 802, 2657, and 314.}
\label{fig:more-sampling-results-vggsound2vas}
\end{figure*}

\begin{figure*}
\begin{center}
\includegraphics[width=\linewidth]{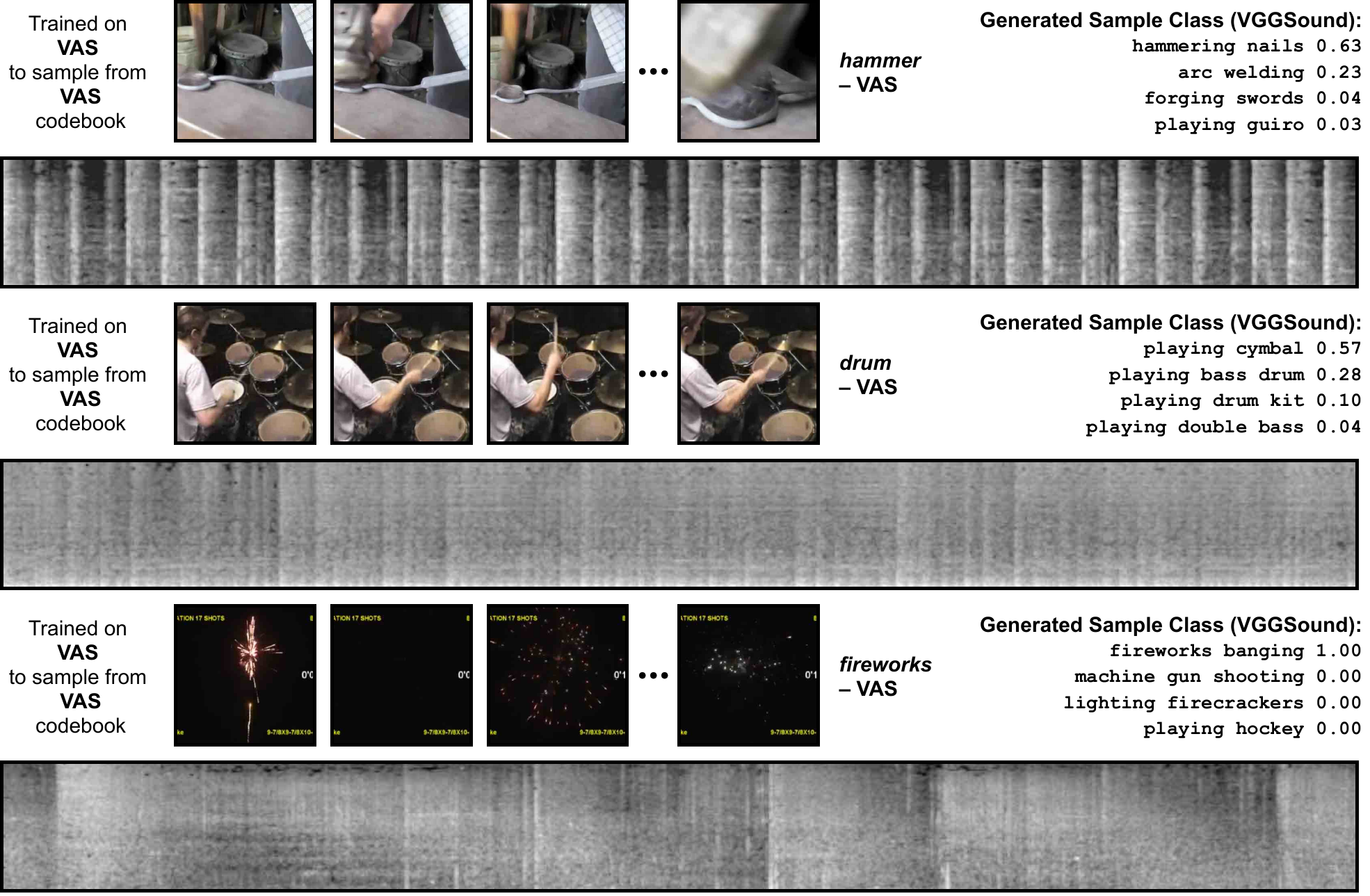}
\end{center}
\PlaceText{23.08mm}{91.61mm}{
\includemedia[ addresource=mp3/hammer_audio.mp3, flashvars={ source=mp3/hammer_audio.mp3 &autoPlay=false } ]{
\transparent{0.8} \color{green} \tiny \parbox{\dimexpr\linewidth-2\fboxsep-2\fboxrule\relax}{\faPlayCircleO~\textsf{Click to Play \\ in Adobe Reader \\[2.8em]}}}{APlayer.swf}
}
\PlaceText{23.08mm}{62.61mm}{
\includemedia[ addresource=mp3/drum_audio.mp3, flashvars={ source=mp3/drum_audio.mp3 &autoPlay=false } ]{
\transparent{0.8} \color{green} \tiny \parbox{\dimexpr\linewidth-2\fboxsep-2\fboxrule\relax}{\faPlayCircleO~\textsf{Click to Play \\ in Adobe Reader \\[2.8em]}}}{APlayer.swf}
}
\PlaceText{23.08mm}{34.11mm}{
\includemedia[ addresource=mp3/fireworks_audio.mp3, flashvars={ source=mp3/fireworks_audio.mp3 &autoPlay=false } ]{
\transparent{0.8} \color{green} \tiny \parbox{\dimexpr\linewidth-2\fboxsep-2\fboxrule\relax}{\faPlayCircleO~\textsf{Click to Play \\ in Adobe Reader \\[2.8em]}}}{APlayer.swf}
}
\vspace{-7ex}
   \caption{\normalsize Samples produced by conditional cross-modal sampler trained on the VAS dataset with the VAS codebook (setting (c) with 212 input features -- see Sec.~\ref{sec:results_sampling}. Also, see Fig.~\ref{fig:sampling_main} in Sec.~\ref{sec:results_sampling}. The num. of training videos for \textit{hammer}, \textit{drum}, \& \textit{fireworks}: 319, 2477, and 2986.}
\label{fig:more-sampling-results-vas2vas}
\end{figure*}

\paragraph{More Visually-Guided Sound Generation Results}
In Figures~\ref{fig:more-sampling-results-vggsound}, \ref{fig:more-sampling-results-vggsound2vas}, and \ref{fig:more-sampling-results-vas2vas}, we provide more visually guided samples for all three settings (a, b, c) described in detail in Section~\ref{sec:results_sampling}.
Our model is capable of generating relevant and high fidelity spectrograms for multiple data classes.
Note that some VAS classes have a small number of videos in the training set (\eg 802 for \textit{gun}, 314 for \textit{cough}, or 319 for \textit{hammer} classes), yet the model is capable of generating high-fidelity samples.

\begin{figure*}
\begin{center}
\includegraphics[width=\linewidth]{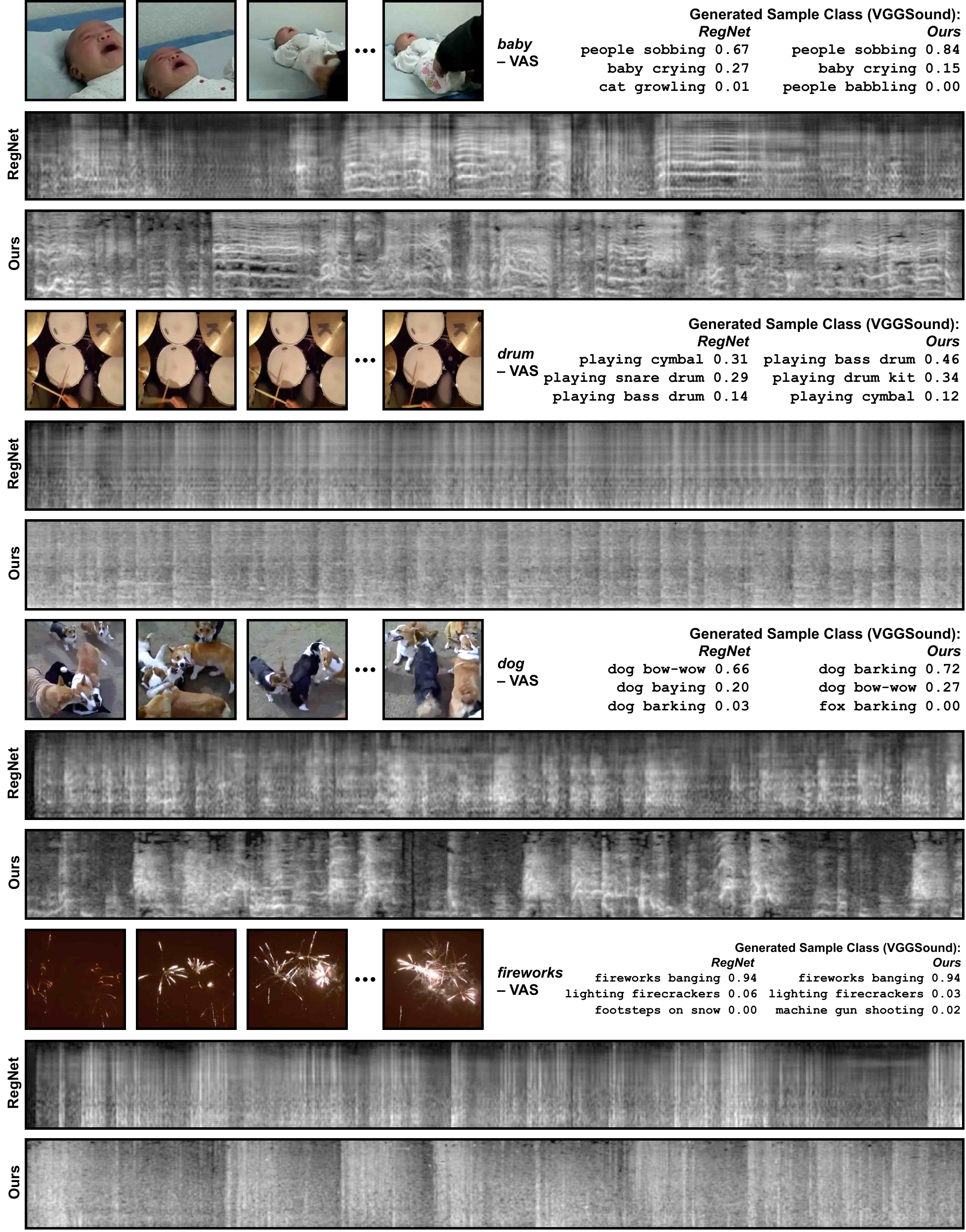}
\end{center}
\PlaceText{23.08mm}{194.11mm}{
\includemedia[ addresource=mp3/regnet_baby_audio.mp3, flashvars={ source=mp3/regnet_baby_audio.mp3 &autoPlay=false } ]{
\transparent{0.8} \color{green} \tiny \parbox{\dimexpr\linewidth-2\fboxsep-2\fboxrule\relax}{\faPlayCircleO~\textsf{Click to Play \\ in Adobe Reader \\[2.8em]}}}{APlayer.swf}
}
\PlaceText{23.08mm}{181.11mm}{
\includemedia[ addresource=mp3/ours_baby_audio.mp3, flashvars={ source=mp3/ours_baby_audio.mp3 &autoPlay=false } ]{
\transparent{0.8} \color{green} \tiny \parbox{\dimexpr\linewidth-2\fboxsep-2\fboxrule\relax}{\faPlayCircleO~\textsf{Click to Play \\ in Adobe Reader \\[2.8em]}}}{APlayer.swf}
}
\PlaceText{23.08mm}{153.11mm}{
\includemedia[ addresource=mp3/regnet_drum_audio.mp3, flashvars={ source=mp3/regnet_drum_audio.mp3 &autoPlay=false } ]{
\transparent{0.8} \color{green} \tiny \parbox{\dimexpr\linewidth-2\fboxsep-2\fboxrule\relax}{\faPlayCircleO~\textsf{Click to Play \\ in Adobe Reader \\[2.8em]}}}{APlayer.swf}
}
\PlaceText{23.08mm}{139.61mm}{
\includemedia[ addresource=mp3/ours_drum_audio.mp3, flashvars={ source=mp3/ours_drum_audio.mp3 &autoPlay=false } ]{
\transparent{0.8} \color{green} \tiny \parbox{\dimexpr\linewidth-2\fboxsep-2\fboxrule\relax}{\faPlayCircleO~\textsf{Click to Play \\ in Adobe Reader \\[2.8em]}}}{APlayer.swf}
}
\PlaceText{23.08mm}{111.61mm}{
\includemedia[ addresource=mp3/regnet_dog_audio.mp3, flashvars={ source=mp3/regnet_dog_audio.mp3 &autoPlay=false } ]{
\transparent{0.8} \color{green} \tiny \parbox{\dimexpr\linewidth-2\fboxsep-2\fboxrule\relax}{\faPlayCircleO~\textsf{Click to Play \\ in Adobe Reader \\[2.8em]}}}{APlayer.swf}
}
\PlaceText{23.08mm}{98.11mm}{
\includemedia[ addresource=mp3/ours_dog_audio.mp3, flashvars={ source=mp3/ours_dog_audio.mp3 &autoPlay=false } ]{
\transparent{0.8} \color{green} \tiny \parbox{\dimexpr\linewidth-2\fboxsep-2\fboxrule\relax}{\faPlayCircleO~\textsf{Click to Play \\ in Adobe Reader \\[2.8em]}}}{APlayer.swf}
}
\PlaceText{23.08mm}{70.11mm}{
\includemedia[ addresource=mp3/regnet_fireworks_audio.mp3, flashvars={ source=mp3/regnet_fireworks_audio.mp3 &autoPlay=false } ]{
\transparent{0.8} \color{green} \tiny \parbox{\dimexpr\linewidth-2\fboxsep-2\fboxrule\relax}{\faPlayCircleO~\textsf{Click to Play \\ in Adobe Reader \\[2.8em]}}}{APlayer.swf}
}
\PlaceText{23.08mm}{56.61mm}{
\includemedia[ addresource=mp3/ours_fireworks_audio.mp3, flashvars={ source=mp3/ours_fireworks_audio.mp3 &autoPlay=false } ]{
\transparent{0.8} \color{green} \tiny \parbox{\dimexpr\linewidth-2\fboxsep-2\fboxrule\relax}{\faPlayCircleO~\textsf{Click to Play \\ in Adobe Reader \\[2.8em]}}}{APlayer.swf}
}
\vspace{-5ex}
   \caption{\normalsize Comparison to the state-of-the-art model (RegNet \cite{chen2020generating}) on the VAS dataset. The baseline model is trained separately for each class and, in total, has 2+ times more parameters than our model. Moreover, the baseline relies on the WaveNet \cite{oord2016wavenet} vocoder, also, trained separated for every class. In contrast, our model is trained with all classes simultaneously while the vocoder is trained on a different dataset (309-class VGGSound, Sec.~\ref{sec:vocoder}). In addition, our model is more than 100 times faster during sampling than the baseline. Adobe Reader can be used to play audio. Ours is reported using the setting (c) with 212 features and a class in the condition. See more details in Sec.~\ref{sec:results-regnet} and Tab.~\ref{tab:regnet}. Best viewed if zoomed-in.}
\label{fig:more-regnet-comparisons}
\end{figure*}

\paragraph{More Baseline Comparison Results}
More results of the comparison to the baseline (RegNet \cite{chen2020generating}) on the VAS dataset are provided in Figure~\ref{fig:more-regnet-comparisons}.
Our model provides higher quality results than the baseline while supporting multiple data classes in a single model and having more than two times fewer parameters (also see Section~\ref{sec:results-regnet}).

\paragraph{Generating ``Infinite'' Samples}
In Figure~\ref{fig:drum-solo} we provide a long sample that we call ``the great drum solo''.
While being trained on significantly shorter segments ($\sim$10 seconds), the model can generate lengthy, relevant,
and high fidelity samples.

\begin{figure*}
\begin{center}
\includegraphics[width=\linewidth]{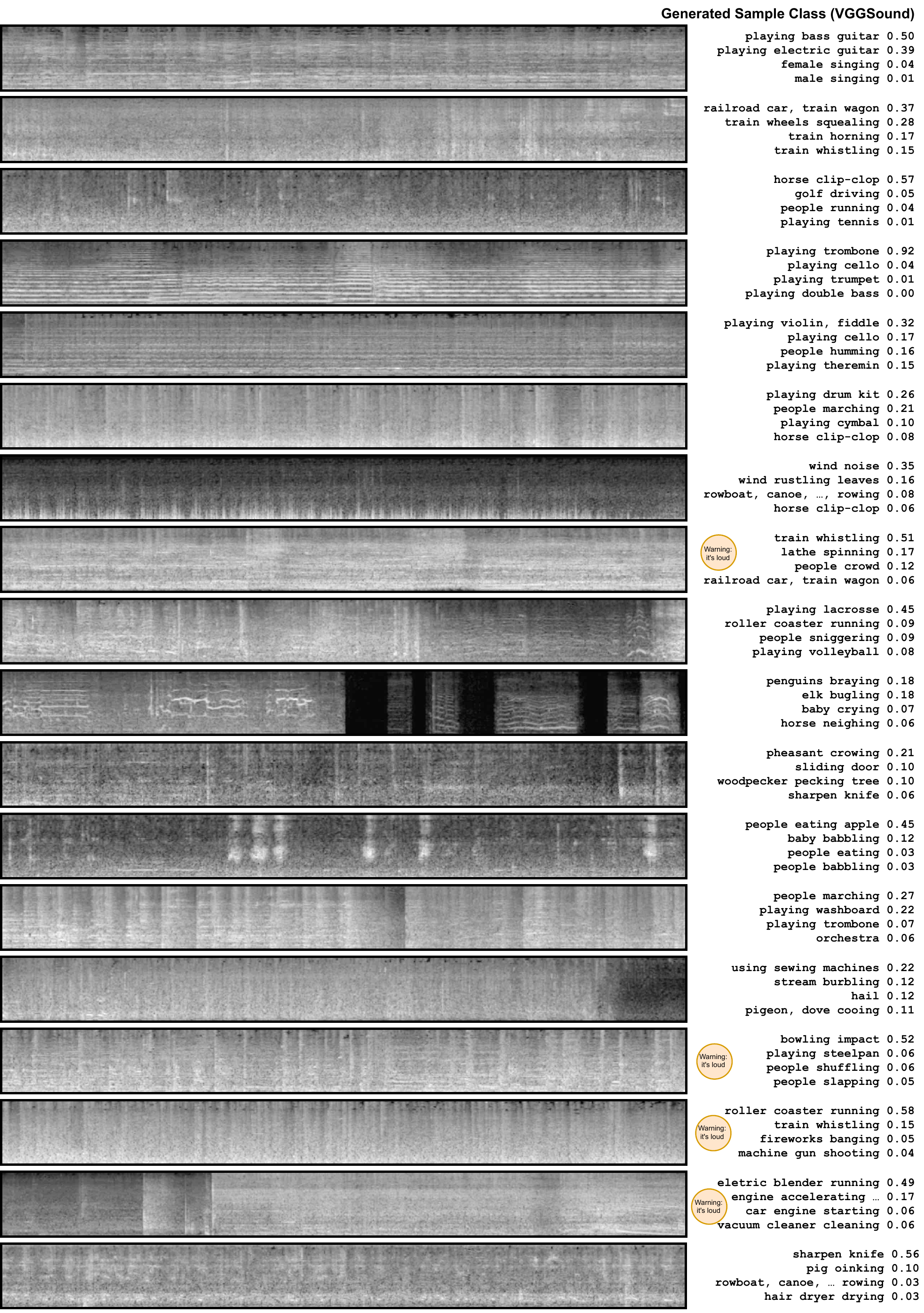}
\PlaceText{23.08mm}{207.11mm}{
\includemedia[ addresource=mp3/no-feats-vggsound/nopix_2021-06-30T16-29-56_vocoder.mp3, flashvars={ source=mp3/no-feats-vggsound/nopix_2021-06-30T16-29-56_vocoder.mp3 &autoPlay=false } ]{
\transparent{0.8} \color{green} \tiny \parbox{\dimexpr 0.7\linewidth-2\fboxsep-2\fboxrule\relax}{\faPlayCircleO~\textsf{Click to Play \\ in Adobe Reader \\[1.5em]}}}{APlayer.swf}
}
\PlaceText{23.08mm}{197.11mm}{
\includemedia[ addresource=mp3/no-feats-vggsound/nopix_2021-06-30T16-30-24_vocoder.mp3, flashvars={ source=mp3/no-feats-vggsound/nopix_2021-06-30T16-30-24_vocoder.mp3 &autoPlay=false } ]{
\transparent{0.8} \color{green} \tiny \parbox{\dimexpr 0.7\linewidth-2\fboxsep-2\fboxrule\relax}{\faPlayCircleO~\textsf{Click to Play \\ in Adobe Reader \\[1.5em]}}}{APlayer.swf}
}
\PlaceText{23.08mm}{187.11mm}{
\includemedia[ addresource=mp3/no-feats-vggsound/nopix_2021-06-30T16-30-40_vocoder.mp3, flashvars={ source=mp3/no-feats-vggsound/nopix_2021-06-30T16-30-40_vocoder.mp3 &autoPlay=false } ]{
\transparent{0.8} \color{green} \tiny \parbox{\dimexpr 0.7\linewidth-2\fboxsep-2\fboxrule\relax}{\faPlayCircleO~\textsf{Click to Play \\ in Adobe Reader \\[1.5em]}}}{APlayer.swf}
}
\PlaceText{23.08mm}{177.11mm}{
\includemedia[ addresource=mp3/no-feats-vggsound/nopix_2021-06-30T16-30-57_vocoder.mp3, flashvars={ source=mp3/no-feats-vggsound/nopix_2021-06-30T16-30-57_vocoder.mp3 &autoPlay=false } ]{
\transparent{0.8} \color{green} \tiny \parbox{\dimexpr 0.7\linewidth-2\fboxsep-2\fboxrule\relax}{\faPlayCircleO~\textsf{Click to Play \\ in Adobe Reader \\[1.5em]}}}{APlayer.swf}
}
\PlaceText{23.08mm}{167.11mm}{
\includemedia[ addresource=mp3/no-feats-vggsound/nopix_2021-06-30T16-31-14_vocoder.mp3, flashvars={ source=mp3/no-feats-vggsound/nopix_2021-06-30T16-31-14_vocoder.mp3 &autoPlay=false } ]{
\transparent{0.8} \color{green} \tiny \parbox{\dimexpr 0.7\linewidth-2\fboxsep-2\fboxrule\relax}{\faPlayCircleO~\textsf{Click to Play \\ in Adobe Reader \\[1.5em]}}}{APlayer.swf}
}
\PlaceText{23.08mm}{157.11mm}{
\includemedia[ addresource=mp3/no-feats-vggsound/nopix_2021-06-30T16-31-41_vocoder.mp3, flashvars={ source=mp3/no-feats-vggsound/nopix_2021-06-30T16-31-41_vocoder.mp3 &autoPlay=false } ]{
\transparent{0.8} \color{green} \tiny \parbox{\dimexpr 0.7\linewidth-2\fboxsep-2\fboxrule\relax}{\faPlayCircleO~\textsf{Click to Play \\ in Adobe Reader \\[1.5em]}}}{APlayer.swf}
}
\PlaceText{23.08mm}{147.11mm}{
\includemedia[ addresource=mp3/no-feats-vggsound/nopix_2021-06-30T16-31-56_vocoder.mp3, flashvars={ source=mp3/no-feats-vggsound/nopix_2021-06-30T16-31-56_vocoder.mp3 &autoPlay=false } ]{
\transparent{0.8} \color{green} \tiny \parbox{\dimexpr 0.7\linewidth-2\fboxsep-2\fboxrule\relax}{\faPlayCircleO~\textsf{Click to Play \\ in Adobe Reader \\[1.5em]}}}{APlayer.swf}
}
\PlaceText{23.08mm}{137.11mm}{
\includemedia[ addresource=mp3/no-feats-vggsound/nopix_2021-06-30T16-32-11_vocoder.mp3, flashvars={ source=mp3/no-feats-vggsound/nopix_2021-06-30T16-32-11_vocoder.mp3 &autoPlay=false } ]{
\transparent{0.8} \color{green} \tiny \parbox{\dimexpr 0.7\linewidth-2\fboxsep-2\fboxrule\relax}{\faPlayCircleO~\textsf{Click to Play \\ in Adobe Reader \\[1.5em]}}}{APlayer.swf}
}
\PlaceText{23.08mm}{127.11mm}{
\includemedia[ addresource=mp3/no-feats-vggsound/nopix_2021-06-30T16-32-25_vocoder.mp3, flashvars={ source=mp3/no-feats-vggsound/nopix_2021-06-30T16-32-25_vocoder.mp3 &autoPlay=false } ]{
\transparent{0.8} \color{green} \tiny \parbox{\dimexpr 0.7\linewidth-2\fboxsep-2\fboxrule\relax}{\faPlayCircleO~\textsf{Click to Play \\ in Adobe Reader \\[1.5em]}}}{APlayer.swf}
}
\PlaceText{23.08mm}{117.11mm}{
\includemedia[ addresource=mp3/no-feats-vggsound/nopix_2021-06-30T16-32-44_vocoder.mp3, flashvars={ source=mp3/no-feats-vggsound/nopix_2021-06-30T16-32-44_vocoder.mp3 &autoPlay=false } ]{
\transparent{0.8} \color{green} \tiny \parbox{\dimexpr 0.7\linewidth-2\fboxsep-2\fboxrule\relax}{\faPlayCircleO~\textsf{Click to Play \\ in Adobe Reader \\[1.5em]}}}{APlayer.swf}
}
\PlaceText{23.08mm}{107.11mm}{
\includemedia[ addresource=mp3/no-feats-vggsound/nopix_2021-06-30T16-32-58_vocoder.mp3, flashvars={ source=mp3/no-feats-vggsound/nopix_2021-06-30T16-32-58_vocoder.mp3 &autoPlay=false } ]{
\transparent{0.8} \color{green} \tiny \parbox{\dimexpr 0.7\linewidth-2\fboxsep-2\fboxrule\relax}{\faPlayCircleO~\textsf{Click to Play \\ in Adobe Reader \\[1.5em]}}}{APlayer.swf}
}
\PlaceText{23.08mm}{97.11mm}{
\includemedia[ addresource=mp3/no-feats-vggsound/nopix_2021-06-30T16-33-20_vocoder.mp3, flashvars={ source=mp3/no-feats-vggsound/nopix_2021-06-30T16-33-20_vocoder.mp3 &autoPlay=false } ]{
\transparent{0.8} \color{green} \tiny \parbox{\dimexpr 0.7\linewidth-2\fboxsep-2\fboxrule\relax}{\faPlayCircleO~\textsf{Click to Play \\ in Adobe Reader \\[1.5em]}}}{APlayer.swf}
}
\PlaceText{23.08mm}{87.11mm}{
\includemedia[ addresource=mp3/no-feats-vggsound/nopix_2021-06-30T16-33-45_vocoder.mp3, flashvars={ source=mp3/no-feats-vggsound/nopix_2021-06-30T16-33-45_vocoder.mp3 &autoPlay=false } ]{
\transparent{0.8} \color{green} \tiny \parbox{\dimexpr 0.7\linewidth-2\fboxsep-2\fboxrule\relax}{\faPlayCircleO~\textsf{Click to Play \\ in Adobe Reader \\[1.5em]}}}{APlayer.swf}
}
\PlaceText{23.08mm}{77.11mm}{
\includemedia[ addresource=mp3/no-feats-vggsound/nopix_2021-06-30T16-34-12_vocoder.mp3, flashvars={ source=mp3/no-feats-vggsound/nopix_2021-06-30T16-34-12_vocoder.mp3 &autoPlay=false } ]{
\transparent{0.8} \color{green} \tiny \parbox{\dimexpr 0.7\linewidth-2\fboxsep-2\fboxrule\relax}{\faPlayCircleO~\textsf{Click to Play \\ in Adobe Reader \\[1.5em]}}}{APlayer.swf}
}
\PlaceText{23.08mm}{67.11mm}{
\includemedia[ addresource=mp3/no-feats-vggsound/nopix_2021-06-30T16-35-18_vocoder.mp3, flashvars={ source=mp3/no-feats-vggsound/nopix_2021-06-30T16-35-18_vocoder.mp3 &autoPlay=false } ]{
\transparent{0.8} \color{green} \tiny \parbox{\dimexpr 0.7\linewidth-2\fboxsep-2\fboxrule\relax}{\faPlayCircleO~\textsf{Click to Play \\ in Adobe Reader \\[1.5em]}}}{APlayer.swf}
}
\PlaceText{23.08mm}{57.11mm}{
\includemedia[ addresource=mp3/no-feats-vggsound/nopix_2021-06-30T16-35-31_vocoder.mp3, flashvars={ source=mp3/no-feats-vggsound/nopix_2021-06-30T16-35-31_vocoder.mp3 &autoPlay=false } ]{
\transparent{0.8} \color{green} \tiny \parbox{\dimexpr 0.7\linewidth-2\fboxsep-2\fboxrule\relax}{\faPlayCircleO~\textsf{Click to Play \\ in Adobe Reader \\[1.5em]}}}{APlayer.swf}
}
\PlaceText{23.08mm}{47.11mm}{
\includemedia[ addresource=mp3/no-feats-vggsound/nopix_2021-06-30T16-35-45_vocoder.mp3, flashvars={ source=mp3/no-feats-vggsound/nopix_2021-06-30T16-35-45_vocoder.mp3 &autoPlay=false } ]{
\transparent{0.8} \color{green} \tiny \parbox{\dimexpr 0.7\linewidth-2\fboxsep-2\fboxrule\relax}{\faPlayCircleO~\textsf{Click to Play \\ in Adobe Reader \\[1.5em]}}}{APlayer.swf}
}
\PlaceText{23.08mm}{37.11mm}{
\includemedia[ addresource=mp3/no-feats-vggsound/nopix_2021-06-30T16-36-13_vocoder.mp3, flashvars={ source=mp3/no-feats-vggsound/nopix_2021-06-30T16-36-13_vocoder.mp3 &autoPlay=false } ]{
\transparent{0.8} \color{green} \tiny \parbox{\dimexpr 0.7\linewidth-2\fboxsep-2\fboxrule\relax}{\faPlayCircleO~\textsf{Click to Play \\ in Adobe Reader \\[1.5em]}}}{APlayer.swf}
}
\end{center}
\vspace{-2ex}
   \caption{\normalsize Random samples from a model without visual conditioning. Both the codebook and the transformer are trained on audio clips from VGGSound (the setting (a), \textit{No feats}).}
\label{fig:no-feats-vggsound}
\end{figure*}

\begin{figure*}
\begin{center}
\includegraphics[width=\linewidth]{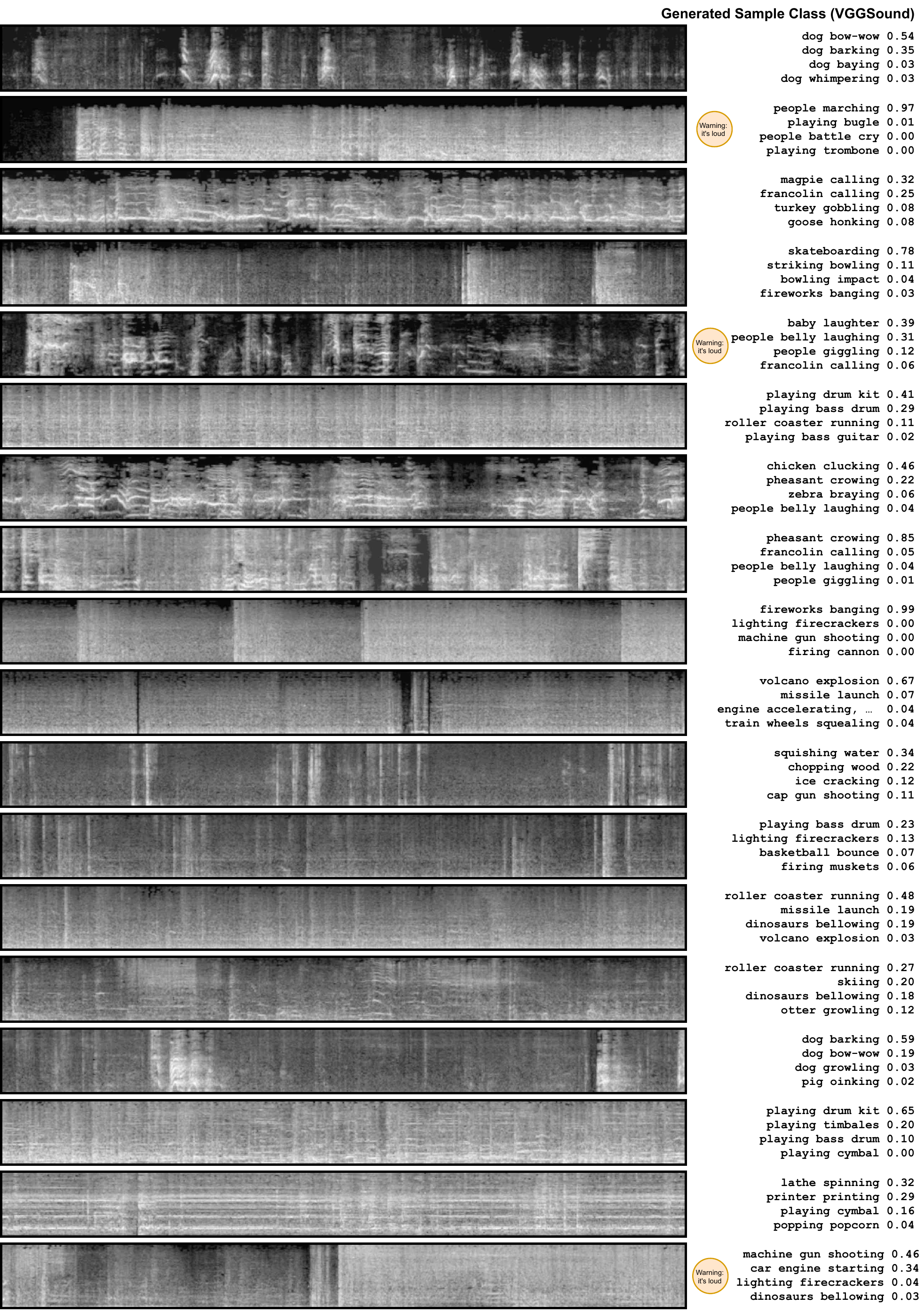}
\PlaceText{20.08mm}{207.11mm}{
\includemedia[ addresource=mp3/no-feats-vas/nopix_2021-06-30T22-38-19_vocoder.mp3, flashvars={ source=mp3/no-feats-vas/nopix_2021-06-30T22-38-19_vocoder.mp3 &autoPlay=false } ]{
\transparent{0.8} \color{green} \tiny \parbox{\dimexpr 0.7\linewidth-2\fboxsep-2\fboxrule\relax}{\faPlayCircleO~\textsf{Click to Play \\ in Adobe Reader \\[1.5em]}}}{APlayer.swf}
}
\PlaceText{20.08mm}{197.11mm}{
\includemedia[ addresource=mp3/no-feats-vas/nopix_2021-06-30T22-38-33_vocoder.mp3, flashvars={ source=mp3/no-feats-vas/nopix_2021-06-30T22-38-33_vocoder.mp3 &autoPlay=false } ]{
\transparent{0.8} \color{green} \tiny \parbox{\dimexpr 0.7\linewidth-2\fboxsep-2\fboxrule\relax}{\faPlayCircleO~\textsf{Click to Play \\ in Adobe Reader \\[1.5em]}}}{APlayer.swf}
}
\PlaceText{20.08mm}{187.11mm}{
\includemedia[ addresource=mp3/no-feats-vas/nopix_2021-06-30T22-38-48_vocoder.mp3, flashvars={ source=mp3/no-feats-vas/nopix_2021-06-30T22-38-48_vocoder.mp3 &autoPlay=false } ]{
\transparent{0.8} \color{green} \tiny \parbox{\dimexpr 0.7\linewidth-2\fboxsep-2\fboxrule\relax}{\faPlayCircleO~\textsf{Click to Play \\ in Adobe Reader \\[1.5em]}}}{APlayer.swf}
}
\PlaceText{20.08mm}{177.11mm}{
\includemedia[ addresource=mp3/no-feats-vas/nopix_2021-06-30T22-39-02_vocoder.mp3, flashvars={ source=mp3/no-feats-vas/nopix_2021-06-30T22-39-02_vocoder.mp3 &autoPlay=false } ]{
\transparent{0.8} \color{green} \tiny \parbox{\dimexpr 0.7\linewidth-2\fboxsep-2\fboxrule\relax}{\faPlayCircleO~\textsf{Click to Play \\ in Adobe Reader \\[1.5em]}}}{APlayer.swf}
}
\PlaceText{20.08mm}{167.11mm}{
\includemedia[ addresource=mp3/no-feats-vas/nopix_2021-06-30T22-39-16_vocoder.mp3, flashvars={ source=mp3/no-feats-vas/nopix_2021-06-30T22-39-16_vocoder.mp3 &autoPlay=false } ]{
\transparent{0.8} \color{green} \tiny \parbox{\dimexpr 0.7\linewidth-2\fboxsep-2\fboxrule\relax}{\faPlayCircleO~\textsf{Click to Play \\ in Adobe Reader \\[1.5em]}}}{APlayer.swf}
}
\PlaceText{20.08mm}{157.11mm}{
\includemedia[ addresource=mp3/no-feats-vas/nopix_2021-06-30T22-39-30_vocoder.mp3, flashvars={ source=mp3/no-feats-vas/nopix_2021-06-30T22-39-30_vocoder.mp3 &autoPlay=false } ]{
\transparent{0.8} \color{green} \tiny \parbox{\dimexpr 0.7\linewidth-2\fboxsep-2\fboxrule\relax}{\faPlayCircleO~\textsf{Click to Play \\ in Adobe Reader \\[1.5em]}}}{APlayer.swf}
}
\PlaceText{20.08mm}{147.11mm}{
\includemedia[ addresource=mp3/no-feats-vas/nopix_2021-06-30T22-39-44_vocoder.mp3, flashvars={ source=mp3/no-feats-vas/nopix_2021-06-30T22-39-44_vocoder.mp3 &autoPlay=false } ]{
\transparent{0.8} \color{green} \tiny \parbox{\dimexpr 0.7\linewidth-2\fboxsep-2\fboxrule\relax}{\faPlayCircleO~\textsf{Click to Play \\ in Adobe Reader \\[1.5em]}}}{APlayer.swf}
}
\PlaceText{20.08mm}{137.11mm}{
\includemedia[ addresource=mp3/no-feats-vas/nopix_2021-06-30T22-39-59_vocoder.mp3, flashvars={ source=mp3/no-feats-vas/nopix_2021-06-30T22-39-59_vocoder.mp3 &autoPlay=false } ]{
\transparent{0.8} \color{green} \tiny \parbox{\dimexpr 0.7\linewidth-2\fboxsep-2\fboxrule\relax}{\faPlayCircleO~\textsf{Click to Play \\ in Adobe Reader \\[1.5em]}}}{APlayer.swf}
}
\PlaceText{20.08mm}{127.11mm}{
\includemedia[ addresource=mp3/no-feats-vas/nopix_2021-06-30T22-40-14_vocoder.mp3, flashvars={ source=mp3/no-feats-vas/nopix_2021-06-30T22-40-14_vocoder.mp3 &autoPlay=false } ]{
\transparent{0.8} \color{green} \tiny \parbox{\dimexpr 0.7\linewidth-2\fboxsep-2\fboxrule\relax}{\faPlayCircleO~\textsf{Click to Play \\ in Adobe Reader \\[1.5em]}}}{APlayer.swf}
}
\PlaceText{20.08mm}{117.11mm}{
\includemedia[ addresource=mp3/no-feats-vas/nopix_2021-06-30T22-40-28_vocoder.mp3, flashvars={ source=mp3/no-feats-vas/nopix_2021-06-30T22-40-28_vocoder.mp3 &autoPlay=false } ]{
\transparent{0.8} \color{green} \tiny \parbox{\dimexpr 0.7\linewidth-2\fboxsep-2\fboxrule\relax}{\faPlayCircleO~\textsf{Click to Play \\ in Adobe Reader \\[1.5em]}}}{APlayer.swf}
}
\PlaceText{20.08mm}{107.11mm}{
\includemedia[ addresource=mp3/no-feats-vas/nopix_2021-06-30T22-40-42_vocoder.mp3, flashvars={ source=mp3/no-feats-vas/nopix_2021-06-30T22-40-42_vocoder.mp3 &autoPlay=false } ]{
\transparent{0.8} \color{green} \tiny \parbox{\dimexpr 0.7\linewidth-2\fboxsep-2\fboxrule\relax}{\faPlayCircleO~\textsf{Click to Play \\ in Adobe Reader \\[1.5em]}}}{APlayer.swf}
}
\PlaceText{20.08mm}{97.11mm}{
\includemedia[ addresource=mp3/no-feats-vas/nopix_2021-06-30T22-40-56_vocoder.mp3, flashvars={ source=mp3/no-feats-vas/nopix_2021-06-30T22-40-56_vocoder.mp3 &autoPlay=false } ]{
\transparent{0.8} \color{green} \tiny \parbox{\dimexpr 0.7\linewidth-2\fboxsep-2\fboxrule\relax}{\faPlayCircleO~\textsf{Click to Play \\ in Adobe Reader \\[1.5em]}}}{APlayer.swf}
}
\PlaceText{20.08mm}{87.11mm}{
\includemedia[ addresource=mp3/no-feats-vas/nopix_2021-06-30T22-41-11_vocoder.mp3, flashvars={ source=mp3/no-feats-vas/nopix_2021-06-30T22-41-11_vocoder.mp3 &autoPlay=false } ]{
\transparent{0.8} \color{green} \tiny \parbox{\dimexpr 0.7\linewidth-2\fboxsep-2\fboxrule\relax}{\faPlayCircleO~\textsf{Click to Play \\ in Adobe Reader \\[1.5em]}}}{APlayer.swf}
}
\PlaceText{20.08mm}{77.11mm}{
\includemedia[ addresource=mp3/no-feats-vas/nopix_2021-06-30T22-41-25_vocoder.mp3, flashvars={ source=mp3/no-feats-vas/nopix_2021-06-30T22-41-25_vocoder.mp3 &autoPlay=false } ]{
\transparent{0.8} \color{green} \tiny \parbox{\dimexpr 0.7\linewidth-2\fboxsep-2\fboxrule\relax}{\faPlayCircleO~\textsf{Click to Play \\ in Adobe Reader \\[1.5em]}}}{APlayer.swf}
}
\PlaceText{20.08mm}{67.11mm}{
\includemedia[ addresource=mp3/no-feats-vas/nopix_2021-06-30T22-43-16_vocoder.mp3, flashvars={ source=mp3/no-feats-vas/nopix_2021-06-30T22-43-16_vocoder.mp3 &autoPlay=false } ]{
\transparent{0.8} \color{green} \tiny \parbox{\dimexpr 0.7\linewidth-2\fboxsep-2\fboxrule\relax}{\faPlayCircleO~\textsf{Click to Play \\ in Adobe Reader \\[1.5em]}}}{APlayer.swf}
}
\PlaceText{20.08mm}{57.11mm}{
\includemedia[ addresource=mp3/no-feats-vas/nopix_2021-06-30T22-43-32_vocoder.mp3, flashvars={ source=mp3/no-feats-vas/nopix_2021-06-30T22-43-32_vocoder.mp3 &autoPlay=false } ]{
\transparent{0.8} \color{green} \tiny \parbox{\dimexpr 0.7\linewidth-2\fboxsep-2\fboxrule\relax}{\faPlayCircleO~\textsf{Click to Play \\ in Adobe Reader \\[1.5em]}}}{APlayer.swf}
}
\PlaceText{20.08mm}{47.11mm}{
\includemedia[ addresource=mp3/no-feats-vas/nopix_2021-06-30T22-43-47_vocoder.mp3, flashvars={ source=mp3/no-feats-vas/nopix_2021-06-30T22-43-47_vocoder.mp3 &autoPlay=false } ]{
\transparent{0.8} \color{green} \tiny \parbox{\dimexpr 0.7\linewidth-2\fboxsep-2\fboxrule\relax}{\faPlayCircleO~\textsf{Click to Play \\ in Adobe Reader \\[1.5em]}}}{APlayer.swf}
}
\PlaceText{20.08mm}{37.11mm}{
\includemedia[ addresource=mp3/no-feats-vas/nopix_2021-06-30T22-44-02_vocoder.mp3, flashvars={ source=mp3/no-feats-vas/nopix_2021-06-30T22-44-02_vocoder.mp3 &autoPlay=false } ]{
\transparent{0.8} \color{green} \tiny \parbox{\dimexpr 0.7\linewidth-2\fboxsep-2\fboxrule\relax}{\faPlayCircleO~\textsf{Click to Play \\ in Adobe Reader \\[1.5em]}}}{APlayer.swf}
}

\end{center}
\vspace{-2ex}
   \caption{\normalsize Random samples from a model without visual conditioning. Both the codebook and the transformer are trained on audio clips from VAS (the setting (c), \textit{No feats}).}
\label{fig:no-feats-vas}
\end{figure*}

\paragraph{Samples without Condition (\textit{No Feats})}
In Figures~\ref{fig:no-feats-vggsound} and \ref{fig:no-feats-vas} we show randomly drawn samples from a model trained on the VGGSound and VAS datasets with the visual condition.
The samples are diverse, unique, and have high fidelity.

\begin{figure*}
\begin{center}
\includegraphics[width=\linewidth]{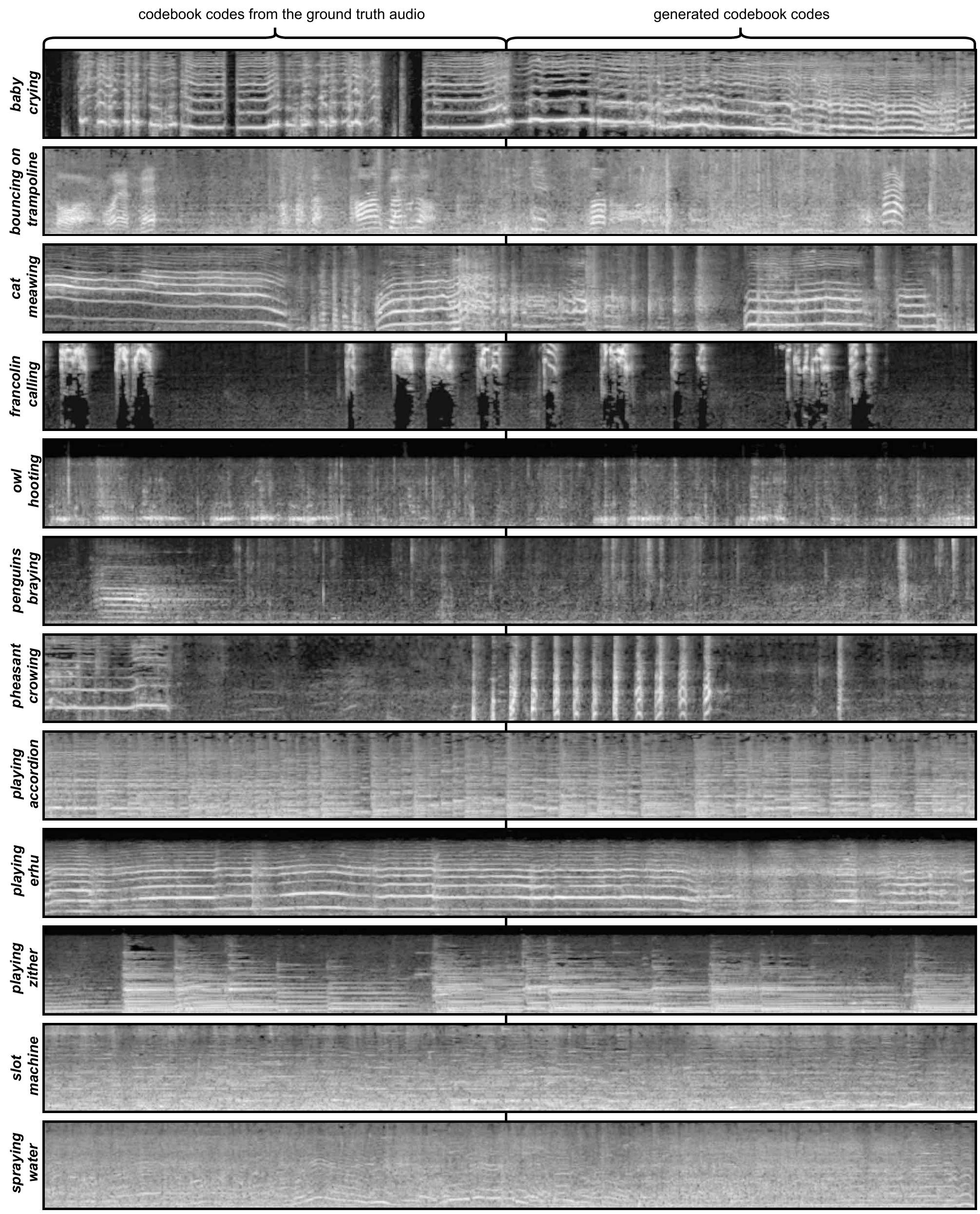}
\PlaceText{29.08mm}{195.11mm}{
\includemedia[ addresource=mp3/half-samples/half_2021-06-15T10-41-35_vocoder.mp3, flashvars={ source=mp3/half-samples/half_2021-06-15T10-41-35_vocoder.mp3 &autoPlay=false } ]{
\transparent{0.8} \color{green} \tiny \parbox{\dimexpr 0.9\linewidth-2\fboxsep-2\fboxrule\relax}{\faPlayCircleO~\textsf{Click to Play \\ in Adobe Reader \\[2.5em]}}}{APlayer.swf}
}
\PlaceText{29.08mm}{182.31mm}{
\includemedia[ addresource=mp3/half-samples/half_2021-06-15T10-41-05_vocoder.mp3, flashvars={ source=mp3/half-samples/half_2021-06-15T10-41-05_vocoder.mp3 &autoPlay=false } ]{
\transparent{0.8} \color{green} \tiny \parbox{\dimexpr 0.9\linewidth-2\fboxsep-2\fboxrule\relax}{\faPlayCircleO~\textsf{Click to Play \\ in Adobe Reader \\[2.5em]}}}{APlayer.swf}
}
\PlaceText{29.08mm}{169.51mm}{
\includemedia[ addresource=mp3/half-samples/half_2021-06-15T10-41-56_vocoder.mp3, flashvars={ source=mp3/half-samples/half_2021-06-15T10-41-56_vocoder.mp3 &autoPlay=false } ]{
\transparent{0.8} \color{green} \tiny \parbox{\dimexpr 0.9\linewidth-2\fboxsep-2\fboxrule\relax}{\faPlayCircleO~\textsf{Click to Play \\ in Adobe Reader \\[2.5em]}}}{APlayer.swf}
}
\PlaceText{29.08mm}{156.70999999999998mm}{
\includemedia[ addresource=mp3/half-samples/half_2021-06-15T10-42-23_vocoder.mp3, flashvars={ source=mp3/half-samples/half_2021-06-15T10-42-23_vocoder.mp3 &autoPlay=false } ]{
\transparent{0.8} \color{green} \tiny \parbox{\dimexpr 0.9\linewidth-2\fboxsep-2\fboxrule\relax}{\faPlayCircleO~\textsf{Click to Play \\ in Adobe Reader \\[2.5em]}}}{APlayer.swf}
}
\PlaceText{29.08mm}{143.91000000000003mm}{
\includemedia[ addresource=mp3/half-samples/half_2021-06-15T10-41-20_vocoder.mp3, flashvars={ source=mp3/half-samples/half_2021-06-15T10-41-20_vocoder.mp3 &autoPlay=false } ]{
\transparent{0.8} \color{green} \tiny \parbox{\dimexpr 0.9\linewidth-2\fboxsep-2\fboxrule\relax}{\faPlayCircleO~\textsf{Click to Play \\ in Adobe Reader \\[2.5em]}}}{APlayer.swf}
}
\PlaceText{29.08mm}{131.11mm}{
\includemedia[ addresource=mp3/half-samples/half_2021-06-15T10-38-56_vocoder.mp3, flashvars={ source=mp3/half-samples/half_2021-06-15T10-38-56_vocoder.mp3 &autoPlay=false } ]{
\transparent{0.8} \color{green} \tiny \parbox{\dimexpr 0.9\linewidth-2\fboxsep-2\fboxrule\relax}{\faPlayCircleO~\textsf{Click to Play \\ in Adobe Reader \\[2.5em]}}}{APlayer.swf}
}
\PlaceText{29.08mm}{118.31mm}{
\includemedia[ addresource=mp3/half-samples/half_2021-06-15T10-40-08_vocoder.mp3, flashvars={ source=mp3/half-samples/half_2021-06-15T10-40-08_vocoder.mp3 &autoPlay=false } ]{
\transparent{0.8} \color{green} \tiny \parbox{\dimexpr 0.9\linewidth-2\fboxsep-2\fboxrule\relax}{\faPlayCircleO~\textsf{Click to Play \\ in Adobe Reader \\[2.5em]}}}{APlayer.swf}
}
\PlaceText{29.08mm}{105.51mm}{
\includemedia[ addresource=mp3/half-samples/half_2021-06-15T10-38-41_vocoder.mp3, flashvars={ source=mp3/half-samples/half_2021-06-15T10-38-41_vocoder.mp3 &autoPlay=false } ]{
\transparent{0.8} \color{green} \tiny \parbox{\dimexpr 0.9\linewidth-2\fboxsep-2\fboxrule\relax}{\faPlayCircleO~\textsf{Click to Play \\ in Adobe Reader \\[2.5em]}}}{APlayer.swf}
}
\PlaceText{29.08mm}{92.71mm}{
\includemedia[ addresource=mp3/half-samples/half_2021-06-15T10-39-29_vocoder.mp3, flashvars={ source=mp3/half-samples/half_2021-06-15T10-39-29_vocoder.mp3 &autoPlay=false } ]{
\transparent{0.8} \color{green} \tiny \parbox{\dimexpr 0.9\linewidth-2\fboxsep-2\fboxrule\relax}{\faPlayCircleO~\textsf{Click to Play \\ in Adobe Reader \\[2.5em]}}}{APlayer.swf}
}
\PlaceText{29.08mm}{79.91mm}{
\includemedia[ addresource=mp3/half-samples/half_2021-06-15T10-42-48_vocoder.mp3, flashvars={ source=mp3/half-samples/half_2021-06-15T10-42-48_vocoder.mp3 &autoPlay=false } ]{
\transparent{0.8} \color{green} \tiny \parbox{\dimexpr 0.9\linewidth-2\fboxsep-2\fboxrule\relax}{\faPlayCircleO~\textsf{Click to Play \\ in Adobe Reader \\[2.5em]}}}{APlayer.swf}
}
\PlaceText{29.08mm}{67.11mm}{
\includemedia[ addresource=mp3/half-samples/half_2021-06-15T10-40-29_vocoder.mp3, flashvars={ source=mp3/half-samples/half_2021-06-15T10-40-29_vocoder.mp3 &autoPlay=false } ]{
\transparent{0.8} \color{green} \tiny \parbox{\dimexpr 0.9\linewidth-2\fboxsep-2\fboxrule\relax}{\faPlayCircleO~\textsf{Click to Play \\ in Adobe Reader \\[2.5em]}}}{APlayer.swf}
}
\PlaceText{29.08mm}{54.31mm}{
\includemedia[ addresource=mp3/half-samples/half_2021-06-15T10-39-52_vocoder.mp3, flashvars={ source=mp3/half-samples/half_2021-06-15T10-39-52_vocoder.mp3 &autoPlay=false } ]{
\transparent{0.8} \color{green} \tiny \parbox{\dimexpr 0.9\linewidth-2\fboxsep-2\fboxrule\relax}{\faPlayCircleO~\textsf{Click to Play \\ in Adobe Reader \\[2.5em]}}}{APlayer.swf}
}
\end{center}
\vspace{-2ex}
   \caption{\normalsize Priming generation with codebook codes obtained for the ground truth audio. Samples are selected randomly from the VGGSound test set. Both the codebook and the transformer are trained on audio clips from VGGSound without visual conditioning (the setting (a), \textit{No Feats}).}
\label{fig:half-gt-vggsound}
\end{figure*}

\begin{figure*}
\begin{center}
\includegraphics[width=\linewidth]{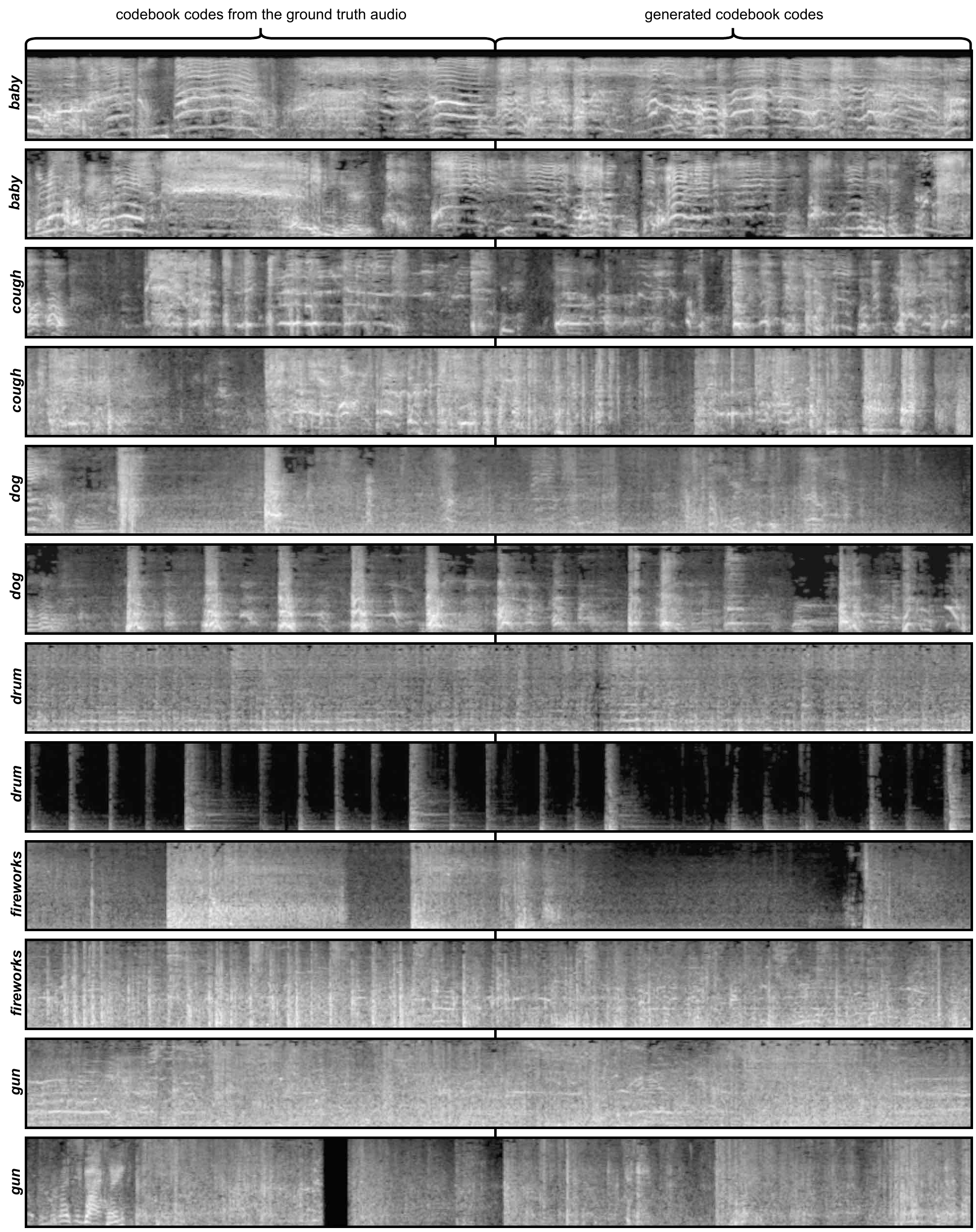}
\PlaceText{23.08mm}{196.61mm}{
\includemedia[ addresource=mp3/half-samples/half_2021-06-15T09-21-17_vocoder.mp3, flashvars={ source=mp3/half-samples/half_2021-06-15T09-21-17_vocoder.mp3 &autoPlay=false } ]{
\transparent{0.8} \color{green} \tiny \parbox{\dimexpr 0.9\linewidth-2\fboxsep-2\fboxrule\relax}{\faPlayCircleO~\textsf{Click to Play \\ in Adobe Reader \\[2.5em]}}}{APlayer.swf}
}
\PlaceText{23.08mm}{183.56mm}{
\includemedia[ addresource=mp3/half-samples/half_2021-06-15T09-21-00_vocoder.mp3, flashvars={ source=mp3/half-samples/half_2021-06-15T09-21-00_vocoder.mp3 &autoPlay=false } ]{
\transparent{0.8} \color{green} \tiny \parbox{\dimexpr 0.9\linewidth-2\fboxsep-2\fboxrule\relax}{\faPlayCircleO~\textsf{Click to Play \\ in Adobe Reader \\[2.5em]}}}{APlayer.swf}
}
\PlaceText{23.08mm}{170.51mm}{
\includemedia[ addresource=mp3/half-samples/half_2021-06-15T09-22-45_vocoder.mp3, flashvars={ source=mp3/half-samples/half_2021-06-15T09-22-45_vocoder.mp3 &autoPlay=false } ]{
\transparent{0.8} \color{green} \tiny \parbox{\dimexpr 0.9\linewidth-2\fboxsep-2\fboxrule\relax}{\faPlayCircleO~\textsf{Click to Play \\ in Adobe Reader \\[2.5em]}}}{APlayer.swf}
}
\PlaceText{23.08mm}{157.45999999999998mm}{
\includemedia[ addresource=mp3/half-samples/half_2021-06-15T09-21-44_vocoder.mp3, flashvars={ source=mp3/half-samples/half_2021-06-15T09-21-44_vocoder.mp3 &autoPlay=false } ]{
\transparent{0.8} \color{green} \tiny \parbox{\dimexpr 0.9\linewidth-2\fboxsep-2\fboxrule\relax}{\faPlayCircleO~\textsf{Click to Play \\ in Adobe Reader \\[2.5em]}}}{APlayer.swf}
}
\PlaceText{23.08mm}{144.41000000000003mm}{
\includemedia[ addresource=mp3/half-samples/half_2021-06-15T09-23-51_vocoder.mp3, flashvars={ source=mp3/half-samples/half_2021-06-15T09-23-51_vocoder.mp3 &autoPlay=false } ]{
\transparent{0.8} \color{green} \tiny \parbox{\dimexpr 0.9\linewidth-2\fboxsep-2\fboxrule\relax}{\faPlayCircleO~\textsf{Click to Play \\ in Adobe Reader \\[2.5em]}}}{APlayer.swf}
}
\PlaceText{23.08mm}{131.36mm}{
\includemedia[ addresource=mp3/half-samples/half_2021-06-15T09-23-10_vocoder.mp3, flashvars={ source=mp3/half-samples/half_2021-06-15T09-23-10_vocoder.mp3 &autoPlay=false } ]{
\transparent{0.8} \color{green} \tiny \parbox{\dimexpr 0.9\linewidth-2\fboxsep-2\fboxrule\relax}{\faPlayCircleO~\textsf{Click to Play \\ in Adobe Reader \\[2.5em]}}}{APlayer.swf}
}
\PlaceText{23.08mm}{118.30999999999999mm}{
\includemedia[ addresource=mp3/half-samples/half_2021-06-15T09-25-20_vocoder.mp3, flashvars={ source=mp3/half-samples/half_2021-06-15T09-25-20_vocoder.mp3 &autoPlay=false } ]{
\transparent{0.8} \color{green} \tiny \parbox{\dimexpr 0.9\linewidth-2\fboxsep-2\fboxrule\relax}{\faPlayCircleO~\textsf{Click to Play \\ in Adobe Reader \\[2.5em]}}}{APlayer.swf}
}
\PlaceText{23.08mm}{105.25999999999999mm}{
\includemedia[ addresource=mp3/half-samples/half_2021-06-15T09-25-40_vocoder.mp3, flashvars={ source=mp3/half-samples/half_2021-06-15T09-25-40_vocoder.mp3 &autoPlay=false } ]{
\transparent{0.8} \color{green} \tiny \parbox{\dimexpr 0.9\linewidth-2\fboxsep-2\fboxrule\relax}{\faPlayCircleO~\textsf{Click to Play \\ in Adobe Reader \\[2.5em]}}}{APlayer.swf}
}
\PlaceText{23.08mm}{92.21mm}{
\includemedia[ addresource=mp3/half-samples/half_2021-06-15T09-26-20_vocoder.mp3, flashvars={ source=mp3/half-samples/half_2021-06-15T09-26-20_vocoder.mp3 &autoPlay=false } ]{
\transparent{0.8} \color{green} \tiny \parbox{\dimexpr 0.9\linewidth-2\fboxsep-2\fboxrule\relax}{\faPlayCircleO~\textsf{Click to Play \\ in Adobe Reader \\[2.5em]}}}{APlayer.swf}
}
\PlaceText{23.08mm}{79.16mm}{
\includemedia[ addresource=mp3/half-samples/half_2021-06-15T09-26-02_vocoder.mp3, flashvars={ source=mp3/half-samples/half_2021-06-15T09-26-02_vocoder.mp3 &autoPlay=false } ]{
\transparent{0.8} \color{green} \tiny \parbox{\dimexpr 0.9\linewidth-2\fboxsep-2\fboxrule\relax}{\faPlayCircleO~\textsf{Click to Play \\ in Adobe Reader \\[2.5em]}}}{APlayer.swf}
}
\PlaceText{23.08mm}{66.11mm}{
\includemedia[ addresource=mp3/half-samples/half_2021-06-15T09-27-10_vocoder.mp3, flashvars={ source=mp3/half-samples/half_2021-06-15T09-27-10_vocoder.mp3 &autoPlay=false } ]{
\transparent{0.8} \color{green} \tiny \parbox{\dimexpr 0.9\linewidth-2\fboxsep-2\fboxrule\relax}{\faPlayCircleO~\textsf{Click to Play \\ in Adobe Reader \\[2.5em]}}}{APlayer.swf}
}
\PlaceText{23.08mm}{53.06mm}{
\includemedia[ addresource=mp3/half-samples/half_2021-06-15T09-26-51_vocoder.mp3, flashvars={ source=mp3/half-samples/half_2021-06-15T09-26-51_vocoder.mp3 &autoPlay=false } ]{
\transparent{0.8} \color{green} \tiny \parbox{\dimexpr 0.9\linewidth-2\fboxsep-2\fboxrule\relax}{\faPlayCircleO~\textsf{Click to Play \\ in Adobe Reader \\[2.5em]}}}{APlayer.swf}
}
\end{center}
\vspace{-2ex}
   \caption{\normalsize Priming generation with codebook codes obtained for the ground truth audio. Samples are selected randomly from the VAS validation set. Both the codebook and the transformer are trained on audio clips from VAS without visual conditioning (the setting (c), \textit{No Feats}).}
\label{fig:half-gt-vas}
\end{figure*}

\paragraph{Priming Sampling with a Ground Truth Segment}
Figures~\ref{fig:half-gt-vggsound} and \ref{fig:half-gt-vas} illustrate the ability of the sampler to seamlessly continue a ground truth audio with a relevant and novel segment.

\begin{figure*}
\begin{center}
\vspace{-1ex}
\includegraphics[width=\linewidth]{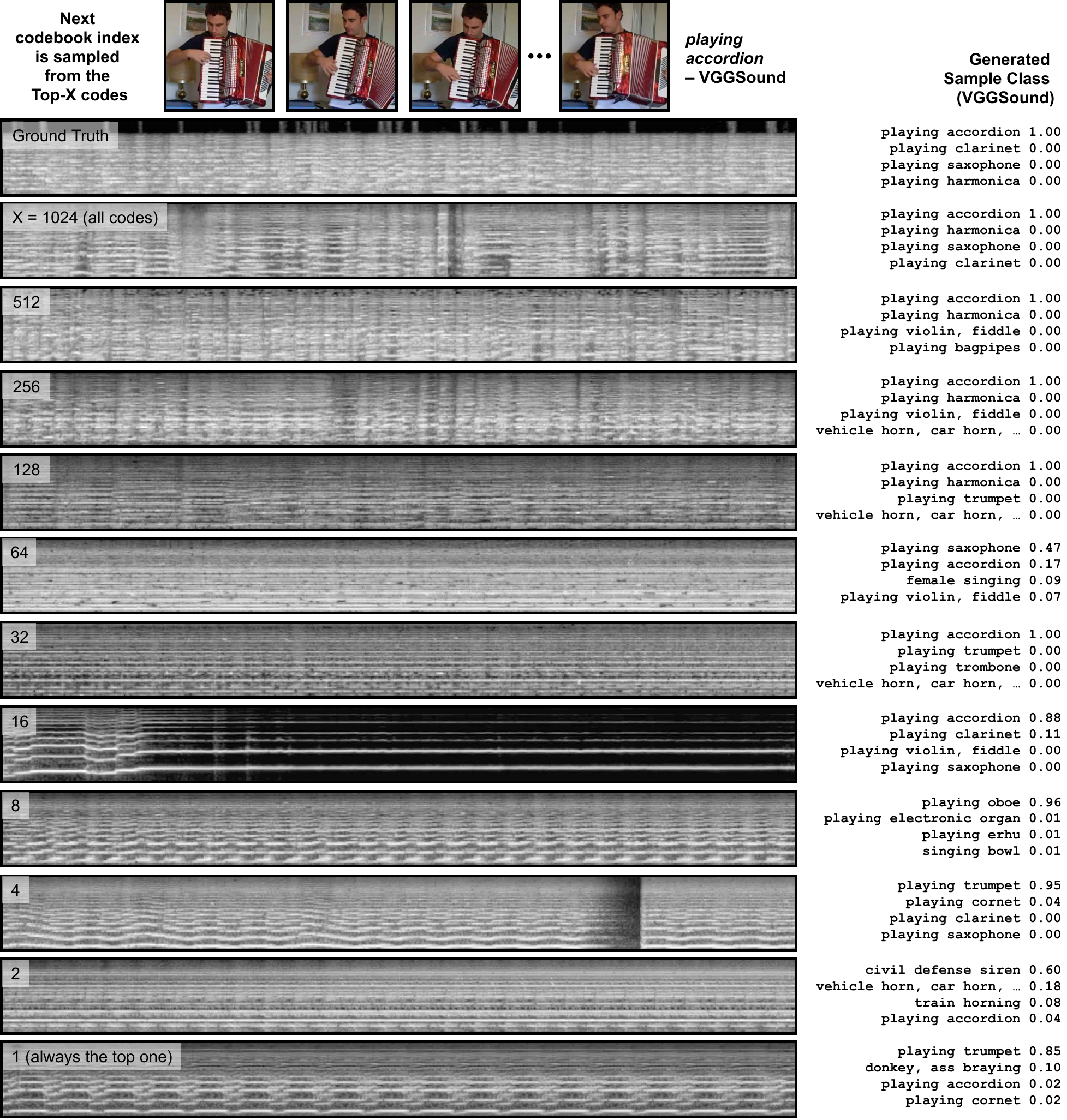}
\PlaceText{44.08mm}{198.61mm}{
\includemedia[ addresource=mp3/topx/JUBdOr8Hes_30000_40000_gt.mp3, flashvars={ source=mp3/topk/JUBdOr8Hes_30000_40000_gt.mp3 &autoPlay=false } ]{
\transparent{0.8} \color{green} \tiny \parbox{\dimexpr 0.6\linewidth-2\fboxsep-2\fboxrule\relax}{\faPlayCircleO~\textsf{Click to Play \\ in Adobe Reader \\[2em]}}}{APlayer.swf}
}
\PlaceText{44.08mm}{188.51mm}{
\includemedia[ addresource=mp3/topx/nopix_2021-06-16T09-14-42_vocoder.mp3, flashvars={ source=mp3/topk/nopix_2021-06-16T09-14-42_vocoder.mp3 &autoPlay=false } ]{
\transparent{0.8} \color{green} \tiny \parbox{\dimexpr 0.6\linewidth-2\fboxsep-2\fboxrule\relax}{\faPlayCircleO~\textsf{Click to Play \\ in Adobe Reader \\[2em]}}}{APlayer.swf}
}
\PlaceText{44.08mm}{178.41000000000003mm}{
\includemedia[ addresource=mp3/topx/nopix_2021-06-16T09-15-36_vocoder.mp3, flashvars={ source=mp3/topk/nopix_2021-06-16T09-15-36_vocoder.mp3 &autoPlay=false } ]{
\transparent{0.8} \color{green} \tiny \parbox{\dimexpr 0.6\linewidth-2\fboxsep-2\fboxrule\relax}{\faPlayCircleO~\textsf{Click to Play \\ in Adobe Reader \\[2em]}}}{APlayer.swf}
}
\PlaceText{44.08mm}{168.31mm}{
\includemedia[ addresource=mp3/topx/nopix_2021-06-16T09-15-59_vocoder.mp3, flashvars={ source=mp3/topk/nopix_2021-06-16T09-15-59_vocoder.mp3 &autoPlay=false } ]{
\transparent{0.8} \color{green} \tiny \parbox{\dimexpr 0.6\linewidth-2\fboxsep-2\fboxrule\relax}{\faPlayCircleO~\textsf{Click to Play \\ in Adobe Reader \\[2em]}}}{APlayer.swf}
}
\PlaceText{44.08mm}{158.20999999999998mm}{
\includemedia[ addresource=mp3/topx/nopix_2021-06-16T09-16-18_vocoder.mp3, flashvars={ source=mp3/topk/nopix_2021-06-16T09-16-18_vocoder.mp3 &autoPlay=false } ]{
\transparent{0.8} \color{green} \tiny \parbox{\dimexpr 0.6\linewidth-2\fboxsep-2\fboxrule\relax}{\faPlayCircleO~\textsf{Click to Play \\ in Adobe Reader \\[2em]}}}{APlayer.swf}
}
\PlaceText{44.08mm}{148.11mm}{
\includemedia[ addresource=mp3/topx/nopix_2021-06-16T09-16-43_vocoder.mp3, flashvars={ source=mp3/topk/nopix_2021-06-16T09-16-43_vocoder.mp3 &autoPlay=false } ]{
\transparent{0.8} \color{green} \tiny \parbox{\dimexpr 0.6\linewidth-2\fboxsep-2\fboxrule\relax}{\faPlayCircleO~\textsf{Click to Play \\ in Adobe Reader \\[2em]}}}{APlayer.swf}
}
\PlaceText{44.08mm}{138.01mm}{
\includemedia[ addresource=mp3/topx/nopix_2021-06-16T09-17-08_vocoder.mp3, flashvars={ source=mp3/topk/nopix_2021-06-16T09-17-08_vocoder.mp3 &autoPlay=false } ]{
\transparent{0.8} \color{green} \tiny \parbox{\dimexpr 0.6\linewidth-2\fboxsep-2\fboxrule\relax}{\faPlayCircleO~\textsf{Click to Play \\ in Adobe Reader \\[2em]}}}{APlayer.swf}
}
\PlaceText{44.08mm}{127.91mm}{
\includemedia[ addresource=mp3/topx/nopix_2021-06-16T09-17-49_vocoder.mp3, flashvars={ source=mp3/topk/nopix_2021-06-16T09-17-49_vocoder.mp3 &autoPlay=false } ]{
\transparent{0.8} \color{green} \tiny \parbox{\dimexpr 0.6\linewidth-2\fboxsep-2\fboxrule\relax}{\faPlayCircleO~\textsf{Click to Play \\ in Adobe Reader \\[2em]}}}{APlayer.swf}
}
\PlaceText{44.08mm}{117.81mm}{
\includemedia[ addresource=mp3/topx/nopix_2021-06-16T09-18-07_vocoder.mp3, flashvars={ source=mp3/topk/nopix_2021-06-16T09-18-07_vocoder.mp3 &autoPlay=false } ]{
\transparent{0.8} \color{green} \tiny \parbox{\dimexpr 0.6\linewidth-2\fboxsep-2\fboxrule\relax}{\faPlayCircleO~\textsf{Click to Play \\ in Adobe Reader \\[2em]}}}{APlayer.swf}
}
\PlaceText{44.08mm}{107.71mm}{
\includemedia[ addresource=mp3/topx/nopix_2021-06-16T09-18-33_vocoder.mp3, flashvars={ source=mp3/topk/nopix_2021-06-16T09-18-33_vocoder.mp3 &autoPlay=false } ]{
\transparent{0.8} \color{green} \tiny \parbox{\dimexpr 0.6\linewidth-2\fboxsep-2\fboxrule\relax}{\faPlayCircleO~\textsf{Click to Play \\ in Adobe Reader \\[2em]}}}{APlayer.swf}
}
\PlaceText{44.08mm}{97.61mm}{
\includemedia[ addresource=mp3/topx/nopix_2021-06-16T09-19-00_vocoder.mp3, flashvars={ source=mp3/topk/nopix_2021-06-16T09-19-00_vocoder.mp3 &autoPlay=false } ]{
\transparent{0.8} \color{green} \tiny \parbox{\dimexpr 0.6\linewidth-2\fboxsep-2\fboxrule\relax}{\faPlayCircleO~\textsf{Click to Play \\ in Adobe Reader \\[2em]}}}{APlayer.swf}
}
\PlaceText{44.08mm}{87.51mm}{
\includemedia[ addresource=mp3/topx/nopix_2021-06-16T09-19-22_vocoder.mp3, flashvars={ source=mp3/topk/nopix_2021-06-16T09-19-22_vocoder.mp3 &autoPlay=false } ]{
\transparent{0.8} \color{green} \tiny \parbox{\dimexpr 0.6\linewidth-2\fboxsep-2\fboxrule\relax}{\faPlayCircleO~\textsf{Click to Play \\ in Adobe Reader \\[2em]}}}{APlayer.swf}
}
\end{center}
\vspace{-5ex}
   \caption{\normalsize Clipping the distribution of the next codebook token to Top-$X$ probabilities provides a control for sample's diversity. $X$ ranges from $X = |\mathcal{Z}|=1024$ to $X=1$ when sampling using the setting (a) with 5 Feats (see Section~\ref{sec:results_sampling}, and more details in Section~\ref{sec:additional-results}). }
\vspace{-2ex}
\label{fig:topx}
\end{figure*}

\begin{figure*}
\begin{center}
\includegraphics[width=\linewidth]{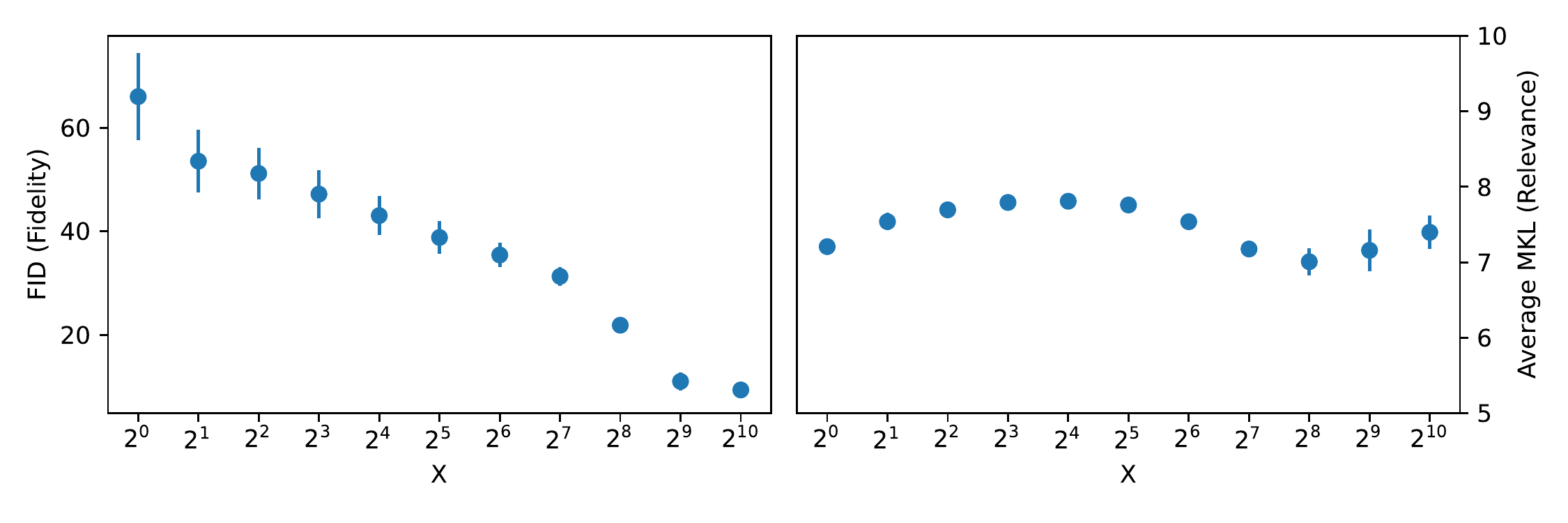}
\end{center}
\vspace{-4ex}
   \caption{\normalsize Impact of the Top-$X$ clipping on fidelity and relevance. The metrics are averaged among all visually conditioned models in the setting (a), see Section~\ref{sec:results_sampling}.}
\label{fig:trade-off}
\end{figure*}

\paragraph{Controlling for Sample Diversity}
Temporal diversity is an important factor of high-fidelity audio.
At the same time, it is challenging to generate a diverse but relevant sample which imposes a trade-off.
Considering the autoregressive property of our sampler, we can control the diversity of the generated sample by clipping the distribution over the next token during sampling.
As mentioned in Section~\ref{sec:second_stage}, at the autoregressive step, the transformer outputs a distribution over the next codebook index $p\big(s_j|\hat{s}_{<j})$, see Eq.~\eqref{eq:autoregressive-step}.
We sample from the multinomial distribution using the weights provided by $p$ that can be clipped to only Top-$X$ probabilities where $X \in \big[1, |\mathcal{Z}|\big]$.
As a result, lower values of $X$ will yield clean, texture-like, and repeating-pattern sounds (low temporal diversity) while higher values will provide more ``random'', less relevant, multi-class samples (high diversity).
The effect is shown in Figures~\ref{fig:topx} \& \ref{fig:trade-off}. 
Another important observation is related to the inability of the Melception to have a strong response of the class on lower $X$ (see the ``Generated Sample Class'' column in Figure~\ref{fig:topx}). 
It is a consequence of the fact that such low diversity examples never occurred during the training of Melception, yet, the audios generated at the lower $X$ do sound as, \eg, \textit{accordion} or another similar instrument.

\begin{figure*}
\begin{center}
\includegraphics[width=\linewidth]{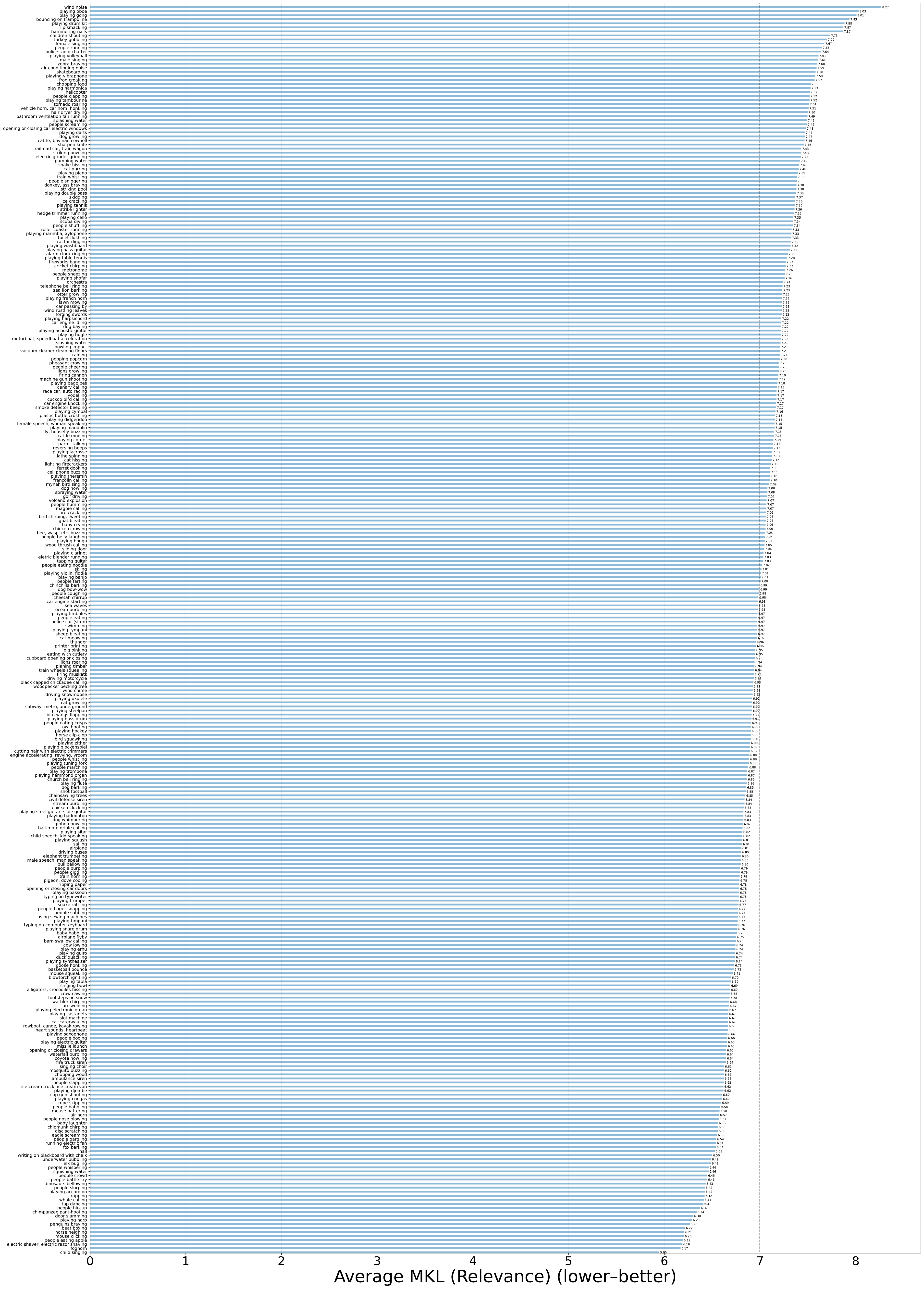}
\end{center}
\vspace{-2ex}
   \caption{\normalsize Relevance per class. We generate 10 samples per video from the VGGSound test set. Each class has ~45 (not more than 50) videos. The model is trained on the VGGSound dataset with VGGSound codebook (setting (a) with\textit{ 5 Feats} (see Section~\ref{sec:results_sampling}). The lower, the more relevant. The average performance is 7.0. Best viewed when zoomed-in.}
\label{fig:relevance-per-class}
\end{figure*}

\paragraph{Relevance per Class}
In Figure~\ref{fig:relevance-per-class} we show how relevant the generated samples are across every class in the VGGSound dataset.
We rely on the setting (a), see Section~\ref{sec:results_sampling} with 5 visual features in condition.
Despite that model performance on a majority of the classes fall into [$7 \pm0.7$] interval of the MKL yet there is still room for improvement in the capabilities of a model to handle multiple classes which we hope to see in future research. 

\begin{figure*}
\begin{center}
\vspace{-10em}
\includegraphics[width=\linewidth]{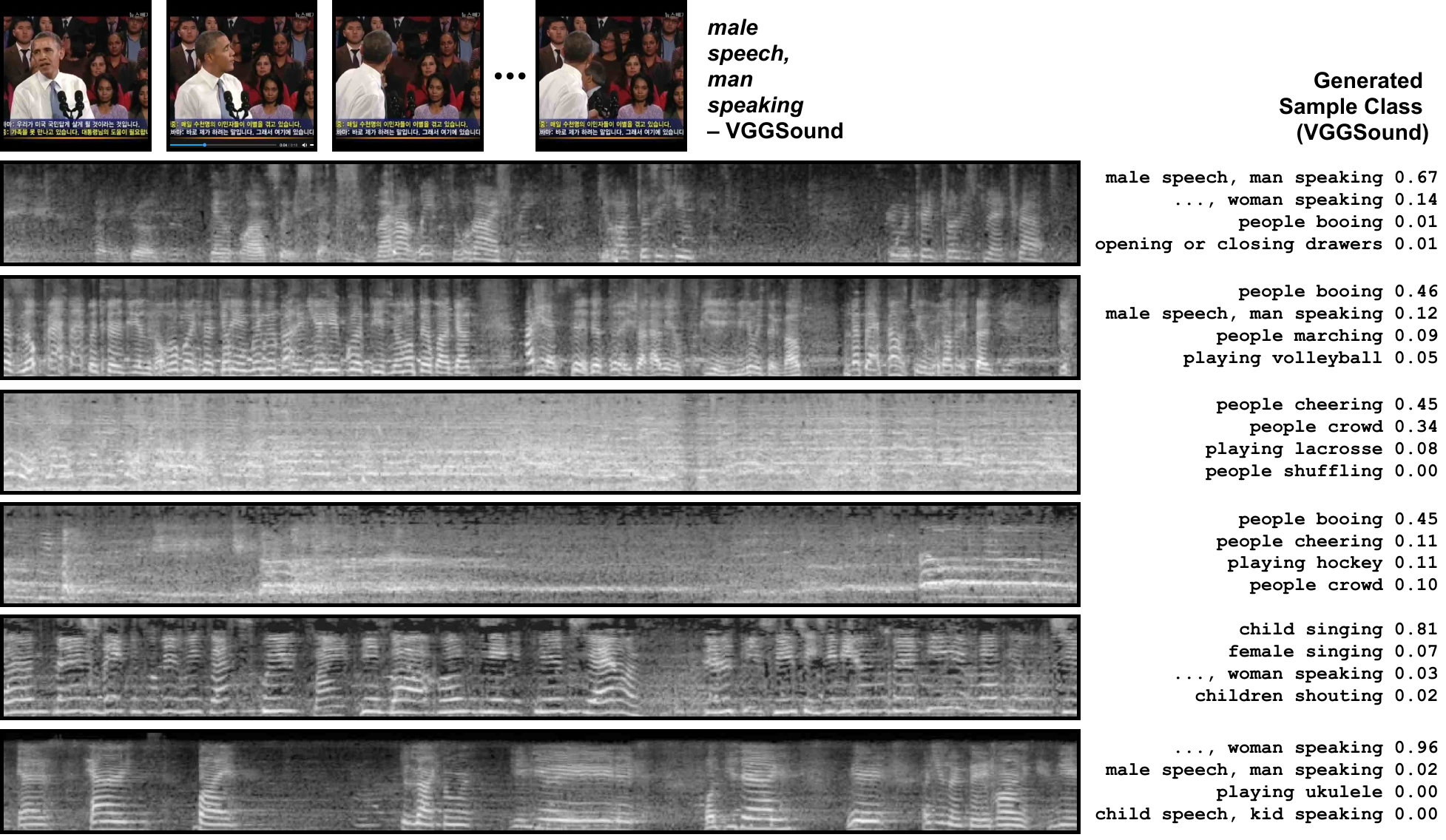}
\PlaceText{20.08mm}{161.11mm}{
\includemedia[ addresource=mp3/variability-in-video/nopix_2021-06-16T11-20-05_vocoder.mp3, flashvars={ source=mp3/variability-in-video/nopix_2021-06-16T11-20-05_vocoder.mp3 &autoPlay=false } ]{
\transparent{0.8} \color{green} \tiny \parbox{\dimexpr 0.7\linewidth-2\fboxsep-2\fboxrule\relax}{\faPlayCircleO~\textsf{Click to Play \\ in Adobe Reader \\[2em]}}}{APlayer.swf}
}
\PlaceText{20.08mm}{151.01mm}{
\includemedia[ addresource=mp3/variability-in-video/nopix_2021-06-16T11-19-50_vocoder.mp3, flashvars={ source=mp3/variability-in-video/nopix_2021-06-16T11-19-50_vocoder.mp3 &autoPlay=false } ]{
\transparent{0.8} \color{green} \tiny \parbox{\dimexpr 0.7\linewidth-2\fboxsep-2\fboxrule\relax}{\faPlayCircleO~\textsf{Click to Play \\ in Adobe Reader \\[2em]}}}{APlayer.swf}
}
\PlaceText{20.08mm}{140.91mm}{
\includemedia[ addresource=mp3/variability-in-video/nopix_2021-06-16T11-18-20_vocoder.mp3, flashvars={ source=mp3/variability-in-video/nopix_2021-06-16T11-18-20_vocoder.mp3 &autoPlay=false } ]{
\transparent{0.8} \color{green} \tiny \parbox{\dimexpr 0.7\linewidth-2\fboxsep-2\fboxrule\relax}{\faPlayCircleO~\textsf{Click to Play \\ in Adobe Reader \\[2em]}}}{APlayer.swf}
}
\PlaceText{20.08mm}{130.81mm}{
\includemedia[ addresource=mp3/variability-in-video/nopix_2021-06-16T11-18-51_vocoder.mp3, flashvars={ source=mp3/variability-in-video/nopix_2021-06-16T11-18-51_vocoder.mp3 &autoPlay=false } ]{
\transparent{0.8} \color{green} \tiny \parbox{\dimexpr 0.7\linewidth-2\fboxsep-2\fboxrule\relax}{\faPlayCircleO~\textsf{Click to Play \\ in Adobe Reader \\[2em]}}}{APlayer.swf}
}
\PlaceText{20.08mm}{120.71mm}{
\includemedia[ addresource=mp3/variability-in-video/nopix_2021-06-16T11-19-21_vocoder.mp3, flashvars={ source=mp3/variability-in-video/nopix_2021-06-16T11-19-21_vocoder.mp3 &autoPlay=false } ]{
\transparent{0.8} \color{green} \tiny \parbox{\dimexpr 0.7\linewidth-2\fboxsep-2\fboxrule\relax}{\faPlayCircleO~\textsf{Click to Play \\ in Adobe Reader \\[2em]}}}{APlayer.swf}
}
\PlaceText{20.08mm}{110.61mm}{
\includemedia[ addresource=mp3/variability-in-video/nopix_2021-06-16T11-20-50_vocoder.mp3, flashvars={ source=mp3/variability-in-video/nopix_2021-06-16T11-20-50_vocoder.mp3 &autoPlay=false } ]{
\transparent{0.8} \color{green} \tiny \parbox{\dimexpr 0.7\linewidth-2\fboxsep-2\fboxrule\relax}{\faPlayCircleO~\textsf{Click to Play \\ in Adobe Reader \\[2em]}}}{APlayer.swf}
}
\end{center}
\vspace{-2ex}
   \caption{\normalsize Given the same visual condition, a model randomly samples different appearances relevant to the input. Both the codebook and the transformer are trained on audio clips from VGGSound (the setting (a), \textit{5 Feats})}
\label{fig:variability_in_video}
\end{figure*}

\paragraph{Variability of Samples in One Video}
Since the generation of a relevant sound given a set of visual features is an ill-posted problem, a model is expected to \textit{sample} unique, diverse, and relevant sounds.
In Figure~\ref{fig:variability_in_video} we provide several samples generated given the same set of visual features.
Specifically, given the provided visual sequence (``a person is talking with a crowd in the background''), it is difficult to guess why the person turned back to the crowd, \eg because they were ``cheering'' or ``booing''.
At the same time, we also notice the limitation of the model, i.\,e. sometimes it confuses the gender or age of a person.
However, we believe these are reasonable mistakes considering the difficulty of the scene.

\paragraph{Qualitative Analysis of the Model Properties (dataset-related issues)}

Continuing the Section~\ref{sec:qualitative-analysis}, VGGSound \cite{chen2020vggsound} dataset is the largest open-domain dataset with strong audio-visual correspondence.
However, due to the automatic nature of the collection procedure, mistakes are not rare.
For instance, we discovered that the class \textit{cupboard opening or closing} has no sounds in neither of the test videos and the videos were from a 3D render.
It leads to a property of Melception, which was trained on VGGSound, to classify a silent audio sample as \textit{cupboard opening or closing} with high confidence.
Also, \textit{playing t\textbf{i}mpani} and \textit{playing t\textbf{y}mpani} are considered as different classes.
Sometimes, most of the test videos for a class are irrelevant.
For example, 95\% of all videos in the \textit{baby babbling} class are not relevant, \textit{strike lighter}, \textit{dog whimpering} with mostly people crying videos, or \textit{metronome} class with a majority of musical videos.
Moreover, there are many pairs/groups of classes which are difficult to differentiate and often have similar examples.
For instance, \textit{airplane} and \textit{airplane flyby}; \textit{wind noise} and \textit{wind rustling leaves}.
Therefore, a more curated large-scale dataset is needed to further improve the performance of the models.

\paragraph{Spectrogram VQGAN as a Neural Audio Codec}
A recent (July 2021) arXiv submission \cite{zeghidour2021soundstream} show-cased a VQVAE architecture with adversarial 
loss, called \textit{SoundStream}, on lossy compression of a waveform with the state-of-the-art results on the 3 kbps bitrate which
works on music and speech datasets.
Since our approach includes sampling from a pre-trained codebook, we can employ our Spectrogram VQGAN pre-trained on
an open-domain dataset as a  neural audio codec \textit{without a change}. 

The bitrate is governed by the size of the codebook (in our case: 1024 i.\,e.\,\,10 bits per code) and the bottleneck 
size ($5\times 53$ for a $9.8$-second audio sample, in our case). Therefore, our approach allows encoding at, approximately, 
\textbf{0.27 kbps} bitrate with the VGGSound codebook and \textbf{0.19 kbps} with the VAS codebook. 

We provide a qualitative comparison of reconstructions of Lyra \cite{kleijn2021generative} (only speech), SoundStream 
(only speech and music), and Spectrogram VQGAN (open-domain) in the supplementary and our project page.
We use the same 3-second examples as provided on the SoundStream project page since the source code for the 
SoundStream has not been released to the public by the time of this submission. 

As a result, despite having one order of magnitude smaller bitrate budget, Spectrogram VQGAN achieves comparable performance with SoundStream in reconstruction quality on music data and produces significantly better reconstructions than Lyra.
However, as we observed before, Spectrogram VQGAN struggles with the fine details of human speech due to the audio preprocessing
(mel-scale spectrogram) and absence of narrow domain pre-training as in Lyra and SoundStream.
We highlight the fact that Spectrogram VQGAN is trained on an open-domain hundred-class dataset (VGGSound) while SoundStream
is trained on music and speech datasets separately.

\vspace{-2ex}

 
\bibliography{src/bibliography}

\end{document}